\documentclass{article}
\usepackage{arxiv_blf,times}

\usepackage[utf8]{inputenc}
\usepackage[T1]{fontenc}
\usepackage{hyperref}
\hypersetup{colorlinks=true,linkcolor=blue,citecolor=blue,urlcolor=blue}
\usepackage{url}
\usepackage{booktabs}
\usepackage{array}
\usepackage{amsfonts}
\usepackage{amsmath}
\usepackage{nicefrac}
\usepackage{microtype}
\usepackage{xcolor}
\usepackage{graphicx}
\usepackage{subcaption}
\usepackage{multirow}
\usepackage{threeparttable}

\usepackage[capitalize,nameinlink]{cleveref}
\crefname{figure}{Fig.}{Figs.}                                                
\crefname{table}{Tab.}{Tabs.}                                                                                                                                             
\crefname{section}{Sec.}{Secs.} 

\usepackage{xspace}
\usepackage{tikz}
\usetikzlibrary{positioning, arrows.meta, calc, shapes.geometric}
\usepackage{algorithm}
\usepackage{algorithmic}
\usepackage{xspace}

\newcommand{\BI}{\text{BI}\xspace}
\newcommand{\BS}{\text{BS}\xspace}

\newcommand{\MBS}{\text{MS}\xspace} 
\newcommand{\eat}[1]{} 
\newcommand{\union}{\xspace \cup \xspace}

\newcommand{\safeLinear}{\text{safe-linear}}

\newcommand{\browseTool}{\texttt{browse-web}\xspace}
\newcommand{\readTool}{\texttt{read-files}\xspace}
\newcommand{\histTool}{\texttt{history-fetch}\xspace}
\newcommand{\modelTool}{\texttt{model-fetch}\xspace}
\newcommand{\comboTool}{\texttt{combo-fetch}\xspace}
\newcommand{\wikiTool}{\texttt{wikipedia-fetch}\xspace}
\newcommand{\urlTool}{\texttt{url-lookup}\xspace}
\newcommand{\submitTool}{\texttt{submit}\xspace}

\newcommand{\actions}{\mathcal{A}}
\newcommand{\sigmoid}{\text{sigmoid}}

\newcommand{\tb}{\text{b}}
\newcommand{\tr}{\text{r}}

\newcommand{\tc}{\text{c}}
\renewcommand{\th}{\text{h}}
\newcommand{\tm}{\text{m}}
\newcommand{\tw}{\text{w}}
\newcommand{\tx}{\text{x}}
\newcommand{\tu}{\text{u}}

\newcommand{\tools}{\mathcal{T}}

\newcommand{\toolsBasic}{\mathcal{T}_{\text{basic}}}
\newcommand{\toolsSearch}{\mathcal{T}_{\text{search}}}

\newcommand{\bi}[1]{#1}    
\newcommand{\dbi}[1]{#1}   
\newcommand{\ms}[1]{#1}    
\newcommand{\Bbi}[1]{\textbf{#1}}   
\newcommand{\Bms}[1]{\textbf{#1}}   

\newcommand{\maxSteps}{10}
\newcommand{\maxTrials}{5}
\newcommand{\nqAB}{400}        
\newcommand{\nrdAB}{791}       

\newcommand{\system}{\texttt{BLF}\xspace}

\newcommand{\BLF}{\texttt{BLF}\xspace}
\newcommand{\uncond}{\ensuremath{\mathrm{s}}\xspace}
\newcommand{\acond}{\ensuremath{\mathrm{sa}}\xspace}
\newcommand{\bcond}{\ensuremath{\mathrm{sab}}\xspace}
\newcommand{\hcond}{\ensuremath{\mathrm{sah}}\xspace}
\newcommand{\hbcond}{\ensuremath{\mathrm{sabh}}\xspace}

\newcommand{\sabcond}{\ensuremath{\sigma\text{-}\mathrm{sab}}\xspace}
\newcommand{\sahbcond}{\ensuremath{\sigma\text{-}\mathrm{sab}\sigma}\xspace}

\newcommand{\Pro}{\textsc{Pro}\xspace}
\newcommand{\Flash}{\textsc{Flash}\xspace}
\newcommand{\GPT}{\textsc{GPT}\xspace}

\title{Agentic Forecasting with Structured Linguistic Beliefs}

\author{
  Kevin Murphy \\
   Department of Computer Science\\
  University of British Columbia\\
  Vancouver, BC, Canada \\
  \texttt{murphyk@cs.ubc.ca} 
}
\date{\today}

\begin{document}

\maketitle

\begin{abstract}
  We propose  the Bayesian Linguistic Forecaster (BLF),  which is a search-based ReAct style agent for (binary) forecasting  that augments its context with a linguistic 'belief state', consisting of a probability estimate and a \emph{structured} natural-language evidence summary. (The belief update is 'Bayesian-inspired' rather than exact Bayesian inference, although we do include a comparison to a more explicit Bayesian approach using LLM-elicited likelihood surrogates.) On 400 historical ForecastBench question-date instances, BLF has the highest overall Brier Index among the compared external agents, and is the only compared method to significantly outperform a crowd+empirical-prior baseline in terms of overall performance.  Our backtesting framework combines date-aware search tools, leakage auditing, and paired comparisons for variance control to provide a rigorous empirical framework for comparing agentic forecasting systems.
\end{abstract}

\section{Introduction}
\label{sec:intro}

Forecasting the probability of future events is a
fundamental challenge with applications in
geopolitics, finance, and public
health~\citep{Tetlock2015,Spiegelhalter2025}.
Recent work has shown that LLMs can approach human-level
forecasting when given web search
access~\citep{halawi2024approaching},
and benchmarks such as
ForecastBench~\citep{forecastbench}
provide standardized evaluation with online leaderboards.

Most forecasting agents
use some kind of ReAct (Reason-then-Act) loop \citep{ReAct},
where the agent
sequentially accumulates search results in context.
However,
this can cause problems with various forms of
``context rot'' (see e.g., \citep{Breunig2025,Liu2026}),
We present \system{} (Bayesian Linguistic Forecaster),
which augments the agent's history with a \emph{semi-structured belief state}
--- a probability estimate paired with natural-language evidence
summaries --- updated by the LLM at each step.
We refer to this loosely as ``Bayesian-style'' updating: the
process is designed to mirror the form of a sequential Bayesian
update (prior + likelihood $\to$ posterior), but the actual
update is just an LLM forward pass.\footnote{
Because we use the LLM to predict the probability at each step,
our system  may not satisfy
the consistency criteria required to correspond to proper Bayesian inference
\citep{Qiu2025,Falck2024}.
In \cref{sec:explicit-bayes}, we compare to an
explicit Bayesian updating procedure using LLM-estimated likelihoods
(or likelihood ratios),
but we show that this gives
lower point-estimate accuracy than our heuristic method.
}

To evaluate \system{} (and other agents),
we develop a four-layer date-leakage defense for backtesting,
which tackles a key problem in the field \citep{Paleka2025}.
Our post-hoc audit shows an approximate
residual leakage rate of only 1.5\%  of retained search results
(\cref{app:leakage}).
In addition, we adopt
a statistical approach to comparing
methods based on     paired analysis with bootstrap confidence intervals
(\cref{app:mixed-effects}).
This controls for the confounding factor of
varying question difficulty,
which  accounts for 62\% of the performance variance (\cref{app:anova}).
Pairing is useful when differences between methods are small,
but has been overlooked by most previous empirical work in this field.

\medskip
\noindent
On \nqAB{} backtesting questions from ForecastBench (FB),
we find that \system{} achieves the highest Brier Index (BI) score on every question
type (Overall, Market, Dataset) compared against the top
methods on the FB leaderboard at the time the work was done
(2026-04-20, see \cref{fig:fb-leaderboard}).
Although LLMs (and the leaderboard) have  been updated since then,
we believe our frozen results still present
a useful comparison of different methods that is still relevant.

\section{Experimental Setup}
\label{sec:setup}

\paragraph{Problem definition.}
\label{sec:problem}

We focus on binary prediction problems of the following type.
Let $Y(t)$ denote the random variable of interest at time $t$.
Let $d$ be the forecast date and $r>d$ be the resolution date.
Our task is to estimate
$P(Y(r)=1 \mid \text{data}(\leq d))$.
We consider two main kinds of questions:
 open-ended text questions
 (e.g., ``Will person X be the president of country C at time $r$?''),
 which  require 
 ``judgemental forecasting''  \citep{lawrence2006};
 and  more traditional time-series forecasting problems
(e.g., ``Will the stock price of company C at time $r$ be greater than its
current value of $v$?''),
which require numerical reasoning, but may also leverage text data where relevant.\footnote{
These binarized time series forecasting problems may seem rather artificial.
However, prediction markets such as Kalshi and Polymarket do contain similar binary questions,
where a continuous outcome is binned into different (mutually exclusive) buckets,
collectively called a "market" \citep{ProphetArena}, 
and users are asked to bet on individual binary 
events.
For example,
\url{https://polymarket.com/event/what-price-will-bitcoin-hit-in-may-2026}
asks ''What price will Bitcoin hit in May?''
with choices such as $<30,000$, $<35,000$, up to $>150,000$ (in USD).
This question was posted on 2026-05-01 and will be resolved on 2026-06-01.
}

\paragraph{Datasets.}
\label{sec:datasets}
We focus on our results on
ForecastBench (FB)~\citep{forecastbench}.
This consists of judgemental forecasting questions 
 from market sources (Polymarket, Manifold, Metaculus, Rand Forecasting Initiative (RFI)),
and (binarized) time series forecasting problems from dataset sources
(yfinance, FRED, DBnomics, Wikipedia, ACLED).
(Formally, the dataset questions require estimating
$P(Y(r_i) > v_d \mid \text{data}(\leq d))$,
where  $v_d$ is a reference value available at time $d$,
and $r_i=d+h_i$ is the resolution date
at up to 8 forecast horizons $h_i$.)

We construct two evaluation tranches from FB (A: 2025-10-26, B: 2025-11-09),
each with 100 market + 100 dataset questions
($\nqAB$ question-date instances across A$\cup$B; 355 distinct underlying questions). Each market question
resolves at a single date, while each dataset question may resolve
at multiple horizons; the 200 market questions yield 200 binary
resolution events, the 200 dataset questions yield 591 events
(roughly three resolved horizons per question, since the 7/30/90-day
horizons resolve within our window), giving $\nrdAB$ binary
resolution events in total.
The dates for these tranches were chosen to meet 3 criteria:
(1) be after the knowledge cutoff of frontier LLMs
(at the time the experiments were done);
(2) maximize number of resolved questions;
(3) maximize overlap with results from publicly submitted
forecasts, to enable paired comparisons.
See \cref{app:FB} for details.

\paragraph{Metrics.}
\label{sec:metrics}
A standard metric for binary forecasters
is the (mean) Brier Score, defined as
$\overline{\BS} = \frac{1}{N}\sum_i (p_i - o_i)^2$,
where
    $p_i \in [0,1]$ is the $i$'th prediction,  $o_i \in \{0,1\}$
is the $i$'th outcome, and $N$ is the number of predictions.
For BS, the best score is 0, and a chance system scores 0.25.
 In \citep{BrierIndex}, they proposed
 the Brier Index, whose values are  slightly more  intuitive.
This is defined as 
$\BI = 100 \times (1 - \sqrt{\overline{\BS}})$,
so the best system scores 100\%, the worst 0\%,
and a chance system scores 50\%.
See \cref{app:metrics} for discussion of other metrics.
When evaluating on FB, overall scores are the unweighted average of market and dataset means,
following the FB methodology.

\paragraph{Backtesting validity.}
\label{sec:leakage}
To minimize parametric leakage,
we select LLMs  whose knowledge cutoff is before
ForecastBench tranche~A (2025-10-26) and tranche~B (2025-11-09).
To minimize leakage from search and tool use, 
 we implement
a four-layer defense:
(1)~search engine date filtering,
(2)~LLM-based leak classifier on results,
(3)~data tool date clamping, and
(4)~URL blocking for resolution sources.
A second-model audit flags post-cutoff information in 1.5\% of retained search results.
Advertised knowledge cutoffs and date filters do not guarantee absence of leakage.
See \cref{app:leakage} for details.

\paragraph{LLMs.}
For the experiments we run (as opposed to historical FB results
we extract from the leaderboard),
we use 3 LLMs:
\Pro=Gemini-3.1-Pro;
\Flash=Gemini-3-Flash;
\GPT=GPT-5.4.
See \cref{tab:llms} for details on these models.

\paragraph{External baselines.}
\label{sec:alternatives}
For our FB experiments, the main external comparison uses three
methods that were top of leaderboard at the time the work was done
(2026-04-20, see \cref{fig:fb-leaderboard}):
Cassi~\citep{Cassi},
Grok~4.20,
and Foresight-32B~\citep{Foresight}.
All these methods use the crowd estimate for market questions as part of their prompt,
and all use search and tools
(see  \cref{app:alternatives} for details).
We also include
\emph{Crowd+emp}, an LLM-free baseline using market prices for market questions
and empirical priors for dataset questions. The NoBel BLF ablation
uses the same search-enabled loop as BLF but without a structured belief state.

\section{Method}
\label{sec:system}
\label{sec:method}

\paragraph{Main loop.}
BLF is a variant of the standard ReAct (Reason-then-Act) agent design  \citep{ReAct},
augmented with structured belief tracking. Independent rollouts can then be
combined into one forecast.
The main loop is illustrated in Figure~\ref{fig:system},
and the pseudocode is in Algorithm~\ref{alg:agent}.
At each step $t$, the LLM gets passed the previous message history (context)
$m_{t-1}$,
performs reasoning,
and then  returns  an updated belief state $b_{t-1}$
(representing a structured summary of the past) and a recommended
action $a_t$, which can be a specialized tool call (such as looking up
information in a database) or a web search.
Next the agent executes the action and gets results $o_t$,
and then it updates its message history by either
appending $(b_{t-1}, a_t, o_t)$ to $m_{t-1}$ (full-BLF)
or just appending  $(b_{t-1}, a_t, o_t)$ to the initial context $m_0$  (Markov-BLF);
\Cref{sec:markov} evaluates whether retaining older history improves accuracy,
without establishing that the summary is a sufficient statistic.
The loop runs for up to $T_{\max}=\maxSteps$ steps,
although the agent can choose to submit its prediction early
if it is confident enough.

\paragraph{Belief state.}
The linguistic belief state  is a semi-structured JSON object containing:
a probability estimate $p \in [0,1]$,
a confidence level,
key evidence for/against, and open questions,
which is
helpful for passing to the search query generation prompt.
(See \cref{app:belief-state} for more details.)
NoBel is a ReAct-like ablation that retains the full history and appends
$(a_t,o_t)$ to $m_{t-1}$, without requesting or maintaining a structured belief.
\eat{
The belief and memory comparisons apply the same arithmetic averaging
to all arms, with neither shrinkage nor calibration. Markov-BLF is therefore
not literally belief-only memory: it also retains the newest observation,
which has not yet been incorporated into $b_{t-1}$.
}

\eat{
\paragraph{What the state represents.}
The probability is the agent's current forecast, while the evidence fields
make its supporting and opposing considerations explicit. Open questions
identify information that could motivate the next tool call; confidence is a
linguistic assessment rather than a calibrated posterior variance. These
fields impose a recurring forecasting task on the agent: relate the accumulated
information to the outcome and identify what remains unresolved. They do not
supply a generative model of documents, enforce independent evidence, or
prevent repeated information from influencing successive assessments.
Thus the proposed mechanism is structured state tracking within a
reasoning-and-action loop. Its value must be established empirically against
an agent with the same tools and access to the conversation.
}

\paragraph{Tools.}
At each step, the agent selects one of several actions:
\texttt{web\_search}
(we use the  Brave search engine with date filtering;
see \cref{app:search} for a discussion),
\texttt{summarize\_results} (read and summarize retrieved pages),
\texttt{lookup\_url} (fetch a specific URL),
source-specific data tools
(e.g., \texttt{fetch\_ts\_yfinance}, \texttt{fetch\_wikipedia\_section}),
or \texttt{submit}.
The submit tool clips probabilities to $[0.02,0.98]$, while the timeout
fallback clips to $[0.05,0.95]$.
A meta-controller selects the set of tools available to the agent
on a per-question-type basis.
See \cref{app:loop}--\cref{app:policies} for the details.

\begin{algorithm}[t]
\footnotesize
\caption{\footnotesize \system and variants.
}
\label{alg:agent}
\begin{algorithmic}[1]
\REQUIRE Fixed context $c$ (instructions, $q$, cutoff $d$, prior), max steps $T$, memory mode
\STATE $m_0 \gets c$ 
\FOR{$t = 1, \ldots, T-1$}
  \STATE $(b_{t-1}, a_t) \gets \text{LLM}_{\mathrm{mode}}(m_{t-1})$
    \hfill \textit{// NoBel emits only $a_t$; $b_{t-1}=\emptyset$}
  \IF{$a_t = \texttt{submit}(\hat{p})$}
    \RETURN $\hat{p}$
  \ENDIF
  \STATE $o_t \gets \text{Env}(a_t; q, d)$
    \hfill \textit{// execute action; search APIs apply date cutoff $d$}
  \IF{$a_t$ is web search}
    \STATE $o_t \gets \text{LeakFilter}(o_t, d)$
      \hfill \textit{// 2nd-pass LLM filter}
  \ENDIF
  \STATE \textbf{if full-BLF:} $m_t \gets m_{t-1} \oplus (b_{t-1}, a_t, o_t)$
  \STATE \textbf{if Markov-BLF:} $m_t \gets m_0 \oplus (b_{t-1}, a_t, o_t)$
  \STATE \textbf{if NoBel:} $m_t \gets m_{t-1} \oplus (a_t, o_t)$  \hfill \textit{// ReAct Baseline}
\ENDFOR
\RETURN $\mathrm{LLM}_{\texttt{submit}}(m_{\mathrm{last}})$
  \hfill \textit{// reserved final call; latest available memory, all modes}
\end{algorithmic}
\end{algorithm}

\eat{
Algorithm~\ref{alg:agent} shows the pilot's common finalization: at most
$T-1$ research calls, then one reserved submit call using the latest memory.
Research can end earlier if its budget is exhausted; early submission is
allowed. The final call requests a probability, not a belief, in every mode.
Historical runs used different timeout fallbacks (\cref{app:loop}); those
are not the pilot's missing-prediction rule. Fixed instructions and remaining
budget metadata are retained in all modes.
}

\paragraph{Crowd and empirical prior.}
 For market questions, the crowd signal (market price)
  is injected into the prompt as an anchor;
  adding it substantially improves market BI.
  For dataset questions, we provide an \emph{empirical prior}
  $\pi_q$ --- the base rate for each source and question subtype.
  The empirical prior has negligible effect on BLF
  (which acquires better data via search),
  but helps the no-LLM baseline.
  See \cref{app:crowd} for details.

  \paragraph{Combining independent trials.}
We run $K$ independent tool-use trajectories and combine their final
probabilities. Arithmetic averaging, $\hat p=K^{-1}\sum_k p_k$, provides
a simple reference. We also evaluate logit averaging and a shrinkage
variant, $\hat p=\sigma(\alpha_q K^{-1}\sum_k\mathrm{logit}(p_k))$.
Here $\alpha_q=\max(f,1-c\sigma_q)$ depends on trial-logit dispersion,
with $(f,c)$ selected by cross-validation; a hierarchical Bayesian
alternative is described in \cref{app:shrinkage}.
When $\alpha_q=1$, shrinkage reduces to logit averaging; smaller values
pull the estimate toward $0.5$.
The belief and memory ablations use arithmetic averaging without calibration.
The external-agent comparisons retain the historical logit-aggregation and
hierarchical-calibration configuration (\cref{tab:default-config}).

In preliminary results (not reported here),
we experimented with the LLM-based aggregation method from 
the AIA Forecaster~\citep{AIA},
which passes the raw reasoning traces (but not the probability estimates)
from all $K$ trials to a final aggregation LLM (that can also perform its own search,
to resolve disagreements).
However, this worked worse than simple mean aggregation on our FB benchmark.

\paragraph{Post-hoc calibration.}
\label{sec:calibration}
We evaluated ordinary Platt scaling~\citep{platt1999} and a hierarchical
extension with regularized source-specific offsets (\cref{app:calibration}).
Although preliminary results (not reported here) showed
small gains on  Kimi K2.5\footnote{
When using Platt calibration,
we saw +0.72/+1.48 BI gains on FB questions with/without the crowd baseline.
When using our hierarchical calibration,
we saw +0.98/+1.80 BI gains.
(The gains are larger on questions without the crowd baseline,
since that tends to act as a very informative prior,
reducing the need for calibration.)
},
we found no statistically resolved overall BI gain for Pro in the historical
calibration comparison. This does not rule out gains on other backbones or subsets.

\eat{
Neither yields a statistically resolved
\emph{overall} BI gain for \Pro in the reported calibration ablation, and
hierarchical scaling offers no established advantage over global scaling
(\cref{tab:cal-comparison}). Benefits on particular subsets or weaker
baselines do not establish a general need for calibration. We preserve the
calibrated configurations in the historical external-agent comparison;
the belief and memory ablations and new trial-count curves use no calibration.
Calibration is not claimed as a contribution of BLF.
}
 
\newcommand{\gc}{\color{gray!40}\vrule width 0.4pt}
\section{Results}
\label{sec:results}
\label{sec:experiments}

In this section, we present our results;
see \cref{app:fb-results} for more details. External comparisons use the
historical aggregation and calibration configuration in \cref{tab:default-config}.

\paragraph{Comparison to SOTA.}
\label{sec:comparison}
\label{sec:comparison-fb}

\begin{figure}[h]
  \centering
  \includegraphics[width=0.8\textwidth]{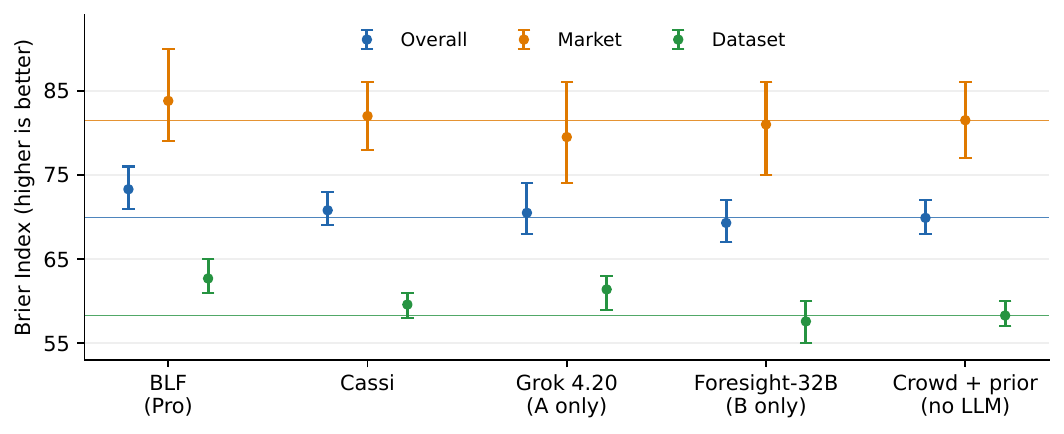}
  \caption{
    {\bf Comparison to external methods and baselines on FB A$\cup$B}.
   We plot  Brier Index,
    split by Overall, Market, and Dataset. Thin horizontal lines mark the
    corresponding crowd+prior baseline means.
    Error bars: rounded 95\% bootstrap intervals from \cref{app:tab:sota-bi}.
    See \cref{tab:mega} for pairwise comparison and significance testing.
  }
  \label{fig:sota-bi-detail}
  \label{fig:sota}
\end{figure}

\begin{table}[t]
\centering
\footnotesize
\caption{
  \small
  {\bf Comparison of BLF (Pro) to 3 external agentic methods.}
  We show Brier Index (higher = better) on FB A$\cup$B.
 All methods use the crowd
  signal for market questions.
  Absolute BI per (question type, tranche) plus paired
  $\Delta$~BI vs \system{} ($\Delta\!<\!0$ = \system{} wins).
  Tranche A: 2025-10-26; B: 2025-11-09; All = A$\cup$B.
  Grok submitted only A and Foresight only on B, so the
  unsubmitted-tranche cells are shown as $\times$.
  Sample sizes for each partition are in
  \cref{tab:mega-n}.
  Significance: $^{***}p{<}0.001$, $^{**}p{<}0.01$, $^*p{<}0.05$,
  $^\text{ns}p{\geq}0.05$ (paired bootstrap, 5{,}000 resamples).
}
\label{tab:mega}
\setlength{\tabcolsep}{3.5pt}
\newcommand{\sm}[1]{{\color{gray!60}#1}}
\begin{tabular}{@{}l !{\gc} rrrl !{\gc} rrrl !{\gc} rrrl @{}}
\toprule
 & \multicolumn{4}{c!{\gc}}{Market} & \multicolumn{4}{c!{\gc}}{Data} & \multicolumn{4}{c}{All} \\
\cmidrule(lr){2-5} \cmidrule(lr){6-9} \cmidrule(lr){10-13}
Method & A & B & All & $\Delta$ & A & B & All & $\Delta$ & A & B & All & $\Delta$ \\
\midrule
\system{} (\Pro)
  & \bi{81.3} & \bi{86.7} & \Bbi{83.8} &
  & \bi{62.4} & \bi{63.1} & \Bbi{62.7} &
  & \bi{71.9} & \bi{74.9} & \Bbi{73.3} & \\
Cassi
  & \bi{80.1} & \bi{84.2} & \bi{82.0} & $-$\dbi{1.7}$^\text{ns}$
  & \bi{61.8} & \bi{57.4} & \bi{59.6} & $-$\dbi{3.1}$^{***}$
  & \bi{71.0} & \bi{70.8} & \bi{70.8} & $-$\dbi{2.4}$^{*}$ \\
Grok
  & \bi{79.5} & $\times$    & \bi{79.5} & $-$\dbi{1.9}$^\text{ns}$
  & \bi{61.4} & $\times$     & \bi{61.4} & $-$\dbi{1.0}$^\text{ns}$
  & \bi{70.5} & $\times$     & \bi{70.5} & $-$\dbi{1.4}$^\text{ns}$ \\
Foresight
  & $\times$     & \bi{81.0} & \bi{81.0} & $-$\dbi{5.7}$^{***}$
  & $\times$     & \bi{57.6} & \bi{57.6} & $-$\dbi{5.5}$^{***}$
  & $\times$     & \bi{69.3} & \bi{69.3} & $-$\dbi{5.6}$^{***}$ \\
\bottomrule
\end{tabular}
\end{table}

\Cref{fig:sota}
compares \system{} (using \Pro) with
external methods and baselines on ForecastBench. BLF has the highest reported
point estimate for Overall, Market, and Dataset BI.
Exploratory paired bootstrap tests on
common resolution events
(shown in \cref{tab:mega})
find significant Overall and Dataset improvements over Cassi
and Foresight. Grok is not statistically distinguishable from BLF on
its tranche-A coverage ($p=0.51$), despite its lower unpaired aggregate score.
Market comparisons have less power: A$\cup$B contains 200 market events, and
Grok and Foresight each cover only one 100-event tranche. Only the market gap
against Foresight reaches significance. The table reports tranche coverage,
paired differences, and significance; \cref{app:comparison,app:mixed-effects}
give evaluation details.

\paragraph{Comparison to LLM-free baseline.}
\Cref{tab:vs-crowd-baseline} shows that
BLF is the only method (in our comparison set)
to significantly beat 
the LLM-free crowd+empirical-prior baseline in terms of overall BI.
Grok and Cassi trend positive but are not significantly above it.

\begin{table}[t]
\centering\footnotesize
\caption{
  {\bf Comparison to the LLM-free crowd+empirical-prior baseline.}
The first row gives absolute BI; subsequent rows give paired $\Delta$BI
(method minus baseline, so positive means better for method) on common coverage.
Within each type, A and B
are the tranches in \cref{tab:mega}; All pools available tranches.
Overall BI equally weights Market and Data. $\times$: no submission.
Stars: $^{***}p<0.001$, $^{**}p<0.01$, $^*p<0.05$, ns otherwise
(5,000 paired resolution-event bootstrap resamples, stratified by type;
unadjusted tests, fitted calibration held fixed).
Only BLF significantly improves Overall BI on its full coverage.}
\label{tab:vs-crowd-baseline}
\setlength{\tabcolsep}{2.5pt}
\begin{tabular}{@{}l rrr !{\gc} rrr !{\gc} rrr@{}}
\toprule
& \multicolumn{3}{c!{\gc}}{Market} & \multicolumn{3}{c!{\gc}}{Data} & \multicolumn{3}{c}{Overall}\\
\cmidrule(lr){2-4}\cmidrule(lr){5-7}\cmidrule(lr){8-10}
Method & A & B & All & A & B & All & A & B & All\\
\midrule
Crowd+prior (absolute) & 81.4 & 81.6 & 81.5 & 56.7 & 59.9 & 58.3 & 69.0 & 70.8 & 69.9 \\
\midrule
BLF (Pro) & $-0.1^{\mathrm{ns}}$ & $+5.1^{\mathrm{ns}}$ & $+2.3^{\mathrm{ns}}$ & $+5.7^{***}$ & $+3.2^{*}$ & $+4.5^{***}$ & $+2.8^{\mathrm{ns}}$ & $+4.1^{**}$ & $+3.4^{**}$ \\
Cassi & $-1.3^{\mathrm{ns}}$ & $+2.7^{\mathrm{ns}}$ & $+0.6^{\mathrm{ns}}$ & $+5.2^{***}$ & $-2.5^{*}$ & $+1.3^{\mathrm{ns}}$ & $+1.9^{\mathrm{ns}}$ & $+0.1^{\mathrm{ns}}$ & $+1.0^{\mathrm{ns}}$ \\
Grok & $-1.9^{\mathrm{ns}}$ & $\times$ & $-1.9^{\mathrm{ns}}$ & $+4.7^{***}$ & $\times$ & $+4.7^{***}$ & $+1.4^{\mathrm{ns}}$ & $\times$ & $+1.4^{\mathrm{ns}}$ \\
Foresight & $\times$ & $-0.6^{\mathrm{ns}}$ & $-0.6^{\mathrm{ns}}$ & $\times$ & $-2.3^{\mathrm{ns}}$ & $-2.3^{\mathrm{ns}}$ & $\times$ & $-1.5^{\mathrm{ns}}$ & $-1.5^{\mathrm{ns}}$ \\
\bottomrule

\end{tabular}
\end{table}

\paragraph{Benefits of the belief state.}
\label{sec:component-effects}
\Cref{tab:component-effects} isolates the benefits of the belief state.
The first row for each LLM backbone reports absolute NoBel BI; the next reports
the paired effect of adding the structured belief state.
\Pro and \Flash improve
overall, driven mainly by market questions; \GPT has no resolved overall
gain and a small negative dataset effect.\footnote{
Earlier experiments showed large gains on Kimi-K2.5, but its market
submission rate was only 6.2\% without belief tracking versus 82.4\% with it.
Failed NoBel runs defaulted to 0.5, while belief agents retained an
intermediate forecast. We omit these results because they confound
forecast quality with recovery from submission failures.
}
We do not know exactly why the belief state helps with \Pro and \Flash,
but not with \GPT.
However, we did notice that
the mean within-question cross-trial standard deviation
on market forecasts falls for \Pro
(0.094 to 0.030) and \Flash (0.105 to 0.032), but rises slightly for
\GPT, suggesting some modest  stabilizing effect
of a structured context.
However, 
it is also possible that the structured representation
helps formulate better search queries;
unfortunately it is hard to test for this
in our current experimental setup
(for engineering reasons).

\eat{
\paragraph{Why does belief tracking help some backbones?}
The market concentration is consistent with a role in organizing sequential
research, but does not establish that mechanism. Saved traces show different
responses to the same structure: Pro's median loop shortens from nine to seven
steps, whereas GPT-5 remains at ten
(\cref{app:backbone-diagnostic}).
The mean within-question trial SD on market forecasts also falls for Pro
(0.094 to 0.030) and Flash (0.105 to 0.032), but rises slightly for
GPT-5 (\cref{app:backbone-diagnostic}). This is consistent with stabilizing
otherwise variable forecasts, not proof of better evidence integration.
A model may already perform useful state tracking internally, or an imposed
schema may add overhead; both remain hypotheses. Model size alone cannot be
isolated from training, prompting, and tool-use behavior in this comparison.
We therefore view structured beliefs as a backbone-dependent agent design,
not a universally necessary representation. Fresh-backbone replication and
controls that separate probability elicitation from evidence summarization
are needed to establish transfer and mechanism.
}

\begin{table}[t]
\centering\small
\caption{{\bf Belief state ablation.}
  We report NoBel baseline BI, and full-BLF minus NoBel
$\Delta$BI (higher is better), crowd prior, five trials, arithmetic averaging,
no shrinkage or calibration. Pro/GPT: 400 question--date instances; Flash: 399.
Brackets: exploratory 95\% question-cluster intervals, stratified by type;
bold differences exclude zero. All equally weights Market and Data.
See \cref{sec:component-effects}; intervals use 5,000 paired resamples.
}
\label{tab:component-effects}
\setlength{\tabcolsep}{3pt}
\begin{tabular}{llrrr}
\toprule
Model & Reference / comparison & Market & Data & All\\
\midrule
\Pro & NoBel (absolute BI) & 78.87 & 61.45 & 70.16\\
 & + Belief & $\mathbf{+5.34}\;[3.32,7.72]$ & $\mathbf{+0.67}\;[0.06,1.31]$ & $\mathbf{+3.00}\;[1.95,4.25]$\\
\addlinespace
\Flash & NoBel (absolute BI) & 79.07 & 61.44 & 70.26\\
 & + Belief & $\mathbf{+4.06}\;[2.38,6.02]$ & $-0.07\;[-0.65,0.50]$ & $\mathbf{+1.99}\;[1.12,3.02]$\\
\addlinespace
\GPT & NoBel (absolute BI) & 80.41 & 61.96 & 71.18\\
 & + Belief & $-0.62\;[-1.49,0.21]$ & $\mathbf{-0.22}\;[-0.39,-0.05]$ & $-0.42\;[-0.86,0.002]$\\
\bottomrule

\end{tabular}
\end{table}

\begin{figure}[t]
\centering
\includegraphics[width=0.8\linewidth]{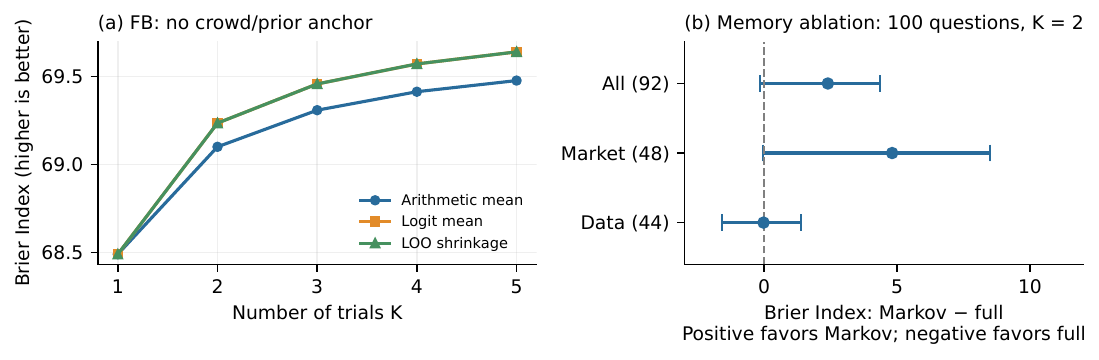}
\caption{
{\bf Effect of multi-trial averaging and Markovian memory.}
  (a) Trial averaging without crowd/prior anchors: 400 ForecastBench
questions, 791 events. Expected loss over trial subsets is converted to BI;
All equally weights Market/Data. Shrinkage uses leave-one-question-out tuning
and coincides with logit averaging.
(b) Separate 100-question memory study, $K=2$.
We show the difference in Brier Index
for BLF-Markov minus BLF-Full,
together with  95\% paired question-cluster intervals.
Counts show complete pairs; positive favors Markov.
All equally weights
Market/Data; every interval includes zero.}
\label{fig:aggregation-main}
\end{figure}

\paragraph{Benefits of averaging multiple trials.}
\label{sec:aggregation}
By default we use arithmetic averaging of the probabilities
from $K=5$ trials per question.
In \cref{fig:aggregation-main}(a)
we show the effect of varying $K$
on pre-computed BLF(\Pro) trajectories collected
without crowd forecasts or empirical-prior anchors
on ForecastBench.
We see  gains with increasing $K$ when using the arithmetic  mean,
and a small boost when using the logit mean.
When using leave-one-out cross-validation,
the shrinkage method selects a shrinkage
coefficient of $\alpha_q=1$, which reduces to standard logit averaging.

\paragraph{Full-history versus Markov memory.}
\label{sec:markov}
\Cref{fig:aggregation-main}(b) compares full BLF,
$m_t\gets m_{t-1}\oplus(b_{t-1},a_t,o_t)$, with Markov-BLF,
$m_t\gets m_0\oplus(b_{t-1},a_t,o_t)$, on 100 fresh questions
(50 Market, 50 Data), selected independently of outcomes.
Both arms use Pro, $K=2$,
at most ten calls including reserved final submission,
and identical tools, including date-filtered search and URL lookup,
but no lookup by index into previously stored documents
(which could violate the Markov property).
Full BLF submits in 191/200 rollouts versus 199/200 for
Markov; 92 questions (48 Market, 44 Data) have both trials complete in both arms.
After averaging the two forecasts, overall BI is 72.72 for full BLF
and 75.12 for Markov: $\Delta\mathrm{BI}=+2.40$
(95\% CI $[-0.17,4.36]$; 10,000 paired question-cluster resamples,
stratified by type). The point estimate favors Markov, mainly on markets,
but every interval includes zero.

\section{Example of a  belief state trace}
\label{app:aibq2-qualitative}
\label{app:aibq2-beliefs}
\label{sec:belief trace}

\begin{figure}[t]
  \centering
  \includegraphics[width=0.7\textwidth]{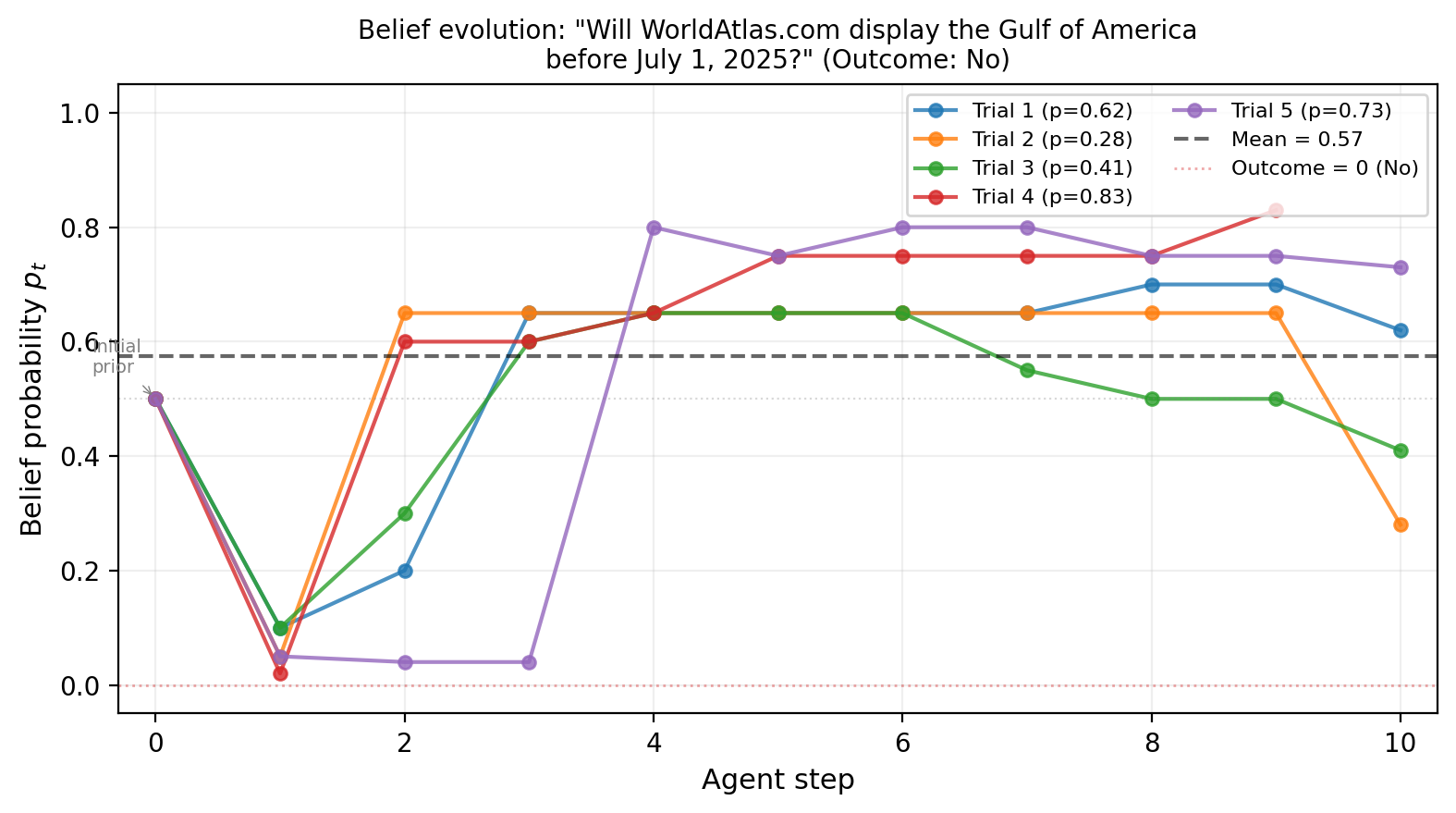}
  \caption{
    \small
        {\bf Belief evolution across 5 trials.}
        The question is ``Will WorldAtlas.com
    display the Gulf of America before July 1, 2025?'' (outcome: No).
    All trials start at $p_0=0.5$ and dip low at steps 1--2
    (initial search finds skepticism), then diverge as different
    search paths lead to different conclusions.
      High inter-trial variance ($\sigma{=}0.20$) motivates aggregation.
  }
  \label{fig:belief-trace}
\end{figure}

\paragraph{Belief evolution.}
In this section, we perform a qualitative analysis of the behavior of the 
\system{} agent using \Pro.
Figure~\ref{fig:belief-trace} shows how the agent's probability
estimate evolves across steps for 5 independent trials
on a single  question.\footnote{
The question is from
the Metaculus AI Benchmark Tournament (Q2 2025),
available at  \url{https://www.metaculus.com/tournament/aibq2/}.
The question
is as follows:
''Will WorldAtlas.com
    display the Gulf of America before July 1, 2025?''
    The context is that President Trump had ordered that the 
    ``Gulf of Mexico'' be renamed to ``Gulf of America''
    on January 20, 2025.
      See
      \url{ https://en.wikipedia.org/wiki/Executive_Order_14172}
      for context.
}
All trials start at $p_0=0.5$ (no information) but quickly diverge
as different search queries lead to different evidence.
The high inter-trial variance ($\sigma=0.20$) motivates multi-trial
aggregation as a way to reduce reliance on any one trajectory;
the mean forecast is 0.57.
The trace from one of the rollouts (Figure~\ref{fig:agent-trace})
shows how the belief state captures key evidence:
Trial~2 correctly identifies that WorldAtlas uses
static maps (not dynamic Google Maps), leading 
this agent to become skeptical that the change will happen before the deadline.
(And indeed the true outcome is ``No''.)

\begin{figure}[t]
  \centering
  \footnotesize
  \fbox{\parbox{0.85\textwidth}{\footnotesize
  \textbf{Question:} Will WorldAtlas.com display the Gulf of America
  before July 1, 2025?\\[1pt]
  \textbf{Trial 2} (forecast: 0.28) --- the ``skeptic'' among 5 trials
  \hfill \mbox{\textbf{Outcome:} No}\\[1pt]
  \renewcommand{\arraystretch}{0.92}
  \begin{tabular}{@{}rp{0.12\linewidth}p{0.74\linewidth}@{}}
  \textbf{Step} & $p_t$ & \textbf{Action \& key evidence} \\[1pt]
  0 & 0.50 & Prior (no information) \\
  1 & 0.05 & \texttt{search}(``Gulf of America Gulf of Mexico'')\\
    & & $\downarrow$ ``Renaming attempts usually fail; base rate $<$1\%'' \\
  2 & 0.65 & \texttt{lookup\_url}(worldatlas.com/\ldots/na.htm)\\
    & & $\uparrow$ Executive Order signed; Google Maps changed label \\
  3--8 & 0.65 & Multiple searches --- inconclusive \\
  9 & 0.65 & \texttt{search}(``worldatlas.com maps API'')\\
    & & $\downarrow$ WorldAtlas uses \emph{static} map images \\
  10 & \textbf{0.28} & \texttt{submit} --- ``Static sites take 1--3 years to update'' \\
  \end{tabular}
  }}
  \caption{
    \footnotesize
    {\bf Agent trace for Trial~2 (Figure~\ref{fig:belief-trace}).}
    The agent initially drops to $p=0.05$ (skeptical prior),
    jumps to 0.65 after finding the executive order (from President Trump),
    then stays at 0.65 for 7 steps while searching inconclusively.
    The static-map concern is incorporated into the final forecast at
    the step-10 submission, when the probability drops to $p=0.28$.
    This trial correctly identifies the key uncertainty ---
    whether the site has updated its static maps --- that the
    other four trials miss.
  }
  \label{fig:agent-trace}
\end{figure}

\paragraph{The Trial~2 ``aha'' moment.}
The drop $0.65\to 0.28$ at Trial~2's submit step 
(\cref{fig:agent-trace}) is  not a
new search hit --- it is the LLM \emph{finally synthesizing} two
pre-existing latent concerns (the \texttt{webimage/countrys/} URL
structure looks like a legacy static page, and WorldAtlas relies on
proprietary static map images that require manual editing) into a new
\texttt{evidence\_against} entry and corresponding
\texttt{key\_uncertainty}. This crystallization happens inside the
final reasoning pass, after the agent decides it has enough material
to commit; it overrides the optimistic priors driven by Trump's EO
and WorldAtlas's other article-title update, and the belief drops by
$0.37$ in a single step. This is precisely the regime that Trial~2
captures correctly and the other four trials miss; the mean does not beat the best individual trial on this example.

\section{Explicit Bayesian Updating}
\label{sec:explicit}
\label{sec:explicit-bayes}

The approach described in \cref{sec:system} is heuristic,
and uses an LLM to compute the probability at each step.
In this section, we describe a more ``orthodox'' approach,
where we update the numerical beliefs with Bayes' rule,
using an LLM to estimate the likelihood or likelihood ratio,
as suggested in  \citep{Papamarkou2026,Amin2026}.

\paragraph{The explicit update.}
We will use this notation:
$s \in \{0,1\}$ is the latent (unknown) outcome to be predicted,
$a_t$ is the search query (action),
$o_t$ the resulting text response, i.e., the observations (concatenated together),
$c$ is the context (prompt containing the question),
and
$b_t=p(s=1|a_{1:t},o_{1:t},c)$
is the belief state at step $t$.
The sequential update for the posterior in log-odds form is
\begin{equation}
  \operatorname{logit} b_T
  \;=\; \operatorname{logit} b_0
        \;+\; \alpha \sum_{t=1}^{T} \lambda_t,
  \qquad
  \lambda_t \;=\; \log \frac{q_t(1)}{q_t(0)}
  \label{eqn:explicit-bayes}
\end{equation}
where each log likelihood  term
$\lambda_t$ depends on $q_t(s)$,
which we discuss below.
Here $\alpha$ is a tempering coefficient we discuss below.

\paragraph{Likelihoods.}
In the simplest case, we have 
$q_t(s)=p(o_t\mid s)$, so $\lambda_t$ is a standard log-likelihood ratio,
but we can also condition on other quantities.
In  \cref{tab:explicit-bayes} we list many possibilities.
Here $h_{t-1}=(a_{1:t-1},o_{1:t-1})$ denotes the raw history, and
$b_{t-1}.h$ denotes the structured belief's evidence summary with its
probability removed, to avoid feeding the running posterior into the likelihood
estimate. Conditioning on $a_t$ accounts for the issued query;
conditioning on $h_{t-1}$ accounts for all the past evidence;
conditioning on
$b_{t-1}.h$ aims to just condition on evidence in the structured belief state.
An exact
chain-rule update would require full-history conditioning or a sufficient
summary and correctly specified likelihoods; neither is guaranteed here.

\paragraph{Eliciting the likelihood.}
To estimate 
the quantities $q_t(s)$,
we follow the procedure in \citet[Eq.~61]{Amin2026}: for each hypothesis
$s\in\{0,1\}$ we ask the LLM (\Flash) to rate,  on an integer
scale $0$--$10$, how \emph{typical} the observations $o_t$
look given the conditioning context $s$ plus relevant history.
(See \cref{app:likelihoodPrompts} for the prompt.)
A typicality judgment is more natural for an LLM than a probability, and
querying each state in a separate call avoids the implicit discriminative
normalization of a single classification prompt.
(A more direct alternative that asks the model for the
log-ratio $\log p(o_t\mid s{=}1)/p(o_t\mid s{=}0)$ in a single call produces
inflated, frequently wrong-signed weights and was uniformly worse.)
Writing $r_t(s)\in[0,10]$ for
the two scores, we define
\begin{equation}
  \lambda_t \;=\; \log \frac{r_t(1)+\epsilon}{r_t(0)+\epsilon},
  \qquad \epsilon = 0.5,
  \label{eqn:per-state}
\end{equation}
where the floor $\epsilon$ keeps a zero score from producing an infinite weight.
The bounded integer scale also caps the magnitude of each $\lambda_t$.
These typicality scores are likelihood surrogates, not normalized text
likelihoods.

\paragraph{Summarizing text observations.}
The raw observation $o_t$ --- the concatenated text of one search step --- averages
$7{,}571$ characters.
So  we also test
a compressed, \emph{task-aware} observation: a single auxiliary LLM call, given the
question $c$ and the issued query $a_t$, rewrites $o_t$ into a short summary,
$  \sigma_t \;=\; \sigma(o_t \mid a_t, c)$,
of mean length $798$ characters ($11\%$ of $o_t$); conditioning the summary on
$a_t$ and $c$ keeps what is relevant to the question. The per-state likelihood
is then evaluated on $\sigma_t$ in place of $o_t$.

\paragraph{Tempering.}
The  $\alpha$ term is
a \emph{tempering} coefficient,
the same as the composite-likelihood
exponent of \citet[Eq.~1]{Papamarkou2026}.
This can be useful because  with conditionally dependent observations,
the naive product over-counts evidence.
If we set $\alpha<1$, it down-weights each likelihood factor
to restore approximately calibrated precision (a Gibbs / power-posterior
correction).
We sweep $\alpha$ on the replay sample and report its best value;
these exploratory results are not held-out validation estimates.

\eat{
We sweep $\alpha$ on the replay sample and report its best value;
these exploratory values are not held-out validation estimates.
We could in principle get better results using
tool-specific (or action-specific) values for $\alpha$,
but we did not try this.
Another mitigation for correlated observations
listed by \citet{Papamarkou2026}, \emph{dependence-aware
evidence pooling} (cluster documents by underlying claim, one $\lambda$ per
cluster), we do not evaluate separately, since conditioning on the running
summary (\bcond, below) already addresses the over-counting it targets.
}

\paragraph{Results.}

\begin{table}[h]
\centering
\footnotesize
\caption{
  {\bf Explicit-Bayes belief update vs the linguistic update (\BLF)}.
We replay all methods on the same evidence sequence. BS first averages
horizons within each question, then averages the 80 questions; BI transforms
this mean, while ECE weights resolution events equally.
Higher BI $=100(1-\sqrt{\mathrm{BS}})$ and lower ECE are better; $\alpha^\star$ is the best tempering
  coefficient; both metrics are evaluated at $\alpha^\star$.
  The top block uses the raw observation $o_t$,
the bottom the task-aware summary $\sigma_t$.
All quantities are implicitly conditioned on the question context $c$
(omitted for brevity).
}
\label{tab:explicit-bayes}
\setlength{\tabcolsep}{4pt}
\renewcommand{\arraystretch}{0.92}
\begin{tabular}{llccc}
\toprule
Method & $q_t(s)$ & $\alpha^\star$ & BI & ECE \\
\midrule
\BLF (linguistic) & --- & --- & \textbf{70.38} & \textbf{0.083} \\
Prior (market / $0.5$) & --- & 0.00 & 63.08 & 0.117 \\
\midrule
\uncond & $p(o_t\mid s)$ & 0.05 & 63.11 & 0.114 \\
\acond & $p(o_t\mid s,a_t)$ & 0.05 & 63.13 & 0.112 \\
\bcond & $p(o_t\mid s,a_t,b_{t-1}.h)$ & 0.35 & 64.93 & 0.097 \\
\hcond & $p(o_t\mid s,a_t,h_{t-1})$ & 0.00 & 63.08 & 0.117 \\
\hbcond & $p(o_t\mid s,a_t,b_{t-1}.h,h_{t-1})$ & 0.20 & 64.29 & 0.093 \\
\midrule
\sabcond & $p(\sigma_t\mid s,a_t,b_{t-1}.h)$ & 0.35 & 65.13 & 0.107 \\
$\sigma\text{-}\mathrm{sa}\sigma$ & $p(\sigma_t\mid s,a_t,\sigma_{1:t-1})$ & 0.20 & 65.18 & 0.126 \\
\sahbcond & $p(\sigma_t\mid s,a_t,b_{t-1}.h,\sigma_{1:t-1})$ & 0.35 & 65.73 & 0.103 \\
\bottomrule
\end{tabular}
\end{table}
At the shared $\alpha=0.35$, removing the belief summary increases BS by
$0.0057$ (95\% paired question-bootstrap CI $[-0.0122,0.0290]$);
its incremental benefit is unresolved. Alpha was selected on this sample,
and the old and new likelihood scores were elicited on different dates.

Some results, using
$n=80$ tranche-A questions ($154$ events),
are shown in \cref{tab:explicit-bayes}.
\eat{
The prior-only baseline (corresponding to \alpha=0)
gets a Brier Score of 
BS = 0.1363,
and expected calibration error (ECE) score
of ECE = 0.1170.
}
Of all the likelihood-based approaches,
the best one (in terms of BI)
uses $q_t(s) = p(\sigma_t\mid s,a_t,b_{t-1}.h,\sigma_{1:t-1})$,
which conditions on the belief state plus the past (summarized)
evidence. This is analogous to the non-Markovian version of BLF,
and gets BI=65.73.
Interestingly, this is slightly better than just
conditioning on the past without the belief state,
$q_t(s) = p(\sigma_t\mid s,a_t,\sigma_{1:t-1})$,
which gets BI=65.18,
or just conditioning on the belief state alone,
$q_t(s) = p(\sigma_t\mid s,a_t,b_{t-1}.h)$,
which gets BI=65.13.
However, all of these methods are worse than BLF,
which gets BI=70.38, and ECE=0.083.

\section[Related work]{Related work\protect\footnote{For more related work, see \cref{app:related}.}}
\label{sec:related}

Retrieval-augmented forecasting systems combine search, reasoning, aggregation,
and calibration~\citep{halawi2024approaching,Hsieh2024,AIA}.
BLF follows the sequential reasoning-and-action pattern of ReAct~\citep{ReAct};
its belief representation is a structured running summary with a probability,
not a new exact Bayesian inference algorithm. Our contribution is
a careful empirical study of this representation. 
BayesAgent~\citep{BayesAgent} instead constructs explicit graphical models;
likelihood-based proposals~\citep{Papamarkou2026,Amin2026} motivate our negative
results in \cref{sec:explicit-bayes}.
Model training for forecasting~\citep{Turtel2026,Chandak2026,scott2026forecasting}
is complementary to these changes to the agent harness.

\section{Conclusion}
\label{sec:conclusion}

We studied structured linguistic beliefs as
a form of context engineering for search-enabled forecasting agents.
BLF has the highest overall BI point estimate among the compared external
agents on \nqAB{} ForecastBench question-date instances.

\paragraph{Limitations and future work.}
Our evaluation is restricted to binary outcomes and backtesting; it does not
establish live forecasting performance or complete cost-normalized superiority.
Cutoff filters reduce, but cannot eliminate, leakage risk.
The
memory comparison is exploratory and excludes incomplete paired submissions.
Future work should test non-binary outcomes,
improve numerical tools, train models for calibration and evidence integration,
and validate the system in live forecasting.


\clearpage
\subsection*{AI use statement}

In this work, we used generative AI tools to help write the code,
launch and analyse experiments, and help with writing and figure preparation.

\subsection*{Ethics}

Forecasting systems can have
sensitive uses, including geopolitical forecasting,
so we recommend caution and human oversight.

\subsection*{Reproducibility statement}

Code and data will be released on GitHub upon publication.

\bibliographystyle{arxiv_blf}
\bibliography{references}

\begin{thebibliography}{60}
\providecommand{\natexlab}[1]{#1}
\providecommand{\url}[1]{\texttt{#1}}
\expandafter\ifx\csname urlstyle\endcsname\relax
  \providecommand{\doi}[1]{doi: #1}\else
  \providecommand{\doi}{doi: \begingroup \urlstyle{rm}\Url}\fi

\bibitem[Agrawal et~al.(2026)Agrawal, Dutta, Hasan, Karmaker, and
  Dutta]{FinTradeBench}
Yogesh Agrawal, Aniruddha Dutta, Md~Mahadi Hasan, Santu Karmaker, and Aritra
  Dutta.
\newblock {FinTradeBench}: A financial reasoning benchmark for {LLMs}.
\newblock \emph{arxiv}, 2026.
\newblock URL \url{https://arxiv.org/abs/2603.19225}.

\bibitem[Ahamed et~al.(2026)Ahamed, Parmar, Goyal, Song, Le, Cheng, Li,
  Palangi, Yoon, and Pfister]{TFRBench}
Md~Atik Ahamed, Mihir Parmar, Palash Goyal, Yiwen Song, Long~T. Le, Qiang
  Cheng, Chun-Liang Li, Hamid Palangi, Jinsung Yoon, and Tomas Pfister.
\newblock {TFRBench}: A reasoning benchmark for evaluating forecasting systems.
\newblock \emph{arxiv}, 2026.
\newblock URL \url{https://arxiv.org/abs/2604.05364}.

\bibitem[Aitchison et~al.(2026)Aitchison, Jeen, Shevlane, and
  Day]{Aitchison2026ensemble}
Matthew Aitchison, Scott Jeen, Toby Shevlane, and Ben Day.
\newblock Diversity is the strength of the {AI} crowd.
\newblock In \emph{ICML Workshop on Forecasting}, June 2026.
\newblock URL \url{http://dx.doi.org/10.48550/arXiv.2606.29661}.

\bibitem[Alur et~al.(2025)Alur, Stadie, Kang, Chen, McManus, Rickert, Lee,
  Federici, Zhu, Fogerty, Williamson, Lozinski, Linsky, and Sekhon]{AIA}
Rohan Alur, Bradly~C. Stadie, Daniel Kang, Ryan Chen, Matt McManus, Michael
  Rickert, Tyler Lee, Michael Federici, Richard Zhu, Dennis Fogerty, Hayley
  Williamson, Nina Lozinski, Aaron Linsky, and Jasjeet~S. Sekhon.
\newblock {AIA} forecaster: Technical report.
\newblock \emph{arxiv}, 2025.
\newblock URL \url{https://arxiv.org/abs/2511.07678}.

\bibitem[Amin(2026)]{Amin2026}
Danial Amin.
\newblock Bayesian orchestration of multi-{LLM} agents for cost-aware
  sequential decision-making.
\newblock \emph{arXiv [cs.AI]}, January 2026.
\newblock URL \url{http://dx.doi.org/10.48550/arXiv.2601.01522}.

\bibitem[Anthropic(2025)]{anthropic-search}
Anthropic.
\newblock Web search tool.
\newblock
  \url{https://docs.anthropic.com/en/docs/agents-and-tools/tool-use/web-search-tool},
  2025.

\bibitem[Auzina et~al.(2026)Auzina, Strüber, Hernández-Gutiérrez, Goel,
  Prabhu, and Bethge]{Auzina2026}
Ilze~Amanda Auzina, Joschka Strüber, Sergio Hernández-Gutiérrez, Shashwat
  Goel, Ameya Prabhu, and Matthias Bethge.
\newblock Intrinsic credit assignment for long horizon interaction.
\newblock \emph{arXiv [cs.LG]}, February 2026.
\newblock URL \url{http://dx.doi.org/10.48550/arXiv.2602.12342}.

\bibitem[Bastani et~al.(2025)Bastani, Kucinskas, and Karger]{Bastani2025}
Houtan Bastani, Simas Kucinskas, and Ezra Karger.
\newblock How well can large language models predict the future?, 2025.
\newblock URL
  \url{https://forecastingresearch.substack.com/p/ai-llm-forecasting-model-forecastbench-benchmark}.

\bibitem[Breunig(2025)]{Breunig2025}
Drew Breunig.
\newblock How long contexts fail, June 2025.
\newblock URL
  \url{https://www.dbreunig.com/2025/06/22/how-contexts-fail-and-how-to-fix-them.html}.

\bibitem[{Cassi AI}(2025)]{Cassi}
{Cassi AI}.
\newblock Cassi: {AI}-powered forecasting.
\newblock \url{https://cassi-ai.com/}, 2025.

\bibitem[Chandak et~al.(2026)Chandak, Goel, Prabhu, Hardt, and
  Geiping]{Chandak2026}
Nikhil Chandak, Shashwat Goel, Ameya Prabhu, Moritz Hardt, and Jonas Geiping.
\newblock Scaling open-ended reasoning to predict the future.
\newblock \emph{arxiv}, 2026.
\newblock URL \url{https://arxiv.org/abs/2512.25070}.

\bibitem[Cheng et~al.(2026)Cheng, Liu, and Long]{PolyBench}
Pu~Cheng, Juncheng Liu, and Yunshen Long.
\newblock {PolyBench}: Benchmarking {LLM} forecasting and trading capabilities
  on live prediction market data.
\newblock \emph{arxiv}, 2026.
\newblock URL \url{https://arxiv.org/abs/2604.14199}.

\bibitem[Damani et~al.(2026)Damani, Puri, Slocum, Shenfeld, Choshen, Kim, and
  Andreas]{Damani2026}
Mehul Damani, Isha Puri, Stewart Slocum, Idan Shenfeld, Leshem Choshen, Yoon
  Kim, and Jacob Andreas.
\newblock Beyond binary rewards: Training {LMs} to reason about their
  uncertainty.
\newblock In \emph{ICLR}, 2026.
\newblock URL \url{https://openreview.net/forum?id=ASQ649zdHm}.

\bibitem[Efron \& Morris(1973)Efron and Morris]{efron1973}
Bradley Efron and Carl Morris.
\newblock Stein's estimation rule and its competitors---an empirical {B}ayes
  approach.
\newblock \emph{Journal of the American Statistical Association}, 68\penalty0
  (341):\penalty0 117--130, 1973.

\bibitem[Falck et~al.(2024)Falck, Wang, and Holmes]{Falck2024}
Fabian Falck, Ziyu Wang, and Chris Holmes.
\newblock Is in-context learning in large language models bayesian? a
  martingale perspective.
\newblock In \emph{ICML}, June 2024.
\newblock URL \url{https://arxiv.org/abs/2406.00793}.

\bibitem[Frick et~al.(2025)Frick, Chen, Tennyson, Li, Chiang, Angelopoulos, and
  Stoica]{promptToLeaderboard}
Evan Frick, Connor Chen, Joseph Tennyson, Tianle Li, Wei-Lin Chiang,
  Anastasios~N. Angelopoulos, and Ion Stoica.
\newblock Prompt-to-leaderboard: Prompt-adaptive {LLM} evaluations, 2025.
\newblock URL \url{https://arxiv.org/abs/2502.14855}.

\bibitem[Gneiting \& Raftery(2007)Gneiting and Raftery]{Gneiting2007}
Tilmann Gneiting and Adrian~E. Raftery.
\newblock Strictly proper scoring rules, prediction, and estimation.
\newblock \emph{{J. Amer. Statist. Assoc}}, 102:\penalty0 359--378, 2007.

\bibitem[Goel et~al.(2026)Goel, Chandak, Arun, Prabhu, Staab, Hardt,
  Andriushchenko, and Geiping]{Goel2026}
Shashwat Goel, Nikhil Chandak, Arvindh Arun, Ameya Prabhu, Steffen Staab,
  Moritz Hardt, Maksym Andriushchenko, and Jonas Geiping.
\newblock {FutureSim}: Replaying world events to evaluate adaptive agents.
\newblock \emph{arXiv [cs.LG]}, May 2026.
\newblock URL \url{http://dx.doi.org/10.48550/arXiv.2605.15188}.

\bibitem[Google(2025)]{gemini-grounding}
Google.
\newblock Grounding with {Google} search.
\newblock \url{https://ai.google.dev/gemini-api/docs/grounding}, 2025.

\bibitem[Guan et~al.(2024)Guan, Peng, Wang, Hou, and Li]{OpenEP}
Yong Guan, Hao Peng, Xiaozhi Wang, Lei Hou, and Juanzi Li.
\newblock {OpenEP}: Open-ended future event prediction.
\newblock \emph{arxiv}, 2024.
\newblock URL \url{https://arxiv.org/abs/2408.06578}.

\bibitem[Halawi et~al.(2024)Halawi, Zhang, Yueh-Han, and
  Steinhardt]{halawi2024approaching}
Danny Halawi, Fred Zhang, Chen Yueh-Han, and Jacob Steinhardt.
\newblock Approaching human-level forecasting with language models.
\newblock \emph{arxiv}, 2024.
\newblock URL \url{https://arxiv.org/abs/2402.18563}.

\bibitem[Hsieh et~al.(2024)Hsieh, Fu, and Chen]{Hsieh2024}
Elvis Hsieh, Preston Fu, and Jonathan Chen.
\newblock Reasoning and tools for human-level forecasting.
\newblock \emph{arxiv}, 2024.
\newblock URL \url{https://arxiv.org/abs/2408.12036}.

\bibitem[Huang et~al.(2026)Huang, Shen, Wang, Meng, Liu, Duchene, Wang, and
  Bhatt]{BayesAgent}
Hengguan Huang, Xing Shen, Songtao Wang, Lingfa Meng, Dianbo Liu,
  David~Alejandro Duchene, Hao Wang, and Samir Bhatt.
\newblock {BayesAgent}: {Bayesian} agentic reasoning under uncertainty via
  verbalized probabilistic graphical modeling.
\newblock In \emph{AAAI}, 2026.
\newblock URL \url{https://arxiv.org/abs/2406.05516}.

\bibitem[Jeen et~al.(2026{\natexlab{a}})Jeen, Aitchison, Clark, Shevlane, and
  Day]{Jeen2026}
Scott Jeen, Matthew Aitchison, Maximilian Anthony~Hugh Clark, Toby Shevlane,
  and Ben Day.
\newblock Reaching the frontier of {AI} forecasting with reinforcement
  learning.
\newblock In \emph{{ICML} Workshop on Forecasting as a New Frontier of
  Intelligence}, June 2026{\natexlab{a}}.
\newblock URL \url{https://openreview.net/forum?id=lbpDR9pj5F}.

\bibitem[Jeen et~al.(2026{\natexlab{b}})Jeen, Aitchison, and
  {Mantic}]{scott2026forecasting}
Scott Jeen, Matthew Aitchison, and {Mantic}.
\newblock Training {LLMs} to predict world events.
\newblock \emph{Thinking Machines Lab: News}, 2026{\natexlab{b}}.
\newblock URL
  \url{https://thinkingmachines.ai/news/training-llms-to-predict-world-events/}.

\bibitem[Karger et~al.(2025)Karger, Bastani, Yueh-Han, Jacobs, Halawi, Zhang,
  and Tetlock]{forecastbench}
Ezra Karger, Houtan Bastani, Chen Yueh-Han, Zachary Jacobs, Danny Halawi, Fred
  Zhang, and Philip Tetlock.
\newblock {ForecastBench}: A dynamic benchmark of {AI} forecasting
  capabilities.
\newblock In \emph{ICLR}, 2025.

\bibitem[Karkar \& Chopra(2025)Karkar and Chopra]{Karkar2025}
Chinmay Karkar and Paras Chopra.
\newblock Future is unevenly distributed: Forecasting ability of {LLMs} depends
  on what we're asking.
\newblock \emph{arxiv}, 2025.
\newblock URL \url{https://arxiv.org/abs/2511.18394}.

\bibitem[Kucinskas et~al.(2025)Kucinskas, Bastani, and
  Karger]{forecastbenchUpdate}
Simas Kucinskas, Houtan Bastani, and Ezra Karger.
\newblock {ForecastBench}: An updated ranking methodology, 2025.
\newblock URL
  \url{https://forecastbench.org/assets/pdfs/forecastbench_updated_methodology.pdf}.

\bibitem[Kucinskas et~al.(2026)Kucinskas, Bastani, and Reynolds]{BrierIndex}
Simas Kucinskas, Houtan Bastani, and Matt Reynolds.
\newblock Making forecasting scores easier to interpret: Introducing the brier
  index, 2026.
\newblock URL
  \url{https://forecastingresearch.substack.com/p/introducing-the-brier-index}.

\bibitem[Lawrence et~al.(2006)Lawrence, Goodwin, O'Connor, and
  \"{O}nkal]{lawrence2006}
Michael Lawrence, Paul Goodwin, Marcus O'Connor, and Dilek \"{O}nkal.
\newblock Judgmental forecasting: A review of progress over the last 25 years.
\newblock \emph{International Journal of Forecasting}, 22\penalty0
  (3):\penalty0 493--518, 2006.

\bibitem[Lee et~al.(2026)Lee, Nair, Zhang, Lee, Khattab, and Finn]{metaHarness}
Yoonho Lee, Roshen Nair, Qizheng Zhang, Kangwook Lee, Omar Khattab, and Chelsea
  Finn.
\newblock {Meta-Harness: End-to-End Optimization of Model Harnesses}.
\newblock \emph{arxiv}, 2026.
\newblock URL \url{https://arxiv.org/abs/2603.28052}.

\bibitem[Li(2026)]{Li2026probes}
Bojie Li.
\newblock {Incompressible Knowledge Probes: Estimating Black-Box LLM Parameter
  Counts via Factual Capacity}.
\newblock \emph{arXiv preprint arXiv:2604.24827}, 2026.

\bibitem[Li et~al.(2026)Li, Wang, Lahib, Xia, and Pi]{SimulatedIgnorance}
Zehan Li, Yuxuan Wang, Ali~El Lahib, Ying-Jieh Xia, and Xinyu Pi.
\newblock Simulated ignorance fails: A systematic study of {LLM} behaviors on
  forecasting problems before model knowledge cutoff.
\newblock \emph{arxiv}, 2026.
\newblock URL \url{https://arxiv.org/abs/2601.13717}.

\bibitem[{Lightning Rod Labs}(2025)]{Foresight}
{Lightning Rod Labs}.
\newblock Foresight-32b: An {LLM}-based forecasting system, 2025.
\newblock URL
  \url{https://blog.lightningrod.ai/p/using-the-future-to-train-prediction-models}.

\bibitem[Liptay et~al.(2026)Liptay, Schwarz, Poyiadzi, Wildman, and
  Bosse]{Liptay2026}
Tom Liptay, Dan Schwarz, Rafael Poyiadzi, Jack Wildman, and Nikos~I Bosse.
\newblock Evaluating strategic reasoning in forecasting agents.
\newblock \emph{arXiv [cs.AI]}, April 2026.
\newblock URL \url{http://dx.doi.org/10.48550/arXiv.2604.26106}.

\bibitem[Liu et~al.(2026{\natexlab{a}})Liu, Chen, Wang, Zeng, Guo, Hu, Yin,
  Huang, Hao, Yang, Cheng, Yao, Yin, Liu, Cheng, Li, Ma, Wang, Qiu, Liu, Zhang,
  Liu, Wang, Yin, He, Liao, Tian, Zhu, Dai, Zhang, Liu, Zhang, Wu, Gao, Chen,
  Yao, Wen, Aditya~Prakash, Blanchet, Wang, Si, and Huang]{FutureXPro}
Jiashuo Liu, Siyuan Chen, Zaiyuan Wang, Zhiyuan Zeng, Jiacheng Guo, Liang Hu,
  Lingyue Yin, Suozhi Huang, Wenxin Hao, Yang Yang, Zerui Cheng, Zixin Yao,
  Lingyue Yin, Haoxin Liu, Jiayi Cheng, Yuzhen Li, Zezhong Ma, Bingjie Wang,
  Bingsen Qiu, Xiao Liu, Zeyang Zhang, Zijian Liu, Jinpeng Wang, Mingren Yin,
  Tianci He, Yali Liao, Yixiao Tian, Zhenwei Zhu, Anqi Dai, Ge~Zhang, Jingkai
  Liu, Kaiyuan Zhang, Wenlong Wu, Xiang Gao, Xinjie Chen, Zhixin Yao, Zhoufutu
  Wen, B~Aditya~Prakash, Jose Blanchet, Mengdi Wang, Nian Si, and Wenhao Huang.
\newblock {FutureX}-pro: Extending future prediction to high-value vertical
  domains.
\newblock \emph{arXiv [cs.AI]}, January 2026{\natexlab{a}}.
\newblock URL \url{http://dx.doi.org/10.48550/arXiv.2601.12259}.

\bibitem[Liu et~al.(2026{\natexlab{b}})Liu, Kang, Wang, Kai, Liang, Zhen,
  Zhong, Yuan, and Song]{Liu2026}
Mingyang Liu, Qingcan Kang, Yuke Wang, Shixiong Kai, Kaichao Liang, Hui-Ling
  Zhen, Tao Zhong, Mingxuan Yuan, and Linqi Song.
\newblock From long news to accurate forecast: Importance-aware fusion and
  {PRM}-guided reflection for time series forecasting.
\newblock \emph{arXiv [cs.AI]}, June 2026{\natexlab{b}}.
\newblock URL \url{http://dx.doi.org/10.48550/arXiv.2606.03097}.

\bibitem[Liu et~al.(2025)Liu, Han, Yu, Li, and You]{TimeR1}
Zijia Liu, Peixuan Han, Haofei Yu, Haoru Li, and Jiaxuan You.
\newblock {Time-R1}: Towards comprehensive temporal reasoning in {LLMs}.
\newblock \emph{arxiv}, 2025.
\newblock URL \url{https://arxiv.org/abs/2505.13508}.

\bibitem[Lou et~al.(2026)Lou, Lázaro-Gredilla, Dedieu, Wendelken, Lehrach, and
  Murphy]{autoHarness}
Xinghua Lou, Miguel Lázaro-Gredilla, Antoine Dedieu, Carter Wendelken,
  Wolfgang Lehrach, and Kevin~P. Murphy.
\newblock {AutoHarness: improving LLM agents by automatically synthesizing a
  code harness}.
\newblock \emph{arxiv}, 2026.
\newblock URL \url{https://arxiv.org/abs/2603.03329}.

\bibitem[{Metaculus}(2026)]{FutureEval}
{Metaculus}.
\newblock {FutureEval}: Continuously updated {AI} forecasting benchmark, 2026.
\newblock URL \url{https://www.metaculus.com/futureeval/}.

\bibitem[Paleka et~al.(2026)Paleka, Goel, Geiping, and Tramer]{Paleka2025}
Daniel Paleka, Shashwat Goel, Jonas Geiping, and Florian Tramer.
\newblock Pitfalls in evaluating language model forecasters.
\newblock \emph{{ICLR}}, 2026.
\newblock URL
  \url{https://proceedings.iclr.cc/paper_files/paper/2026/hash/b9d69c72e4587ddf3fa6352890aefc4f-Abstract-Conference.html}.

\bibitem[Papamarkou et~al.(2026)Papamarkou, Alquier, Bauer, Buntine, Davison,
  Dziugaite, Filippone, Foong, Fortuin, Fouskakis, Frellsen, Hüllermeier,
  Karaletsos, Khan, Kotelevskii, Lahlou, Li, Liu, Lyle, Möllenhoff, Palla,
  Panov, Sale, Schweighofer, Shelmanov, Swaroop, Trapp, Waegeman, Wilson, and
  Zaytsev]{Papamarkou2026}
Theodore Papamarkou, Pierre Alquier, Matthias Bauer, Wray Buntine, Andrew
  Davison, Gintare~Karolina Dziugaite, Maurizio Filippone, Andrew Y~K Foong,
  Vincent Fortuin, Dimitris Fouskakis, Jes Frellsen, Eyke Hüllermeier,
  Theofanis Karaletsos, Mohammad~Emtiyaz Khan, Nikita Kotelevskii, Salem
  Lahlou, Yingzhen Li, Fang Liu, Clare Lyle, Thomas Möllenhoff, Konstantina
  Palla, Maxim Panov, Yusuf Sale, Kajetan Schweighofer, Artem Shelmanov,
  Siddharth Swaroop, Martin Trapp, Willem Waegeman, Andrew~Gordon Wilson, and
  Alexey Zaytsev.
\newblock Position: agentic {AI} orchestration should be bayes-consistent.
\newblock In \emph{ICML}, May 2026.
\newblock URL \url{http://dx.doi.org/10.48550/arXiv.2605.00742}.

\bibitem[Platt(1999)]{platt1999}
John~C. Platt.
\newblock Probabilistic outputs for support vector machines and comparisons to
  regularized likelihood methods.
\newblock In \emph{Advances in Large Margin Classifiers}, pp.\  61--74. MIT
  Press, 1999.

\bibitem[Pratt et~al.(2024)Pratt, Blumberg, Carolino, and Morris]{Pratt2024}
Sarah Pratt, Seth Blumberg, Pietro~Kreitlon Carolino, and Meredith~Ringel
  Morris.
\newblock Can language models use forecasting strategies?
\newblock \emph{arxiv}, 2024.
\newblock URL \url{https://arxiv.org/abs/2406.04446}.

\bibitem[Qiu et~al.(2025)Qiu, Sha, Allen, Kim, Linzen, and van
  Steenkiste]{Qiu2025}
Linlu Qiu, Fei Sha, Kelsey Allen, Yoon Kim, Tal Linzen, and Sjoerd van
  Steenkiste.
\newblock Bayesian teaching enables probabilistic reasoning in large language
  models.
\newblock \emph{Nat. Commun.}, March 2025.
\newblock URL \url{http://arxiv.org/abs/2503.17523}.

\bibitem[Schoenegger et~al.(2024)Schoenegger, Tuminauskaite, Park, and
  Tetlock]{Schoenegger2024crowd}
Philipp Schoenegger, Indre Tuminauskaite, Peter~S. Park, and Philip~E. Tetlock.
\newblock Wisdom of the silicon crowd: {LLM} ensemble prediction capabilities
  rival human crowd accuracy.
\newblock \emph{arxiv}, 2024.
\newblock URL \url{https://arxiv.org/abs/2402.19379}.

\bibitem[Schoenegger et~al.(2025)Schoenegger, Jones, Tetlock, and
  Mellers]{Schoenegger2025prompts}
Philipp Schoenegger, Cameron~R. Jones, Philip~E. Tetlock, and Barbara Mellers.
\newblock Prompt engineering large language models' forecasting capabilities.
\newblock \emph{arxiv}, 2025.
\newblock URL \url{https://arxiv.org/abs/2506.01578}.

\bibitem[Spiegelhalter(2025)]{Spiegelhalter2025}
David Spiegelhalter.
\newblock \emph{The Art of Uncertainty: How to Navigate Chance, Ignorance, Risk
  and Luck}.
\newblock W.W. Norton, 2025.

\bibitem[Stein(1956)]{stein1956}
Charles Stein.
\newblock Inadmissibility of the usual estimator for the mean of a multivariate
  normal distribution.
\newblock In \emph{Proceedings of the Third Berkeley Symposium on Mathematical
  Statistics and Probability}, pp.\  197--206, 1956.

\bibitem[Tetlock \& Gardner(2015)Tetlock and Gardner]{Tetlock2015}
Philip~E. Tetlock and Dan Gardner.
\newblock \emph{Superforecasting: The Art and Science of Prediction}.
\newblock Crown, 2015.

\bibitem[Turtel et~al.(2025)Turtel, Franklin, Skotheim, Hewitt, and
  Schoenegger]{Turtel2025}
Benjamin Turtel, Danny Franklin, Kris Skotheim, Luke Hewitt, and Philipp
  Schoenegger.
\newblock Outcome-based reinforcement learning to predict the future.
\newblock \emph{arxiv}, 2025.
\newblock URL \url{https://arxiv.org/abs/2505.17989}.

\bibitem[Turtel et~al.(2026)Turtel, Wilczewski, Franklin, and
  Skothiem]{Turtel2026}
Benjamin Turtel, Paul Wilczewski, Danny Franklin, and Kris Skothiem.
\newblock Future-as-label: Scalable supervision from real-world outcomes.
\newblock \emph{arxiv}, 2026.
\newblock URL \url{https://arxiv.org/abs/2601.06336}.

\bibitem[Waghmare \& Ziegel(2025)Waghmare and Ziegel]{Waghmare2025}
Kartik Waghmare and Johanna Ziegel.
\newblock Proper scoring rules for estimation and forecast evaluation.
\newblock \emph{arxiv}, 2025.
\newblock URL \url{https://arxiv.org/abs/2504.01781}.

\bibitem[Wang et~al.(2026)Wang, Wang, Deng, and Tang]{Wang2026trade}
Yishu Wang, Yuxuan Wang, Jiaqi Deng, and Hanyang Tang.
\newblock Beyond forecasting: The belief-to-trade layer in prediction-market
  agents.
\newblock \emph{arXiv [cs.AI]}, July 2026.
\newblock URL \url{http://dx.doi.org/10.48550/arXiv.2607.03015}.

\bibitem[Wang et~al.(2025)Wang, Zhou, Yang, Ma, Wang, Dong, and
  Anwar]{CogForecast}
Zhen Wang, Xi~Zhou, Yating Yang, Bo~Ma, Lei Wang, Rui Dong, and Azmat Anwar.
\newblock Beyond inherent cognition biases in {LLM}-based event forecasting:
  {A} multi-cognition agentic framework.
\newblock In \emph{Findings of EMNLP}, 2025.
\newblock URL \url{https://aclanthology.org/2025.findings-emnlp.258/}.

\bibitem[Yang et~al.(2025)Yang, Mahns, Li, Gu, Wu, and Xu]{ProphetArena}
Qingchuan Yang, Simon Mahns, Sida Li, Anri Gu, Jibang Wu, and Haifeng Xu.
\newblock {LLM}-as-a-prophet: Understanding predictive intelligence with
  prophet arena.
\newblock \emph{arxiv}, 2025.
\newblock URL \url{https://arxiv.org/abs/2510.17638}.

\bibitem[Yao et~al.(2023)Yao, Zhao, Yu, Du, Shafran, Narasimhan, and
  Cao]{ReAct}
Shunyu Yao, Jeffrey Zhao, Dian Yu, Nan Du, Izhak Shafran, Karthik Narasimhan,
  and Yuan Cao.
\newblock {ReAct}: Synergizing reasoning and acting in language models.
\newblock In \emph{ICLR}, 2023.
\newblock URL \url{http://dx.doi.org/10.48550/arXiv.2210.03629}.

\bibitem[Zeng et~al.(2025)Zeng, Liu, Chen, He, Liao, Tian, Wang, Wang, Yang,
  Yin, Yin, Zhu, Cai, Chen, Chen, Du, Gao, Guo, Hu, Jiao, Li, Liu, Ni, Wen,
  Zhang, Zhang, Zhou, Blanchet, Qiu, Wang, and Huang]{FutureX}
Zhiyuan Zeng, Jiashuo Liu, Siyuan Chen, Tianci He, Yali Liao, Yixiao Tian,
  Jinpeng Wang, Zaiyuan Wang, Yang Yang, Lingyue Yin, Mingren Yin, Zhenwei Zhu,
  Tianle Cai, Zehui Chen, Jiecao Chen, Yantao Du, Xiang Gao, Jiacheng Guo,
  Liang Hu, Jianpeng Jiao, Xiangsheng Li, Jingkai Liu, Shuang Ni, Zhoufutu Wen,
  Ge~Zhang, Kaiyuan Zhang, Xin Zhou, Jose Blanchet, Xipeng Qiu, Mengdi Wang,
  and Wenhao Huang.
\newblock {FutureX}: An advanced live benchmark for {LLM} agents in future
  prediction.
\newblock \emph{arxiv}, 2025.
\newblock URL \url{https://arxiv.org/abs/2508.11987}.

\bibitem[Zhang et~al.(2026)Zhang, Liu, Johansson, Yitayew, Ohly, and
  Li]{PredictionArena}
Jaden Zhang, Gardenia Liu, Oliver Johansson, Hileamlak Yitayew, Kamryn Ohly,
  and Grace Li.
\newblock Prediction arena: Benchmarking {AI} models on real-world prediction
  markets.
\newblock \emph{arxiv}, 2026.
\newblock URL \url{https://arxiv.org/abs/2604.07355}.

\bibitem[Zou et~al.(2022)Zou, Xiao, Jia, Kwon, Mazeika, Li, Song, Steinhardt,
  Evans, and Hendrycks]{Zou2022}
Andy Zou, Tristan Xiao, Ryan Jia, Joe Kwon, Mantas Mazeika, Richard Li, Dawn
  Song, Jacob Steinhardt, Owain Evans, and Dan Hendrycks.
\newblock Forecasting future world events with neural networks.
\newblock In \emph{NeurIPS (Datasets and Benchmarks)}, 2022.
\newblock URL \url{https://arxiv.org/abs/2206.15474}.

\end{thebibliography}

\newpage
\appendix

\section{Experimental Setup Details}
\label{app:setup}

\subsection{Datasets}

We use two binary-question benchmarks: AIBQ2 (113 questions from the
Metaculus AI Benchmark Tournament, Q2 2025) and ForecastBench (FB,
$\sim$500 questions per fortnightly round). Full descriptions, exam
construction, base rates, and a representativeness analysis are in
\cref{app:datasets}.

\subsection{Metrics}
\label{app:metrics}

We use multiple metrics to evaluate performance, as defined below.

\begin{itemize}

    \item \textbf{Metaculus Baseline Score}:
    $\MBS = 100 \log_2(p_o / \pi_0) = 100(\log_2 p_o + 1)$ where
    $p$ is the predicted probability of a YES outcome,
    $p_o=p$ if $o=1$, $p_o = 1-p$ if $o=0$, and $\pi_0=0.5$ is the uniform baseline.\footnote{
  See 
 \url{https://www.metaculus.com/help/scores-faq/\#baseline-score}
  for the full derivation.
  Metaculus also defines a \textbf{Metaculus Peer Score (MPS)}
that measures performance relative to other forecasters
rather than a uniform baseline:
$\text{MPS} = 100 \times (\ln(p) - \ln(\text{GM}(q_i)))$,
where $\text{GM}(q_i)$ is the geometric mean of all
other predictions for the same question.
  (See \url{https://www.metaculus.com/help/scores-faq/\#peer-score}.)
We do not use MPS in this paper, but it could be
approximated by using the crowd estimate
as the reference prediction (cf.\ the adjusted Brier score below).
}
A perfect prediction ($p_o=1$) gives $\MBS=100$,
  while the baseline ($\pi_0=0.5$) gives $\MBS=0$.
MS punishes incorrect predictions in the tails more severely
than Brier score. Higher is better.

   \item \textbf{Brier Score}: $\BS = (p - o)^2$,
    where $p$ is the forecast and $o \in \{0,1\}$ the outcome.
    Lower is better (despite the use of the word ``score''),
    and predicting the uniform distribution of $p=0.5$ scores 0.25.
    BS is unscaled unless a table explicitly labels BS$\times100$
    (so 0.25 becomes 25).

    \item \textbf{Brier Index}:
    $\BI = 100 \times (1 - \sqrt{\overline{\BS}})$,
    where $\overline{\BS} = \frac{1}{N}\sum_i (p_i - o_i)^2$
    is the mean Brier score across $N$ predictions.
    Higher is better; always-0.5 scores 50\%.
    Note the square root is applied \emph{after} averaging,
    not per question.
    See \citep{BrierIndex} for details.
    Within a common evaluation split, BI and mean BS give equivalent rankings.
    Averaging Market and Data BI need not preserve the ranking of
    the equally weighted mean of their BS values.

       \item \textbf{Adjusted Brier Score}:
    $\text{ABS}_i = \BS_i - \gamma_i + \bar\gamma$ is the difficulty adjusted
    Brier score for question~$i$, where $\gamma_i$ is an
    estimate of question $i$'s difficulty, and $\bar\gamma$ is the reference
    mean difficulty used to center the adjustment \citep{Bastani2025,forecastbenchUpdate}.
    For market questions, $\gamma_i = (m_i - o_i)^2$ where
    $m_i$ is the market price (crowd estimate).
    For dataset questions, 
    $\gamma_i$ is a fixed effect estimated across
    all methods on the ForecastBench.\footnote{
      The $\gamma_i$ values are available at
      \url{https://www.forecastbench.org/datasets/}.
      However, relying on $\gamma_i$ means this metric
    cannot be used for novel datasets. In \cref{app:mixed-effects},
    we use a paired analysis 
    to achieve a similar effect in a more general way.
    }
  ABS controls for question difficulty, so a method that
    does well on hard questions scores better than one that
    only does well on easy questions. ABS was used
    as the official metric on the FB leaderboard until March 2026,
    when they switched to adjusted Brier index (see below).

    \item \textbf{Adjusted Brier Index}:
    $\text{ABI} = 100 \times \bigl(1 - \sqrt{\max(0,\,\overline{\text{ABS}})}\bigr)$,
    where $\overline{\text{ABS}} = \frac{1}{N}\sum_i \text{ABS}_i$ is the mean
    adjusted Brier score across $N$ predictions.
    As of March 2026, this is the official metric used by the FB leaderboard
    \citep{BrierIndex}.\footnote{
      Per Eq.~(2) of \citet{forecastbenchUpdate}, ABS is rescaled so that
      the always-0.5 forecaster scores 0.25 (matching raw BS), but
      $\text{ABS}_i = b_i - \hat\gamma_j + \overline{\hat\gamma}$ can still
      be negative for individual questions (whenever the forecaster
      out-performs the difficulty estimate $\hat\gamma_j$ on question $j$),
      and the mean $\overline{\text{ABS}}$ can in principle dip below zero
      for a forecaster who consistently beats the difficulty estimate
      across its evaluation subset. The $\max(0,\cdot)$ clamp keeps the
      index well-defined; in practice all top FB-leaderboard methods (and
      all methods reported in this paper) have positive
      $\overline{\text{ABS}}$, so the clamp is inert.
    }

\end{itemize}

MS and BS metrics are proper  scoring rules,
which reward well-calibrated forecasts
(see e.g., \citep{Gneiting2007,Waghmare2025}).
Technically speaking BI is not proper,
but it converges to properness as the sample size increases
 (see \citep{BrierIndex} for details).
 (Adjusted versions do not affect whether a metric is proper or not.)

\subsection{External methods we compare to}
\label{app:alternatives}

\begin{figure}[h]
  \centering
  \includegraphics[width=\textwidth]{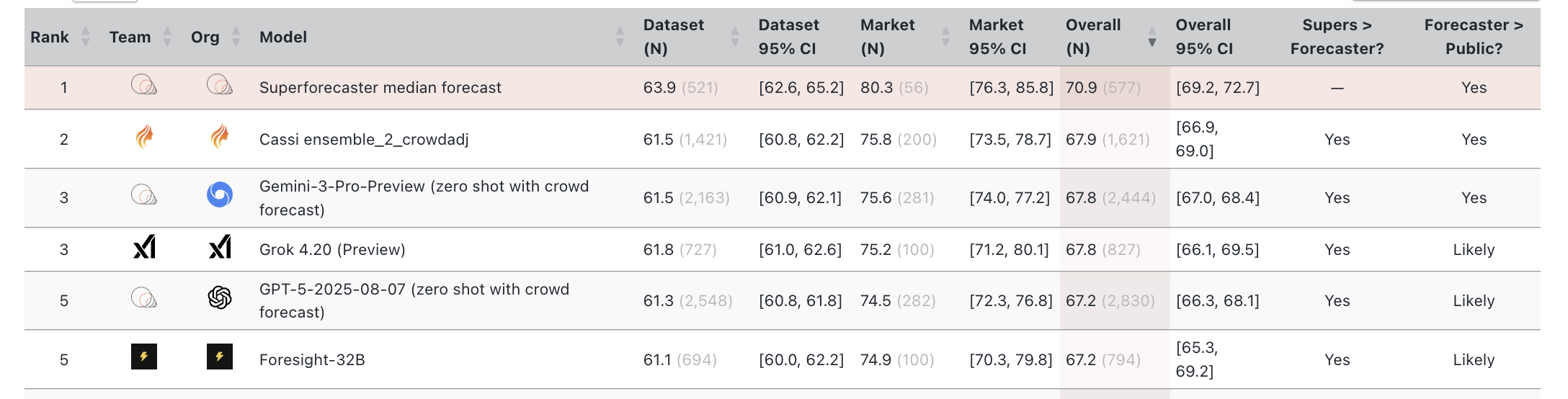}
  \caption{
    \small
    Screenshot of the ForecastBench tournament leaderboard
    (\url{https://www.forecastbench.org/leaderboards/\#tournament})
    as of 2026-04-20, showing the top 5 methods.
    The metric is  Adjusted Brier Index.
    Note that each method's ABI is computed on a different
    set of resolved questions, so direct comparison of
    leaderboard scores is approximate;
    our paired analysis (Table~\ref{tab:mega}) provides
    rigorous comparisons on shared questions.
  }
  \label{fig:fb-leaderboard}
  \label{app:tab:alternatives}
\end{figure}

The historical leaderboard snapshot
of \url{https://www.forecastbench.org/datasets/}
from 2026-04-20 
is shown in \cref{fig:fb-leaderboard}.
The term `zero-shot'' denotes a single probability response
without search or tools, it does not mean they do not use the crowd prior;
we omit all such zero-shot methods,
and just retain the following search-based agents:
\begin{itemize}
  \item \textbf{Cassi} (ensemble+crowd-adj): ABI=67.9; both tranches A and B.
  \item \textbf{Grok~4.20}: ABI=67.8; tranche A only.
  \item \textbf{Foresight-32B}: ABI=67.2; tranche B only.
\end{itemize}

\eat{
For questions after a frozen model's knowledge cutoff, a no-search baseline
cannot obtain recent evidence beyond its supplied context. Its performance
therefore confounds forecasting quality with access to timely information.
Forecasting systems need an up-to-date information source, whether search,
source-specific tools, supplied observations, or continually updated model
knowledge. Our main comparison focuses on agents, while historical zero-shot
results remain in the appendix as diagnostics of information access and
calibration. We do not treat gains over those baselines as evidence for the
structured-belief mechanism. The Gemini zero-shot leaderboard submission is
unavailable on our tranche dates; our own zero-shot runs are diagnostic
ablations, not reproductions of that external system.
}

\subsection{LLMs used}
\label{app:llms}

\Cref{tab:llms} lists every LLM used in this paper, with
provider, version, knowledge-cutoff date, maximum context, an
order-of-magnitude effective parameter-count estimate (closed
proprietary models from \citealp{Li2026probes}; open models
from official model cards), and per-token cost (USD per million
tokens, OpenRouter pricing as of \today).
Knowledge-cutoff and context-window values come from the
respective providers; cost figures are subject to change.

\begin{table}[h]
\centering
\small
\setlength{\tabcolsep}{4pt}
 \begin{threeparttable}
\caption{
\small
LLMs used in this paper. Parameter estimates for closed models 
are from~\citet{Li2026probes}.
Cost estimates: USD per million input/output tokens via OpenRouter
or direct provider.
Input tokens can be cheaper when there is a cache hit.
Cutoff dates are provider-reported knowledge cutoffs.
}
\label{tab:llms}
\begin{tabular}{@{}llcccc@{}}
\toprule
LLM (alias)        & Full version                & Cutoff   & Context & Params       & Cost (\$/M in / \$/M out) \\
\midrule
\multicolumn{6}{@{}l}{\emph{Main agent}} \\
Pro                & gemini-3.1-pro-preview      & Jan 2025 & 1M    & ?            & \$2 / \$12  \\
Flash              & gemini-3-flash-preview      & Jan 2025 & 1M    & ?            & \$0.50 / \$3 \\
GPT-5              & gpt-5.4                     & Aug 2025 & 1M  & $\sim$2T      & \$2.50 / \$15 \\
\midrule
\multicolumn{6}{@{}l}{\emph{Auxiliary roles (search filtering, summarization, leakage audit)}} \\
Flash      & gemini-3-flash-preview      & Jan 2025 & 1M    & ?            & \$0.50 / \$3 \\
Grok-4.1-fast & x-ai/grok-4.1-fast   & May 2025 & 2M    & ?            & \$0.20 / \$0.50 \\
\bottomrule
\end{tabular}
\end{threeparttable}
\end{table}

\subsection{Compute requirements}
\label{app:compute}


Our system uses 3 LLMs:
for the main agent, for search results date filtering,
and for search results summarization.
(We optionally use a 4th LLM, Grok-4.1-fast, to perform
post-hoc leakage auditing, as described in \cref{sec:detective}.)
We use Gemini-3.1-Pro for most of our agent experiments
(although we also experiment with other models),
and use Gemini-3-Flash for the search filtering and summarization
because it is much cheaper.
(An agent that takes 10 steps and returns 10 search results per step
requires 100 summarization LLM calls.)

We access all LLMs through OpenRouter (\url{https://openrouter.ai/}),
  using the LiteLLM interface (\url{https://github.com/BerriAI/litellm}),                                                                              
  which provides a uniform way to use multiple LLM providers (e.g., Google, OpenAI,
  Anthropic, Moonshot).                                                                                                                                
  All experiments are launched from a  laptop.
  A single question takes 1--8 minutes depending on API latency
  and the number of steps taken by the agent.                                                                                     
  (We limit the maximum number of steps to 10, and the maximum time to 10 minutes;                                                                           
  if the agent has not submitted by then, we take its last estimate of $\hat{p}$.)
  Using 50 parallel workers, a full evaluation                                                                                                         
  (5 trials on 400 questions) completes in 1--3 hours wall clock.   

  The agentic methods consume 50--100M tokens across 5 trials,
  the large majority of which are input tokens (system prompt,
  tool schemas, accumulated conversation). At Gemini-3.1-Pro
  pricing (\$2/M input, \$12/M output), evaluating one method on
  the full $A \cup B$ dataset costs approximately \$250. Two
  factors keep this well below the naive uncached estimate
  ($\sim$\$1000): (i) Gemini's automatic context caching reuses
  the stable prefix (system + tools + initial user message) across
  the steps of an agent loop, billing the cached portion at
  $\sim$25\% of the base input rate, and (ii) output tokens, which
  carry the higher per-token rate, are a small fraction of total
  tokens because most steps emit short tool calls or short final
  answers rather than long free-text responses.


\clearpage
\section{Backtesting validity}
\label{app:leakage}
\label{app:leakage-setup}

A key concern with backtesting is information leakage: the LLM's
parametric knowledge or web search results may contain post-cutoff
information that would not be available in a live setting.
(In our backtesting context, the cutoff date is set equal to the
forecast due date.)

\subsection{Mitigating search leakage.}

As mentioned in the main text, we adopt
a four-layer defense:
(1)~search engine date filtering,
(2)~LLM-based leak classifier on results,
(3)~date clamping for tools, and
(4)~URL blocking for resolution sources.
We give more details below.

\paragraph{Stage 1.}
The first layer is to add date range filtering to web search,
so that we do not retrieve pages that are date stamped after the cutoff
date.
Specifically,
we use Brave Search, which supports a
\texttt{freshness} parameter to restrict results to before
the cutoff date. We evaluated other engines
(Perplexity, AskNews, Exa, Serper) but they either did not
support reliable date filtering, or returned only snippets
(not full page text), which harmed downstream performance.

Note that  algorithmic date filtering at the search engine level is
critical: without it, we would need to retrieve many more
results knowing that the LLM-based classifier (layer 2) will
discard most of them, which would be very inefficient.

Also, 
note that web search is controlled entirely through our own tool schemas
passed via the function-calling API, so we can enforce date filtering.
The core LLM could choose to ``spontaneously'' do search on its own without us asking it to;
however,
both Gemini and Anthropic models require explicit opt-in to enable
built-in search~\citep{gemini-grounding, anthropic-search},
which we avoid using.

\paragraph{Stage 2.}
The second layer is to use an LLM classifier to decide if any
results that pass the first stage are after the cutoff date based on their content (not meta data).
We use Gemini-3-Flash for this, since it is fast and cheap and reasonably
reliable.
This stage is necessary because a document might have a time stamp that is before the cutoff date, 
and hence pass the stage 1 filter, but it may have been updated and contain information
from after the cutoff.

\paragraph{Stage 3.}
The tools we use, such as the time series tools discussed in 
\cref{app:tools}, all support  date filtering 
which enforces the requested observation-date cutoff. This alone does not
guarantee historical data-vintage correctness for sources that revise past values.
(For example, we can retrieve yfinance history up to the cutoff date,
or retrieve a  Wikipedia snapshotted page from before the cutoff.)

\paragraph{Stage 4.}
Finally, since sometimes the question or its resolution criteria contain a URL
which might have the resolved outcome
(e.g.
\url{https://manifold.markets/AlexLiesman/will-ding-liren-repeat-as-world-che}),
we block access to such URLs for both web search and the lookup-url tool.

\subsection{Mitigating parametric leakage.}

Besides leakage from web search, the LLM will contain certain forms of knowledge
(including potentially the true outcome of a given forecasting question)
up until its knowledge cutoff date; the amount of memorized world knowledge
scales roughly log-linearly with effective parameter count
\citep{Li2026probes}, so larger frontier models are at higher
parametric-leakage risk.
To avoid this risk, we only use models whose cutoff date is before
the date range of the questions we test on.\footnote{
  An extreme example would be the "Talkie" model from \url{https://talkie-lm.com/introducing-talkie},
  which was only trained on data up until 1930.
  However, this is a very small model that is not good at reasoning or tool use,
  so is not capabable of reliable forecasting.
}


\subsection{Post-hoc leakage audit.}
\label{sec:detective}

To assess the effectiveness of our leakage mitigation techniques,
we run a separate ``leak detective'' that re-classifies all saved
search results and URL lookups conducted by the agent,
as well as examining the reasoning traces.
For the detective, we use Grok-4.1-Fast.
This is fast and cheap, but somewhat uncorrelated with 
the runtime filter (Gemini-3-Flash).

\paragraph{Search leakage analysis}
After classifying all the search results from our \system{} applied to tranche A of FB,
we produced the confusion matrix shown in Table~\ref{app:tab:leakage}.
We see that the runtime filter has
93.8\% recall (catches 320 of 341 leaks) but also drops 577
clean results (29.9\% false positive rate).
Crucially, of the 1{,}377 results the agent actually sees
(the ``kept'' column), only 21 are leaks --- an undetected
flag rate of $1.5\%$ among retained results. These are second-LLM
verdicts, not human-verified ground truth. At the question level, at least one
result was flagged for 14 of 76 web-search questions (18.4\%), or 14 of all
200 tranche-A questions (7.0\%). These exposure rates do not directly measure
how much the flagged evidence affected forecasts. 
\eat{)
A more
conservative \emph{per-question} metric is the fraction of
questions for which the agent's full search history surfaces
at least one leak: $14/76 = 18.4\%$ among the questions that
actually use web search (the rest --- mostly DBnomics --- bypass
search entirely or use only structured tool calls). Counted
against the full 200-question tranche A this is $14/200 = 7.0\%$
of questions exposed to at least one leak. The leaks are
concentrated rather than uniform: most fall on a small number
of questions whose topic was heavily updated post-cutoff
(e.g.\ ongoing geopolitical events). Even at the 7--18\% per-question
rate, leak exposure is still much smaller than the residual
noise sources we quantify in \cref{app:anova} (35\% of
trial-level BI variance is residual inter-trial stochasticity;
question difficulty alone accounts for 62\%).
}
However, if this is a problem,
we could use  a stronger classifier to  reduce FPs even further, while maintaining the high leak detection rate.

\begin{table}[h]
\centering
\caption{
  \small
  {\bf Leakage audit confusion matrix} ($n=2{,}272$ search results
  across 9 sources from Tranche~A).
  Rows: post-hoc verdict from a second LLM classifier (Grok-4.1).
  Columns: runtime filter decision (Gemini-3-Flash).
 }
\label{app:tab:leakage}
\begin{tabular}{lcc}
\toprule
& Filter: Kept & Filter: Dropped \\
\midrule
Post-hoc: Clean & 1{,}354 & 577 \\
Post-hoc: Leak  & 21 & 320 \\
\bottomrule
\end{tabular}
\end{table}

\paragraph{Parametric leakage analysis}
To assess parametric leakage,
we scan the LLM's reasoning traces (including
chain-of-thought) for mentions of post-cutoff events, dates,
or outcomes that could not have been known at the
forecast date.
The retained backbones' advertised knowledge cutoffs precede the
ForecastBench forecast dates. This is a selection criterion, not proof
that their post-training data contain no later information. An absence
of explicit future facts in a reasoning trace likewise cannot rule out
parametric leakage.

\paragraph{Error analysis}
The runtime filter (Gemini-3-Flash) and post-hoc detective
(Grok-4.1) are intentionally different models to reduce
correlated errors. However, both make mistakes, as we now
discuss.

One kind of error is when Flash is too aggressive.
For example,
 a search result about historical weather patterns in Dijon
 (from \url{https://www.weather2visit.com/europe/france/dijon-june.htm})
mentioned ``June 2025'' in a generic seasonal context
(e.g.\ ``What's the weather like in Dijon in June 2025?'').
Flash wrongly classified this as a leak (``describes the weather in
June 2025 which is after the March 2025 cutoff''), but
Grok correctly recognized it as generic climatological
information, not a post-cutoff observation.

Another kind of error is when Grok is too aggressive.
For example, a search result about the Ethiopia conflict 
(from \url{https://acleddata.com/update/ethiopia-weekly-update-13-august-2024})
contained a
link titled ``Ethiopia situation update (30 April 2025)''
while the cutoff was April 27, 2025. Flash kept the result
because the \emph{content} of the page did not include
post-cutoff information --- only the link title referenced
a future date. However, Grok wrongly flagged this as a leak.

These examples illustrate the fundamental difficulty of
date-leakage classification: it requires understanding whether
a date reference describes a \emph{specific observed event}
(leak) or a \emph{generic/scheduled reference} (not a leak).
We could get better results if we use a more powerful classifier,
but this would also be more expensive, so we choose not to do so.

\clearpage
\section{Method Details}
\label{app:system}
\label{app:method}

\subsection{Linguistic Belief State}
\label{app:belief-state}

The core innovation of our system is the \emph{Linguistic
belief state}: a semi-structured representation that the LLM
maintains and updates at each step of the agent loop.
The belief state consists of:
\begin{itemize}
  \item A \textbf{probability estimate} $p \in [0,1]$ for the
    binary outcome;
  \item A \textbf{confidence level} (low/medium/high);
  \item \textbf{Key evidence} for and against the outcome
    (natural-language summaries);
  \item \textbf{Open questions} the agent plans to investigate next.
\end{itemize}
At each step, the LLM receives the full message history
(including the current belief state) alongside
new evidence (search results, tool outputs), and produces an
updated belief state, which is appended to the history
for the next step.
See \cref{app:loop} for details.

This approach contrasts with two common alternatives:
(1)~\emph{text accumulation}, where all retrieved evidence is
appended to an ever-growing context\footnote{
  For examples of this kind of method, 
  See e.g., \url{https://github.com/ai-prophet/mini-prophet/} (public)
  and \citep{AIA} (private).
}, which may exceed
the model's effective attention span; and
(2)~\emph{batch-search then reason}, where multiple queries are
issued in parallel, and then the model reasons about all of them
all at once, without iterative (sequential) refinement of the search queries.

\subsection{Agent Loop}
\label{app:loop}



\begin{figure}[t]
  \centering
  \resizebox{0.6\columnwidth}{!}{



\begin{tikzpicture}[
    node distance=1.0cm and 0.8cm,
    >=Stealth,
    base/.style={draw, rounded corners, text centered, minimum height=0.85cm, font=\small},
    input/.style={base, fill=yellow!20, text width=2.8cm},
    llmnode/.style={base, fill=blue!12, text width=2.2cm},
    action/.style={base, fill=orange!15, text width=1.5cm, minimum height=0.75cm},
    belief/.style={base, fill=green!10, text width=1.8cm},
    msgnode/.style={base, fill=yellow!25, text width=2.2cm},
    submit/.style={base, fill=gray!20, text width=1.3cm, minimum height=0.75cm},
    store/.style={draw, cylinder, shape border rotate=90, aspect=0.25,
                  minimum height=0.7cm, minimum width=1.3cm, font=\small, fill=white},
    elbl/.style={font=\scriptsize},
    cutoff_lbl/.style={font=\footnotesize\color{gray}, inner sep=2pt}
]

    \node [input] (question) {Question $q$};
    \node [input, right=5cm of question] (cutoff) {Cutoff date $d$};

    \node [msgnode, text width=2.5cm, below=0.8cm of question] (history) {History $m_{t-1}$};

    \node [llmnode, below=1.0cm of history] (llm) {LLM (main)};

    \node [belief, below=1.5cm of llm, xshift=-5cm] (belief) {Belief $b_{t-1}$};
    \node [submit, right=0.4cm of belief] (stop) {Stop};
    \node [action, right=0.4cm of stop] (browse) {Browse};
    \node [action, right=0.4cm of browse] (urllookup) {URL lookup};
    \node [action, right=0.4cm of urllookup] (read) {Read};
    \node [action, right=1.2cm of read] (tools) {Data tools};

    \node [llmnode, below=1.4cm of browse, xshift=-1cm] (filter_b) {LLM (filter)};
    \node [store, right=1.8cm of filter_b] (filestore) {File store};
    \node [llmnode, right=1.8cm of filestore] (summarize) {LLM (summ.)};

    \node [msgnode, below=1.5cm of filter_b, xshift=-0.5cm] (concat) {Concat $\to m_t$};


    \draw [->] (question) -- (history);

    \draw [gray, thick, ->, dotted] (cutoff.south west) -- (browse.north east);
    \draw [gray, thick, ->, dotted] (cutoff.south) .. controls ++(0,-2.5) and ++(3,1) .. (filter_b.east);
    \draw [gray, thick, ->, dotted] (cutoff.south) -- (tools.north);

    \draw [->] (history) -- (llm);

    \draw [->] (llm) -- (belief);
    \draw [->] (llm) -- node[left, elbl] {submit($\hat{p}$)} (stop);
    \draw [->] (llm) -- node[right, elbl] {query} (browse);
    \draw [->] (llm) -- node[right, elbl] {url} (urllookup);
    \draw [->] (llm) -- node[right, elbl] {file ids} (read);
    \draw [->] (llm) -- node[right, elbl] {call(args)} (tools);

    \draw [->] (browse) -- node[left, elbl] {\shortstack[c]{snippets\\+ pages}} (filter_b);
    \draw [->] (filter_b) -- node[above, elbl] {filtered pages} (filestore);
    \draw [->] (filter_b.south) -- node[right, elbl] {filtered results} (concat);

    \draw [->] (urllookup.south) -- ++(0,-0.4) -| node[elbl, above, pos=0.3] {page} (filter_b.north);

    \draw [->] (read.south) -- ++(0, -0.5) -| node[elbl, above left, pos=0.8] {file ids} (filestore);
    \draw [->] (filestore) -- node[above, elbl] {pages} (summarize);
    \draw [->] (summarize.south) |- node[above, elbl, pos=0.55] {summary}
          ($(concat.east) + (0, -0.2)$);

    \draw [->] (tools.south) -- ++(0, -0.8)
          -| ($(summarize.south) + (1.2, 0)$) |-
          node[below, elbl, pos=0.85] {time series}
          ($(concat.east) + (0, -0.5)$);

    \draw [->] (belief) -- (concat);

    \draw [->, looseness=1.2]
        (history.west) .. controls ++(-1.8, -3.5) and ++(-1.8, 2.5) .. (concat.west);

    \draw [->, looseness=1.2]
        (concat.south west) .. controls ++(-2.5, -1) and ++(-2.5, 1) ..
        (history.north west);

\end{tikzpicture}

}
  \caption{
    \small
    {\bf \system agent loop}. At each step, the LLM reads the
    message history $m_{t-1}$ and produces an action $a_t$
    and belief $b_{t-1}$ summarizing evidence through step $t-1$. The action is executed
    in the environment (with cutoff-date restrictions),
    producing an observation $o_t$.
    The loop terminates at submit or \texttt{max\_steps}.}
  \label{fig:system}
\end{figure}

The agent operates as an iterative tool-use loop,
as shown in 
Figure~\ref{fig:system} and Algorithm~\ref{alg:agent}.
To explain in more detail, let us define some notation:
$q$ is the question;
$t$ is the time step;
 $a_t$ is the action chosen by the agent;
$o_t$ is the resulting observation from the environment;
$b_{t-1}$ is the belief state;
and $m_t$ is the (non-Markovian) message history,
$m_0=c$, $m_{t-1}=c\oplus\bigoplus_{j=1}^{t-1}(a_j,o_j,b_{j-1})$ in full-BLF, where $c$ contains fixed instructions, question, cutoff, and prior.
The belief state $b_{t-1}$ is a semi-structured object containing
the agent's probability estimate
$b_{t-1}.p \approx P(\text{outcome}=1 \mid m_{t-1})$,
as well as natural-language evidence summaries,
as we discussed in 
\cref{app:belief-state}.
Consequently, we call our agent the ``Bayesian Linguistic Forecaster''
or \textbf{BLF}.

At each step, the LLM produces both an action $a_t$ and an updated belief
state $b_{t-1}$ in a single generation:
  $(a_t,\, b_{t-1}) = \text{LLM}(m_{t-1})$.
The belief update is embedded as a
structured JSON field (\texttt{updated\_belief}) within
the tool call arguments, so both outputs come from one
LLM call.
This field includes the updated probability $b_{t-1}.p$,
as well as an \texttt{update\_reasoning} string in which the
LLM explains why its belief changed (e.g., ``page contradicts
initial assessment'' --- see Algorithm~\ref{alg:unrolled}
for an example trace). We have found that
requiring the LLM to articulate its
reasoning for each belief update encourages coherent
probabilistic updating.

The action $a_t$ is executed in the environment,
returning an observation,
$o_t = \text{Env}(a_t; q, d)$,
where $d$ is the cutoff date restricting data access.
Finally 
the message history is updated deterministically by
concatenation: $m_t = m_{t-1} \oplus (b_{t-1}, a_t, o_t)$.\footnote{
This process is thus amenable to KV-caching, which can significantly reduce costs.
Gemini and OpenAI cache stable prefixes automatically, with no
client-side action required.
}
The loop returns the submitted probability when the agent calls \texttt{submit}.
At the step limit $T$, the historical timeout fallback returns $b_{T-1}.p$,
which has not incorporated $o_T$. The controlled memory study instead reserves a final submission
call within the budget (\cref{sec:markov}). Markov-BLF replaces the
history update by $m_t=m_0\oplus(b_{t-1},a_t,o_t)$, retaining the newest observation
but discarding older raw observations and reasoning. Remaining budget is
supplied separately to the model in both memory conditions.

The process is structured to \emph{resemble} sequential
Bayesian updating and decision making for a POMDP, but we are
not actually doing Bayesian inference in any technical sense:
the update rule $(a_t, b_{t-1}) = \mathrm{LLM}(m_{t-1})$ is an LLM
forward pass with no explicit likelihood, no marginalisation,
and no formal posterior --- the LLM does whatever it does, and
we treat the structured belief slots as a working-memory
scaffold that constrains the form of the update rather than
its semantics.
A textbook POMDP loop would first perform a
\emph{belief update}, $b_{t-1} = f(m_{t-1})$, integrating all
available information into a Markov-sufficient belief state,
and then an \emph{action selection}, $a_t = \pi(b_{t-1})$, choosing
which tool to call next or whether to submit. We do not compute
a Markov-sufficient belief: actions are chosen with access to
the entire history $m_{t-1}$, and the structured belief $b_{t-1}$ is
an \emph{additional} field of the message --- a structured
scratchpad derived from $m_{t-1}$ --- that we hope helps the
model maintain coherent probabilistic reasoning across steps.
\eat{
The  shrinkage estimator and the calibration step \emph{are}
proper (empirical) Bayesian procedures; the per-step belief
update is a Bayesian-shaped heuristic rather than a Bayesian
algorithm.
}

\begin{algorithm}[h]
\caption{Simplified agentic trace for $T=4$ steps.
(See \cref{fig:agent-trace} for a real trace.)
  The web-search action returns a set of 10 snippets, as well as the corresponding
  page content; the full content is stored in a filing system (not shown),
  and the snippets are concatenated and returned to the agent.
  The summarize-results tool call will read the relevant files
  generated by  a previous web search, and then summarize their contents
  and return the result to the agent.
  }
\label{alg:unrolled}
\begin{algorithmic}[1]
\STATE $m_0 \gets c$ // fixed instructions, question, cutoff, and prior
\STATE $(a_1,b_0) \gets \text{LLM}(m_0)$
    // $a_1{=}(\text{lookup-url}, \text{url})$;\enspace $b_0.p {=} 0.65$ \textit{``Seems likely based on question''}
\STATE $o_1 \gets \text{Env}(a_1)$ // page text
\STATE $m_1 \gets m_{0} \oplus (a_1, o_1, b_0)$
\STATE $(a_2, b_1) \gets \text{LLM}(m_1)$
    // $a_2{=}(\text{search}, \text{query})$;\enspace $b_1.p {=} 0.40$ \textit{``Page contradicts initial assessment''}
\STATE $o_2 \gets \text{Env}(a_2)$ // $(\text{snippet}_1, \ldots, \text{snippet}_{10})$
\STATE $m_2 \gets m_{1} \oplus (a_2, o_2, b_1)$
\STATE $(a_3,b_2) \gets \text{LLM}(m_2)$
    // $a_3{=}(\text{summarize}, [3,6,9])$;\enspace $b_2.p {=} 0.35$ \textit{``Sources confirm opposition''}
\STATE $o_3 \gets \text{Env}(a_3)$ // $(\text{text}_3, \text{text}_6, \text{text}_9)$
\STATE $m_3 \gets m_{2} \oplus (a_3, o_3, b_2)$
\STATE $(a_4, b_3) \gets \text{LLM}(m_3)$
    // $a_4{=}(\text{submit}, 0.30)$;\enspace $b_3.p {=} 0.30$ \textit{``Confident: evidence is clear''}
\end{algorithmic}
\end{algorithm}

\subsection{Web search tools}
\label{app:search}
\label{app:actions}

The agent can issue web search queries via an API.
We call this tool \browseTool(query).
For backtesting, it is crucial that search support date filtering,
to avoid leakage.
We have found that the Brave search engine, which supports a
\texttt{freshness} parameter to restrict results to before
the cutoff date, to be quite reliable,
although we still need the second-stage content-based date filter
discussed in \cref{sec:leakage}.
We briefly evaluated other search engines
(Perplexity, AskNews, Exa, Serper, Tavily) but they 
all had one or more of the following problems:
unreliable (or non-existent) date filtering;
only returned snippets, not full page text;
expensive.
It might be interesting to explore these other
search engines more carefully in future work.
However, it is important to emphasize that \system
can work with any search engine.


For each search, Brave returns up to 10 short snippets by default,
but we also request extra snippets to get more informative results
(which we find generally helps performance).
The snippets are
immediately added to the context,
but the full results are not,
to avoid overwhelming the attention mechanism.
Instead we save search results to a set of local files,
\texttt{search\_\{i\}\_result\_\{j\}.md}.
We give the agent an additional tool,
called \readTool$(\text{id}_1,\ldots,\text{id}_k)$,
which can load selected files into the context
of a sub-LLM (we use Gemini Flash), which returns a summary to the main LLM.
(The summarizer is given a prompt generated by the main LLM,
so it extracts question-specific facts.)
The main agent can choose which  files it wants to read
based on their snippets.
This progressive disclosure approach --- similar to
how Anthropic uses skills --- avoids
flooding the context with noise from irrelevant results
while letting the agent drill into the most
promising leads.

\begin{table}
  \centering
  \caption{
  List of all the tools and their arguments.
  Here $q$ denotes the question, which contains information
  about its source $s_q$, forecast date $f_q$ and resolution date(s) $r_q$,
  which are needed by certain tools.
  The first letter is bolded, to represent the short-hand we use to describe each tool.
}
\label{tab:all-tools}
  \begin{tabular}{ll}
    Type & Tools \\ \hline
    Web search & $\mathbf{b}$: \browseTool(query),
                 $\mathbf{r}$: \readTool(ids) \\
    Time series & $\mathbf{h}$: \histTool($q$),
                  $\mathbf{m}$: \modelTool($q$), \\
                & $\mathbf{c}$: \comboTool($q$) \\
    Other & $\mathbf{u}$: \urlTool(url),
            $\mathbf{w}$: \wikiTool(url, [section]),
            $\mathbf{x}$: \submitTool\\
  \end{tabular}
\end{table}

\subsection{Time-series tools}
\label{app:tools}

As we briefly discussed in \cref{sec:datasets}, all the dataset questions in FB
require the ability to do  time series forecasting, i.e., predicting a value of interest
at multiple fixed time points into the future and then comparing to a reference value,
to create a binary prediction problem
of the form
\begin{equation}
\hat{p}_q^r = P(Y_q(r) > v_q \mid \text{data}(\leq f))
\label{eqn:classifier}
\end{equation}
where $q$ is the question, $r$ is the resolution date, and $f$ is the forecast date.
Most of these dataset sources have publicly available APIs that allow you to retrieve historical
values of the quantity of interest,
which is very useful for tackling this problem.
(The exception is ACLED, whose API requires special permission to use.)
More precisely, these database retrieval tools return the history of values 
for a given question of interest $q$
up to a given date $t$ in the form of a CSV list:
\begin{equation}
\mathcal{H}(q,t;W) = 
\{ (t_j, y_s(t_j)): t - W \leq t_j \leq t  \}
\label{eqn:history}
\end{equation}
where $W$ is the window size (which we set by hand,
depending on the  source).\footnote{
We also consider more sophisticated versions of the history,
that sample past data
at different strides, allowing us to look back further
without creating very long traces,
but we omit this for simplicity.
Furthermore, we always enforce $t \leq f$, so the agent
cannot cheat by looking into the future.
}
For brevity, we define
$\mathcal{H}_q = \mathcal{H}(q,f_q;W_{s_q})$,
where $f_q$ is the forecast date of question $q$
and $s_q$ is the source of question $q$ (e.g.\ a FRED question).
We denote the tool for fetching this history
by $\histTool(q)$.

Note that $\mathcal{H}_q$ set contains $v_q=y_q(f_q)$,
allowing us to use the correct reference value in our comparisons,
as opposed to the potentially stale ``freeze value'' given in the original question.
This is needed because the freeze value is of the form 
$v_q'=y_q(t')$,
from the time $t' \leq f_q$ when the question was
created.
In some cases, $v_q'$ may be quite different to $v_q$
(see e.g., \cref{fig:dbnomics-zoom}),
so forecasters will be comparing to the wrong reference.

Although providing $\mathcal{H}_q$ to the LLM is often helpful,
sometimes we find that the LLM cannot make sense of this raw numerical data.
In such cases, we can optionally compute an estimate 
of  $\hat{p}_q = \{ \hat{p}_q^r: r \in \text{res-dates}(q)\}$ 
directly from 
$\mathcal{H}_q$ 
by  using simple statistical models
(see Appendix~\ref{app:ts-models} for details).
We denote the tool for computing this model-based estimate by
 $\modelTool(q)$.
We can either use this estimate directly (bypassing the LLM),
or pass it back to the LLM,
just like any other tool call.
Currently we use  $\hat{p}_q$ directly for DBnomics,
since we found the LLM struggled with directly interpreting
the raw data  (see \cref{tab:ts-models-dbnomics}).
For yfinance and FRED, we use a combined tool,
$\comboTool(q)$,
that returns  both  $\mathcal{H}_q$ and $\hat{p}_q$
in a single call (this is for efficiency reasons).

\subsection{Other tools}
\label{app:other-tools}

We now describe a few other tools we make available to the agent.

For polymarket and manifold questions, we let the agent query the history of prices for the relevant
question (we limit to last 60 days,
and apply a date cutoff to stop it seeing into the future).
The hope is that this will let the agent decide how much to trust the market estimate.
However, the results so far seem to indicate these tools are not useful beyond just using web search,
perhaps because the LLM does not know how to interpret this data.
We leave improving this to future work.

We also provide the  
$\urlTool(\text{url})$ to retrieve the contents of a specific URL.
This is useful because the question text, or resolution criteria, often contain specific
URLs with relevant information. However, we have to be careful that the agent does not just lookup
the answer (since date filtering cannot be applied to url lookup).
To combat this, we create a blacklist of sites that might contain the true outcome
for the question, such as \url{polymarket.com}.
If the agent tries to perform a $\urlTool$ call with such an argument,
we return an empty string and an error message (in text form).
This way the agent can learn not to try this invalid action
again (c.f., \citep{autoHarness}).
(Such invalid calls should not happen, since the tool descriptions specifies the constraints,
but smaller models sometimes struggle to obey these instructions).

Finally, we discuss a special tool for Wikipedia questions.
FB considers these  dataset questions because they require
the user to predict the value at multiple future resolution dates.
For example, consider the question
``According to Wikipedia, will Alexandra Kosteniuk have an Elo rating on [resolution dates]
that's at least 1\% higher than on [forecast date]?".
It is potentially possible to predict this answer  just using 
information from web search, but it is easier to do if you
can retrieve the values from the data table specified in the resolution criteria
(for the case of Elo ratings of chess players, 
the relevant data is at \url{https://en.wikipedia.org/wiki/FIDE\_rankings}).
See \cref{fig:wiki-fide}.
Of course, we must avoid leakage.
Fortunately the wikipedia API allows retrieval of a page at or before a given 
timestamp, which we set to the forecast date $f$.
We denote this tool by $\wikiTool(\text{url}, f_q)$.

\subsection{Policies}
\label{app:policies}

\begin{table}
  \centering
  \caption{
  \small
  Elementary actions (tools) available to each question,
  depending on its source type.
  Abbreviations are defined in \cref{tab:all-tools}.
  We define $\toolsSearch = \{ \tb, \tr, \tu, \tx\}$;
  this is available to all sources
  except DBnomics (which uses $\toolsBasic = \{ \tu, \tx\}$,
  without search tools).
  Currently, time-series sources use the combo tool $\tc$
  (returning both history and model estimate);
  the separate $\th$ and $\tm$ tools are a planned
  decomposition for future work.
}
\label{tab:tool-sets}
\begin{tabular}{@{}ll@{}}
\toprule
  Source  $s$ & Tools $\tools_s$ \\
  \midrule
  FRED, yfinance & $\{\tc\} \union \toolsSearch$ \\
  DBnomics & $\{\tm\} \union \toolsBasic$ \\
  Polymarket, Manifold & $\{\th\} \union \toolsSearch$ \\
  Wikipedia & $\{\tw\} \union \toolsSearch$ \\
  RFI, Metaculus, ACLED & $\toolsSearch$ \\
  \bottomrule
\end{tabular}
\end{table}

\begin{table}[h]
\centering
\small
\caption{
  \small
 Typical action sequences (traces) for different question types.
  Superscript $*$ denotes repetition ($0$ to $T{-}1$ times).
  Note that Bypass and Zero-shot have fixed traces.
}
\label{tab:traces}
\begin{tabular}{@{}lll@{}}
\toprule
Policy & Sources & Typical trace \\
\midrule
TS + search &
  yfinance, FRED &
  $\tc \to \tb^* \to \tx$ \\
Market + search &
  Polymarket, Manifold &
  $\th \to \tb^* \to \tx$ \\
Wiki + search &
  Wikipedia &
  $\tw \to \tb^* \to \tx$ \\
Search-only &
  Metaculus, RFI, ACLED &
  $\tb^* \to \tx$ \\
Bypass &
  DBnomics &
  $\tm \to \tx$ \\
Zero-shot &
  (ablation) &
  $\tx$ \\
\bottomrule
\end{tabular}
\end{table}

The set of tools (actions) available to the agent depends on the type of question it is tackling,
as shown in  \cref{tab:tool-sets}.
Denote this set by $\mathcal{A}_q$.
At each step $t$, the agent's policy picks the action
$a_t = \pi(m_{t-1};\actions_q, \mathcal{L})$, where $m_{t-1}$ is the incoming message history.
In our case, the policy is implemented by the LLM $\mathcal{L}$,
as discussed in \cref{app:loop}.
In addition, the LLM generates the next belief state,
$(a_t,b_{t-1}) = \pi(m_{t-1};\mathcal{A}_q,\mathcal{L})$;
 we can think of $b_{t-1}$ as an internal ``mental'' action,
 as opposed to $a_t$ which is sent to the external environment.
We call the sequence of actions generated by this process a 
policy trace,
$\tau = (a_1, \ldots, a_T)$,
where $a_t \in \mathcal{A}_q$, 
$|\tau| \leq T$,
and $a_T = x$ (the submit/stop action).
See \cref{tab:traces} for some typical traces.
(Note that the agent can choose to stop at any time, but we also impose an upper
bound of $T_{\max}=\maxSteps$ steps.)

Some traces do not make much sense.
For example, calling \readTool before \browseTool has no effect,
since no files have been retrieved by the search tool. 
In addition, one or more URLs are often provided in the resolution criteria,
so it makes sense to initially perform \urlTool,
before following up with optional \browseTool calls,
since we know that these URLs are relevant to the question,
but web search may just add noise.
Similarly, when a source-specific tool, such as \wikiTool
or \comboTool, is available, it is usually best to call this first.
We currently encourage this behavior by using
source-specific prompt instructions
(e.g., ``call the FRED tool as your first action''),
but we could enforce it more rigorously by using a
meta-controller, which determines which set of actions
are available at each step.
We denote this by $\actions_{q,t}=\pi_2(s_q, t)$,
where $\pi_2$ is the second-level policy,
indexed by source type $s_q$ and step $t$.
This can be viewed as a form of agentic harness
(c.f., \citep{autoHarness,metaHarness}).

Currently we use a very primitive form of meta-controller.
For DBnomics questions, we set
$\actions_{q,1}=\{m\}$ and $\actions_{q,2}=\{x\}$;
this forces the agent to first call the domain-specific modeling tool
and then submit its (model-based) estimate, skipping all further steps.
For all other question types, we use
$\actions_{q,t}=\actions_q = \tools_{s_q}$,
so the base LLM can choose any action it wants in any order,
limited only by the set of tools applicable to that source.
However, since $\pi_2$ depends only on $(s_q, t)$,
it can be represented as a simple lookup table,
which could be easy to learn,
e.g., via offline cross-validation on backtesting data,
or online via a bandit algorithm.
For questions that do not come from a well-defined source 
(i.e., for anything beyond the FB dataset), we can replace $s_q$ with $c_q$,
which is class label that we estimate for the question.
We leave learning the meta-controller to future work.

\subsection{Crowd estimate and empirical priors}
\label{app:crowd}

We optionally provide the LLM with an informative
starting estimate $\pi_q$, controlled by two independent flags:

\textbf{Crowd signal} (\texttt{crowd=0/1}).
For market questions, ForecastBench provides
a crowd estimate $\pi_q = m_q$ known as the "freeze value": 
this is the market price 
(Polymarket, Manifold, RFI) or Community Prediction
(Metaculus).
This is the most powerful single input for market
questions: simply predicting $\hat{p}_q = m_q$
achieves BI $\approx$ \bi{81.5}
(Table~\ref{tab:mega}).
The crowd flag has no effect on dataset questions
(no market price exists).

\textbf{Empirical prior} (\texttt{emp=0/1}).
For dataset questions, no crowd estimate exists.
Instead, we use the \emph{empirical base rate} for
each source and question subtype as $\pi_q$.
This is computed using all data prior to the question being asked,
approximating what an online learner would quickly
converge to.
Table~\ref{tab:empirical-priors} shows the priors.
The emp flag has no effect on market questions
(which use the crowd signal instead).

\begin{table}[h]
\centering
\small
\caption{Empirical priors $\pi_q$ for dataset questions,
  computed from all ForecastBench questions.
  The different question subtypes are explained in \cref{app:data-viz}.
}
\label{tab:empirical-priors}
\begin{tabular}{@{}llr@{}}
\toprule
Source & Question subtype & Prior $\pi_q$ \\
\midrule
ACLED & 10$\times$ spike & 0.00 \\
ACLED & Any increase & 0.23 \\
Wikipedia & Vaccine & 0.00 \\
Wikipedia & FIDE Elo $\geq$1\% & 0.01 \\
Wikipedia & FIDE rank & 0.68 \\
Wikipedia & Swimming WR & 0.99 \\
FRED & (all) & 0.42 \\
yfinance & (all) & 0.58 \\
DBnomics & (all) & 0.56 \\
\bottomrule
\end{tabular}
\end{table}

\noindent
These priors are strong baselines for sources with extreme base
rates: for example, predicting $\pi_q = 0$ for ACLED 10$\times$
spike questions
(\cref{app:fb-acled})
or Wikipedia vaccine questions
(\cref{app:fb-wikipedia}) is nearly optimal.
On dataset questions, simply submitting these priors achieves
BI of \bi{58.3}; our full system (\system{}) improves this
to \bi{62.7}
(see Table~\ref{tab:mega}).
Note that the crowd and emp signals are not always available
(e.g., AIBQ2 has neither market prices nor empirical priors),
so we also evaluate without them
(\texttt{crowd=0, emp=0}).

\subsection{Multi-Trial Aggregation}
\label{app:aggregation}
\label{app:jensen}

We run $K=5$ independent trials per question,
each producing a forecast $p_k \in (0,1)$.
Sometimes these can differ quite a lot, based on the particular search
queries performed during each trial
(see \cref{fig:belief-trace}).

The simplest aggregation method is the arithmetic mean of the probabilites:
\begin{equation}
  \hat{p} = \frac{1}{K}\sum_{k=1}^K p_k
  \label{eqn:mean-agg}
\end{equation}
We now prove that this improves 
 all three metrics (BS, BI, and the log-based \MBS) in expectation
 as $K$ increases.
Consider $K$ forecasts $p_1, \ldots, p_K$ for a question with
outcome $o$.
By Jensen's inequality applied to the convex function
$f(e) = e^2$:
\begin{equation}
  \text{BS}(\bar{p}) = (\bar{p} - o)^2
  \leq \frac{1}{K}\sum_k (p_k - o)^2
  = \overline{\text{BS}(\{p_k \})},
  \label{eq:jensen-bs}
\end{equation}
i.e.\ scoring the arithmetic-mean forecast is no worse than
the average per-trial Brier score; the same statement holds
for the log-based \MBS{} because $-\log p$ is also convex in $p$,
so Jensen gives $\mathrm{MS}(\bar{p}) \geq \overline{\mathrm{MS}(p_k)}$
in the same fashion. Note this does \emph{not} say that averaging
beats every individual trial --- a single trial can outperform
$\bar p$ by chance --- only that it beats the \emph{mean} of the
per-trial scores in expectation, which is the relevant comparison
when we view the trials as exchangeable draws.
Since BI $= 100(1 - \sqrt{\overline{\text{BS}}})$ is a monotone
decreasing function of $\overline{\text{BS}}$,
arithmetic averaging also improves BI.
Concretely, if averaging across $K$ trials reduces
$\overline{\text{BS}}$ from $B_1$ to $B_K \leq B_1$, then
\begin{equation}
  \text{BI}_K = 100(1 - \sqrt{B_K}) \geq 100(1 - \sqrt{B_1}) = \text{BI}_1.
  \label{eq:jensen-bi}
\end{equation}
The BI improvement with more trials is confirmed empirically (see \cref{fig:aggregation-main}).

Logit averaging is another option, but does not consistently outperform
arithmetic averaging in \cref{fig:aggregation-main}.
This is defined as follows:
\begin{equation}
  \hat{p} = \sigmoid\!\left(\frac{1}{K}\sum_{k=1}^K \mathrm{logit}(p_k)\right)
  \label{eqn:logit-agg}
\end{equation}
(Note that this is a special case of the shrinkage estimator in
\cref{app:shrinkage}.)
Unfortunately, the Jensen based proof that avergaging is better
does not apply to logit-space averaging.
In particular, the convexity used above is convexity of $f(e)=e^2$ as a function of
the \emph{probability-space} error $e = p - o$, which justifies the
arithmetic mean $\bar p = \frac{1}{K}\sum_k p_k$. The Brier score
viewed as a function of the logit $y = \mathrm{logit}(p)$, namely
$\mathrm{BS}_y(y) = (\sigma(y) - o)^2$ where $\sigma$ is the sigmoid,
is \emph{not} globally convex in $y$: it has an inflection where the
sigmoid passes through its steepest slope. Jensen's inequality
therefore does \emph{not} directly imply that the logit-space mean
reduces BS or BI relative to the arithmetic mean.
\eat{
  For the
\emph{log-based} \MBS\ (and the cross-entropy more generally) the
situation is different: the loss \emph{is} convex in the logit, since
cross-entropy is $\log(1{+}e^{-(2o-1)y})$, a softplus of a linear
function. Jensen then guarantees only that
$\mathcal{L}\bigl(\bar y\bigr) \leq \bar{\mathcal{L}(y_k)}$ ---
i.e.\ scoring the logit-mean is no worse than the average of the
per-trial scores. This is \emph{not} the same as saying logit-space
averaging beats arithmetic-space averaging; that comparison is
between $\mathcal{L}(\sigma(\bar y))$ and
$\mathcal{L}(\bar{\sigma(y)})$ and does not follow from a one-line
Jensen argument either way. 
}
Our preference for logit-space averaging
on FB is therefore empirical, not formal.

We attribute the empirical benefits of logit space averaging to two practical effects.
First, the logit mean preserves extremity better, producing
predictions further from $0.5$ when individual trials are concordant.
Second, logit-space averaging degrades
gracefully when an individual trial fails (e.g., due to network timeout).
Such failures fall back to $\hat{p}_k = 0.5$, with
$\mathrm{logit}(0.5)=0$, which is the prior on the logit scale ---
so a dead trial pulls the logit mean only by $-\bar y/K$ where
$\bar y$ is the surviving-trial mean logit, rather than by a
fixed prior-distance proportional to $(p - 0.5) / K$ in the
arithmetic case. For a confident prediction $p \approx 0.95$, one
dead trial in five costs the arithmetic mean $\sim$9 percentage
points but the logit mean only $\sim$4.

\subsection{Shrinkage aggregation}
\label{app:shrinkage}

We also explored \emph{hierarchical shrinkage} as a way to aggregate results.
This is 
inspired by James--Stein estimation~\citep{stein1956,efron1973},
which is a form of empirical Bayes estimation for a certain hierarchical
model, as we explain below.

\paragraph{Model.}
We model the $K$ trial forecasts for a question $q$ in logit
space. Let $y_{qk} = \mathrm{logit}(p_{qk})$ and let $\theta_q$ denote
the ``true'' logit-forecast for the question.
The likelihood is
\begin{equation}
  y_{qk} \mid \theta \sim \mathcal{N}(\theta_q, \sigma_q^2),
  \quad k = 1, \ldots, K,
\end{equation}
where $\sigma_q^2$ is estimated as the sample variance of the
logit-transformed forecasts.
We place a Gaussian prior on $\theta_q$ centered at the
\emph{question-specific} logit-prior
$\mu_q = \mathrm{logit}(\pi_q)$,
\begin{equation}
  \theta_q \sim \mathcal{N}(\mu_q,\; \tau^2),
\end{equation}
where $\pi_q$ is the crowd estimate (for market questions
where it is available) or the empirical/base-rate prior (for
dataset questions; \cref{app:crowd}). When neither
signal is available we fall back to the uninformative prior
$\pi_q = 0.5$, equivalently $\mu_q = 0$.
By standard calculations for linear Gaussian models, the
posterior is $\theta_q \mid y_{q,1:K} \sim \mathcal{N}(m_q, v_q)$ with
\begin{equation}
  m_q = \alpha_q \bar{y}_q + (1-\alpha_q) \mu_q,
  \qquad
  \alpha_q = \frac{K\tau^2}{K\tau^2 + \sigma_q^2},
  \label{eqn:shrinkage-posterior}
\end{equation}
where $\bar{y}_q = \frac{1}{K}\sum_k y_{qk}$.

\paragraph{Prediction.}
Given the posterior mean
$\mathbb{E}[\theta_q \mid y_{q,1:K}] = m_q$,
we use the following plugin approximation to the posterior
predictive distribution:
\begin{equation}
  \hat{p}_q \;=\; \sigmoid\!\bigl(\alpha_q\,\bar{y}_q + (1-\alpha_q)\,\mu_q\bigr)
  \label{eqn:shrinkage-agg}
\end{equation}
where $\sigmoid(x) = 1/(1 + e^{-x})$.
When trials agree ($\sigma_q^2$ small) $\alpha_q \approx 1$ and
$\hat{p}_q$ collapses to logit-space averaging of the trials;
when trials disagree $\alpha_q \to 0$ and $\hat{p}_q$ is pulled
toward $\pi_q$ (the crowd/empirical prior, or 0.5 when
neither is available).


\paragraph{Practical implementation.}
\Cref{eqn:shrinkage-agg} requires a per-question
$\alpha_q = K\tau^2 / (K\tau^2 + \sigma_q^2)$, which depends on the
unknown prior variance $\tau^2$ and the (estimable) per-question
logit-trial variance $\sigma_q^2$. 
The standard empirical Bayesian approach is to maximize the marginal likelihood
wrt the hyperparemeter $\tau$.
We adopt the simpler, and more robust, approach of using LOO-CV to estimate $\tau^2$.

We also evaluated a heuristic shrinkage rule
(\cref{sec:aggregation}) based on the same principle:
that we should shrink toward the prior more aggresibely (i.e. smaller $\alpha_q$)
when there is a lot of cross-trial variation (i.e., )$\sigma_q^2$ is large).
More precisely, we adopt the following estimator:
\begin{equation}
  \alpha_q \;=\; \max\!\bigl(f,\; 1 - c \cdot s_q\bigr),
  \qquad s_q = \mathrm{std}\bigl(y_{q1}, \ldots, y_{qK}\bigr),
  \label{eqn:shrinkage-stdform}
\end{equation}
where $s_q$ is the per-question sample standard deviation of trial
logits, $f \in [0,1]$ is a floor on $\alpha_q$, and $c \geq 0$ controls
how fast $\alpha_q$ contracts toward $f$ as trials disagree. The
two hyperparameters $(f, c)$ are global (shared across questions).
Operationally, $f{=}1$ disables shrinkage entirely ($\alpha_q\equiv 1$,
pure logit averaging), while $f{=}0$ allows full shrinkage to the
prior; intermediate $f$ caps how aggressively a single high-disagreement
question can be shrunk. The interaction is mildly degenerate in the
$f{>}0$ regime: whenever $c \cdot s_q \le 1-f$, the slope $c$ has no
effect (the floor $f$ is binding), so LOO effectively selects between
``always at the floor'' ($c$ large) and ``mostly at the ceiling''
($c{\to}0$).

\eat{
We tune $(f, c)$ by leave-one-out cross-validation on a
$11{\times}11$ grid ($f \in \{0, 0.1, \ldots, 1\}$,
$c \in \{0, 0.2, \ldots, 2\}$), picking the pair that minimizes
mean Brier score on the labeled backtest set; we call the
resulting LOO-tuned variant \emph{shrink-prior-loo}.
LOO picks different operating
points across regimes. On FB at c=1 --- where the crowd signal
is in the prompt and inter-trial variance is already small ---
it selects $f{=}1, c{=}0$, so $\alpha_q \equiv 1$ and the
estimator collapses to logit-space averaging
(\cref{eqn:logit-agg}). On FB at c=0 and on AIBQ2 (no crowd
anchor in the prompt, noisier trials) it selects $f{<}1, c{>}0$,
yielding genuine per-question shrinkage toward $\pi_q$ that
improves all three metrics.
}

\paragraph{Connection to James--Stein.}
This method is analogous to James--Stein
shrinkage~\citep{stein1956,efron1973}, which shrinks individual
estimates toward a common mean to reduce total MSE. Two
differences are worth highlighting. First, classical
James--Stein shrinks toward a single global mean; we instead
shrink toward a \emph{per-question} center $\mu_q$ (the
crowd or empirical prior in logit space) when one is available.
Second, our setting can optimize $(f,c)$ wrt log-scoring rules where
overconfident wrong predictions are catastrophically penalized,
rather than being restricted to squared error.

\subsection{Calibration}
\label{app:calibration}

We apply Platt scaling~\citep{platt1999} as a post-processing
calibration step:
\begin{equation}
  \hat{p}_{\text{cal}} = \sigma\bigl(a \cdot \mathrm{logit}(\hat{p}) + b\bigr),
  \label{eq:platt}
\end{equation}
where $a$ and $b$ minimize log loss on the training records. Historical
calibration results use leave-one-record-out cross-validation: each forecast
entry is calibrated using a model fitted to the other entries. Other horizons
or forecast dates from the same underlying question can remain in training,
so these estimates may be optimistic and do not establish performance on
unseen questions. Intervals are conditional on the fitted predictions.
For deployment, a calibrator can be trained on historical labeled data and
applied to new forecasts. This record-level protocol differs from the
question-held-out aggregation analysis in \cref{sec:aggregation}.

For ForecastBench, which spans 9 heterogeneous sources
(Table~\ref{tab:source-composition}), we also evaluate
\emph{hierarchical Platt scaling}: a shared slope $a$ and
intercept $b$, plus per-source offsets
$\delta_s$:
\begin{equation}
  \hat{p}_{\text{cal}} = \sigma\bigl(a \cdot \mathrm{logit}(\hat{p})
  + b + \delta_s \bigr).
  \label{eq:hier-platt}
\end{equation}
When fitting, we add  L2 regularization,
$\lambda \sum_s \delta_s^2$, to the NLL. We use $\lambda = 1.0$
as a fixed default rather than tuning it formally; the metric
is fairly insensitive to $\lambda$ within an order of magnitude
on a benchmark of our size, so we did not run an outer LOO loop
over $\lambda$ values.
The per-source offsets account for systematic miscalibration
that varies across question types (e.g., the model tends
to be overconfident on polymarket questions but
underconfident on dbnomics).


\clearpage
\section{FB: Further  Results}
\label{app:results}
\label{app:fb-results}

This section presents more ForecastBench results.

\begin{table}[h]
\centering
\small
\caption{
  \small
  Historical external-comparison configuration on each benchmark;
Table~\ref{tab:component-effects} and the memory study instead use arithmetic
averaging without calibration.
  AIBQ2 has no crowd signal, empirical prior, or source-specific tools
  (all questions are Metaculus, search-only).
  DBnomics bypasses the LLM entirely and uses a custom KNN tool
  (see \cref{app:dbnomicsModel}).
}
\label{tab:default-config}
\setlength{\tabcolsep}{4pt}
\begin{tabular}{@{}l ll@{}}
\toprule
Component & AIBQ2 & ForecastBench \\
\midrule
LLM               & Gemini-3.1-Pro      & Gemini-3.1-Pro \\
Thinking level     & high                & high \\
Search engine      & Brave               & Brave \\
Source tools       & (none available)    & yfinance, FRED, Wikipedia, DBnomics; polymarket, manifold \\
Belief state       & yes          & yes \\
Agent mode         & sequential ($T_{\max}=\maxSteps$) & sequential ($T_{\max}=\maxSteps$) \\
Crowd signal       & off (not available) & market price for market questions \\
Empirical prior    & off (not available) & base rate for dataset questions \\
Trials             & $K=\maxTrials$      & $K=\maxTrials$ \\
Aggregation        & logit-space LOO shrinkage    & logit-space LOO shrinkage$^*$ \\
Calibration        & Platt (record-LOO)      & hier.\ Platt (per-source, record-LOO) \\
DBnomics           & n/a                 & KNN bypass (no LLM) \\
\bottomrule
\end{tabular}\\
{\footnotesize $^*$On FB at $c{=}1$ the LOO grid selects $f{=}1, c{=}0$
(no shrinkage), so the aggregator collapses to plain logit-space mean
(\cref{eqn:logit-agg}); shrinkage is still \emph{applied} but is a
no-op at the chosen hyperparameters. These historical summaries differ from the regenerated common-pool
trial-count analysis: on FB without anchors, question-held-out tuning selects
no shrinkage (\cref{fig:aggregation-main}a).}
\end{table}

\subsection{FB: Comparison to SOTA}
\label{app:comparison}
\label{app:comparison-fb}

\begin{table}[ht]
\centering\small
\caption{ForecastBench A$\cup$B comparisons under three metrics.
Higher BI and ABI and lower BS are better. Brackets give 95\% intervals:
bootstrap intervals for BI; ABI intervals are approximate.
BS points and split-wise intervals are obtained by transforming the rounded
BI summaries, so are approximate; no overall BS interval is reported.
ABI adjusts for question difficulty (\cref{app:metrics}).
$^*$ denotes an external leaderboard submission; $^\dagger$ denotes
partial coverage (Grok: A only; Foresight: B only).
For BI and BS, All averages Market and Data with equal weight.
See \cref{fig:sota,tab:mega} for visualization and paired BI tests.}
\label{app:tab:sota-bi}
\label{app:tab:sota-abi}
\label{app:tab:sota-bs}
\setlength{\tabcolsep}{2.5pt}\small\renewcommand{\arraystretch}{1.25}
\begin{tabular}{@{}l rrr !{\gc} rrr !{\gc} rrr@{}}
\toprule
& \multicolumn{3}{c!{\gc}}{BI $\uparrow$} & \multicolumn{3}{c!{\gc}}{ABI $\uparrow$} & \multicolumn{3}{c}{BS$\times100$ $\downarrow$}\\
\cmidrule(lr){2-4}\cmidrule(lr){5-7}\cmidrule(lr){8-10}
Method & Market & Data & All & Market & Data & All & Market & Data & All\\
\midrule
\system{} (Pro) & \shortstack{\Bbi{83.8}\\{\scriptsize [\bi{79},\bi{90}]}} & \shortstack{\Bbi{62.7}\\{\scriptsize [\bi{61},\bi{65}]}} & \shortstack{\Bbi{73.3}\\{\scriptsize [\bi{71},\bi{76}]}} & \shortstack{\Bms{84.7}\\{\scriptsize [\ms{82},\ms{88}]}} & \shortstack{\Bms{62.0}\\{\scriptsize [\ms{59},\ms{65}]}} & \shortstack{\Bms{71.0}\\{\scriptsize [\ms{68},\ms{74}]}} & \shortstack{2.6\\{\scriptsize [1.0,4.4]}} & \shortstack{13.9\\{\scriptsize [12.2,15.2]}} & 8.3 \\
Cassi$^*$ & \shortstack{\bi{82.0}\\{\scriptsize [\bi{78},\bi{86}]}} & \shortstack{\bi{59.6}\\{\scriptsize [\bi{58},\bi{61}]}} & \shortstack{\bi{70.8}\\{\scriptsize [\bi{69},\bi{73}]}} & \shortstack{\ms{82.0}\\{\scriptsize [\ms{79},\ms{85}]}} & \shortstack{\ms{59.6}\\{\scriptsize [\ms{57},\ms{63}]}} & \shortstack{\ms{68.7}\\{\scriptsize [\ms{66},\ms{72}]}} & \shortstack{3.2\\{\scriptsize [2.0,4.8]}} & \shortstack{16.3\\{\scriptsize [15.2,17.6]}} & 9.8 \\
Grok~4.20$^{*\dagger}$ & \shortstack{\bi{79.5}\\{\scriptsize [\bi{74},\bi{86}]}} & \shortstack{\bi{61.4}\\{\scriptsize [\bi{59},\bi{63}]}} & \shortstack{\bi{70.5}\\{\scriptsize [\bi{68},\bi{74}]}} & \shortstack{\ms{77.6}\\{\scriptsize [\ms{75},\ms{81}]}} & \shortstack{\ms{61.4}\\{\scriptsize [\ms{58},\ms{65}]}} & \shortstack{\ms{68.4}\\{\scriptsize [\ms{65},\ms{72}]}} & \shortstack{4.2\\{\scriptsize [2.0,6.8]}} & \shortstack{14.9\\{\scriptsize [13.7,16.8]}} & 9.6 \\
Foresight-32B$^{*\dagger}$ & \shortstack{\bi{81.0}\\{\scriptsize [\bi{75},\bi{86}]}} & \shortstack{\bi{57.6}\\{\scriptsize [\bi{55},\bi{60}]}} & \shortstack{\bi{69.3}\\{\scriptsize [\bi{67},\bi{72}]}} & \shortstack{\ms{82.3}\\{\scriptsize [\ms{79},\ms{85}]}} & \shortstack{\ms{57.6}\\{\scriptsize [\ms{54},\ms{61}]}} & \shortstack{\ms{67.5}\\{\scriptsize [\ms{64},\ms{71}]}} & \shortstack{3.6\\{\scriptsize [2.0,6.2]}} & \shortstack{18.0\\{\scriptsize [16.0,20.2]}} & 10.8 \\
Crowd+emp (no LLM) & \shortstack{\bi{81.5}\\{\scriptsize [\bi{77},\bi{86}]}} & \shortstack{\bi{58.3}\\{\scriptsize [\bi{57},\bi{60}]}} & \shortstack{\bi{69.9}\\{\scriptsize [\bi{68},\bi{72}]}} & \shortstack{\ms{81.5}\\{\scriptsize [\ms{79},\ms{84}]}} & \shortstack{\ms{58.3}\\{\scriptsize [\ms{55},\ms{61}]}} & \shortstack{\ms{67.7}\\{\scriptsize [\ms{65},\ms{71}]}} & \shortstack{3.4\\{\scriptsize [2.0,5.3]}} & \shortstack{17.4\\{\scriptsize [16.0,18.5]}} & 10.4 \\
\bottomrule
\end{tabular}
\end{table}

\begin{table}[h]
\centering
\small
\setlength{\tabcolsep}{4pt}
\caption{
  \small
  Sample sizes (number of binary resolution events) per method,
  split (Market / Data), and tranche, for the comparisons in
  \cref{tab:mega}. BLF and Cassi cover both tranches.
  Grok submitted only on 2025-10-26 (tranche A) and Foresight
  only on 2025-11-09 (tranche B); for the paired comparisons
  involving them, we use only the events from their submitted
  tranche to avoid scoring a forecast against a resolution event
  that falls outside the method's actual ask date.
}
\label{tab:mega-n}
\begin{tabular}{@{}l !{\gc} rrr !{\gc} rrr !{\gc} rrr @{}}
\toprule
 & \multicolumn{3}{c!{\gc}}{Market $n$} & \multicolumn{3}{c!{\gc}}{Data $n$} & \multicolumn{3}{c}{All $n$} \\
\cmidrule(lr){2-4} \cmidrule(lr){5-7} \cmidrule(lr){8-10}
Method & A & B & All & A & B & All & A & B & All \\
\midrule
\system{} (Pro)        & 100 & 100 & 200 & 298 & 293 & 591 & 398 & 393 & 791 \\
Cassi                  & 100 & 100 & 200 & 298 & 293 & 591 & 398 & 393 & 791 \\
Grok$^{\dagger}$       & 100 &   0 & 100 & 298 &   0 & 298 & 398 &   0 & 398 \\
Foresight$^{\dagger}$  &   0 & 100 & 100 &   0 & 293 & 293 &   0 & 393 & 393 \\
\midrule
Baseline (no LLM)      & 100 & 100 & 200 & 298 & 293 & 591 & 398 & 393 & 791 \\
\bottomrule
\end{tabular}
\end{table}




\Cref{app:tab:sota-bi} compares BLF with search-enabled external agents
and the LLM-free crowd+empirical-prior baseline using BI, ABI, and BS.
The BLF configuration is in \cref{tab:default-config}.
BLF has the highest point estimates for BI and ABI and the lowest BS.
On market BI, BLF scores 83.8, Cassi 82.0, and the crowd baseline 81.5;
paired comparisons are needed because method coverage differs.

Table~\ref{tab:mega} reports paired-bootstrap comparisons:
\system{} vs each external method; \cref{tab:vs-crowd-baseline} compares
each LLM-based method with the crowd+emp baseline. The headline
finding is that \emph{\system{} is the only LLM-based method that
significantly beats the crowd+emp baseline on overall BI}
($+$\dbi{3.4}$^{**}$); on the dataset subset alone the gap is
$+$\dbi{4.5}$^{***}$. On the market subset the paired test for
\system{} vs.\ baseline is $+$\dbi{2.3}$^\text{ns}$ (the
difference is in the right direction but is within noise on
$n_{\text{mkt}}{=}200$ events); none of the three external methods
beats the baseline on overall BI either, although Grok is
significantly better on the dataset subset alone.

\subsection{FB: Per-source analysis}
\label{sec:bi-per-source}

\begin{figure}[h]
  \centering
  \includegraphics[width=\textwidth]{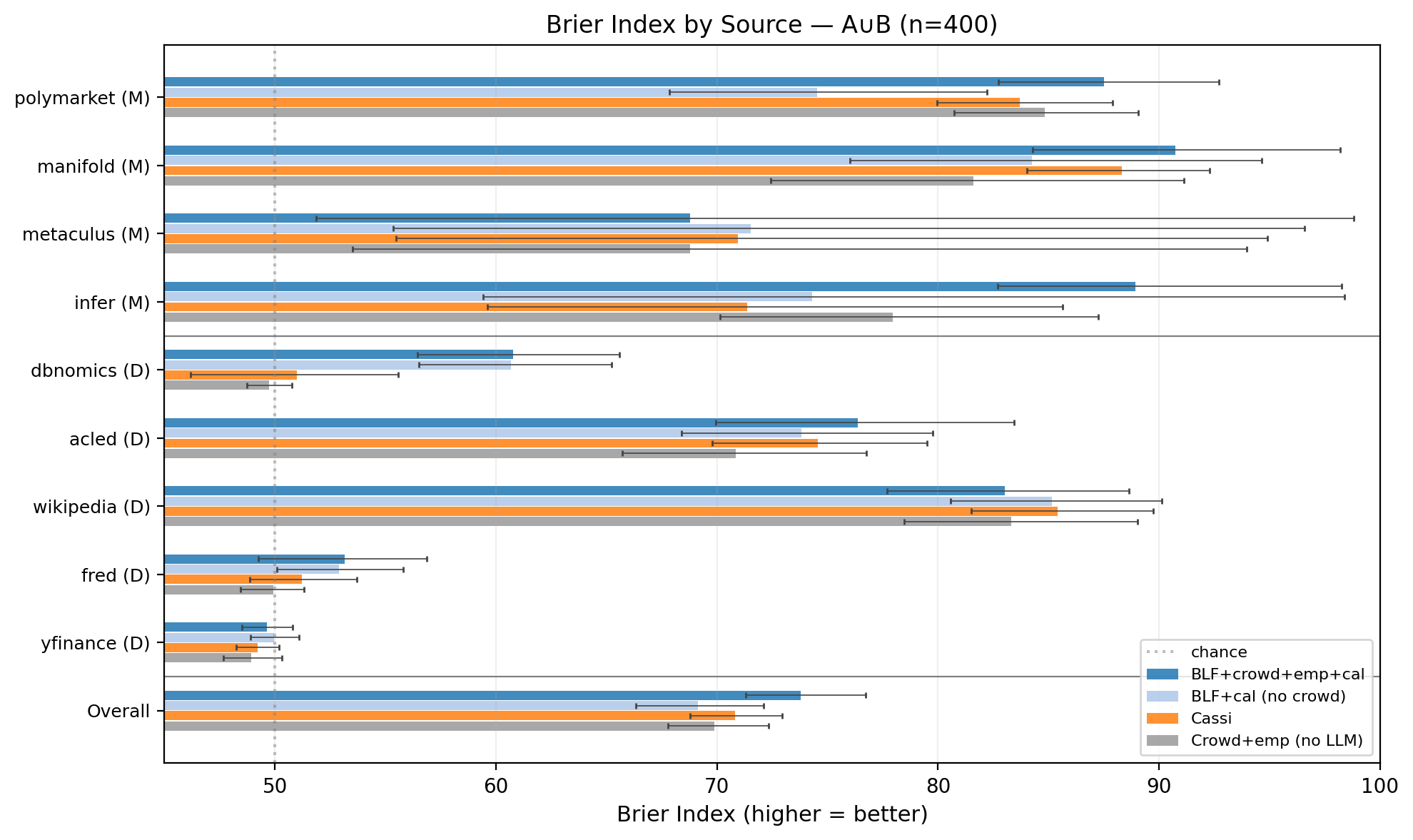}
  \caption{Brier Index by source on FB A$\cup$B, comparing
  \system{} (with and without the crowd anchor on market questions),
  the LLM-free crowd+empirical baseline, and Cassi, the retained external agent covering both tranches.}
  \label{fig:bi-by-source}
\end{figure}

Figure~\ref{fig:bi-by-source} breaks down BI by source,
comparing \system{} against Cassi and two baselines (LLM-free crowd+empirical, and
\system{} without the crowd anchor).
We see that all methods are at chance on yfinance,
due to the random walk nature of stock prices
(see \cref{app:fb-yfinance}).
All methods struggle with FRED
(Federal Reserve Economic Data)
for similar reasons (see \cref{app:fb-fred}),
 although our method has an edge,
 likely due to the simple statistical model
 we use (\cref{eqn:phybrid}),
 whose estimate is made available to the agent to supplement
 the raw data when the fetch-time-series tool is called.
 Our method has an edge on DBnomics for similar reasons:
 the agent uses a simple KNN-based probability estimator,
 bypassing the LLM entirely,
 since we found the LLM struggles with seasonal data
 such as temperature (see the ablation study in 
 \cref{tab:ts-models-dbnomics}).
All methods do quite well on ACLED and Wikipedia,
because the questions are quite easy,
since the base rates are so extreme
(see \cref{app:fb-acled} and
\cref{app:fb-wikipedia}).
On market questions, we outperform other methods by a large margin,
which struggle to beat the prior.
The sources of these gains are studied 
below.


\subsection{Tool Usage}
\label{app:behavior-details}
\begin{figure}[t]
  \centering
  \includegraphics[width=0.6\textwidth]{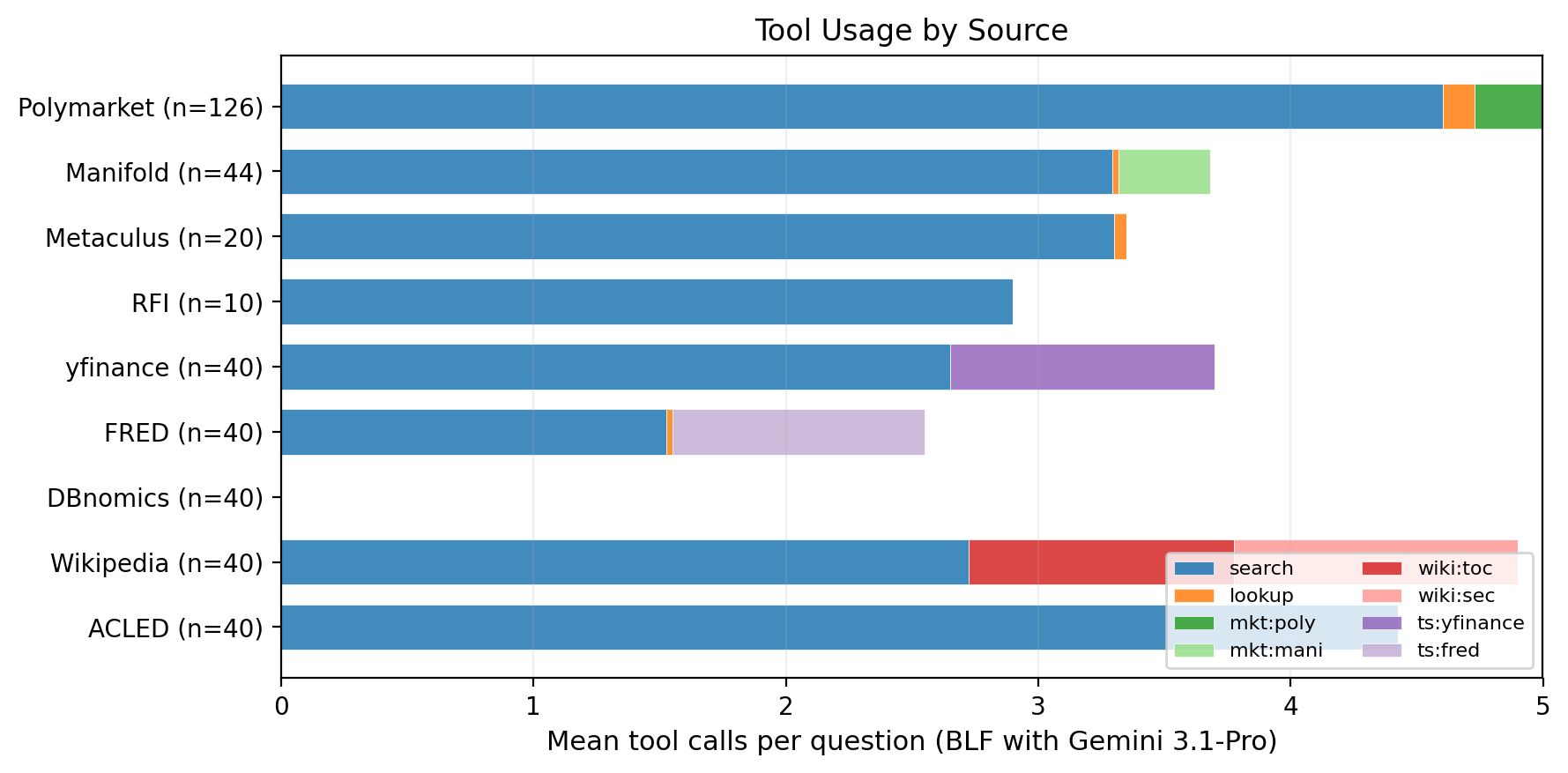}
  \caption{
    \small
    Mean tool calls per question by source for \system{} (trial~1, A$\cup$B).
    Source-specific tools are used selectively.
    DBnomics bypasses the LLM (KNN model).
  }
  \label{fig:tool-calls}
\end{figure}

\paragraph{Tool usage.}
Figure~\ref{fig:tool-calls} shows the mean number of tool calls
per question, broken down by source.
Not surprisingly, web search  dominates across all sources.
Source-specific data tools (market info, Wikipedia sections,
time-series fetchers) are used selectively ---
only for their respective sources.
DBnomics questions are handled entirely by the KNN model
(\cref{app:dbnomicsModel}) with no LLM tool calls.
Polymarket questions use the most tools ($\sim$5.5 per question),
reflecting the additional market-info fetch
on top of web search.

\eat{
\subsection{Protocol for the main-text trial-count comparison}
\label{app:aggregation-curves}
\Cref{fig:aggregation-main}(a) uses the same five saved forecasts for every
curve: \texttt{pro-high-brave-c0-t1}, with 400 question--date instances and
791 resolution events. Each horizon uses its own saved forecast when available.
For each $K$, we enumerate all $\binom{5}{K}$ trial subsets, average squared
error across subsets for each event, then convert to BI. Overall BI equally
weights Market and Data. These are expected losses within a fixed trial pool,
not independent new rollouts at each $K$.

For shrinkage, probabilities are clipped to $[0.001,0.999]$ before taking
logits. We select $f\in\{0,0.1,\ldots,1\}$ and
$c\in\{0,0.2,\ldots,2\}$ by MS on all other question--date instances,
excluding every horizon and trial subset of the held-out instance.
Other dates of the same underlying question can remain in training.
At $K=1$, all methods return the original forecast; no fitted calibration is
applied. Intervals use 5,000 paired question-cluster resamples (seed 2709),
stratified by Market/Data and retaining all horizons. Fits remain fixed in
resampling, so uncertainty is conditional on the fitted forecasts.
The analysis is in \texttt{analysis/aggregation\_curves.py} and its JSON output.
}

\clearpage
\section{ANOVA and Paired Analysis}
\label{app:stats}

\subsection{Variance Decomposition}
\label{app:anova}

We perform a two-way ANOVA (method $\times$ question)
on trial-level BI scores pooled across 12 historical configurations.
This is a descriptive analysis of the original experiment pool, separate
from the matched belief and memory contrasts in \cref{tab:component-effects}.
Each observation is $\BI_{ijt}$ for method $i$, question $j$, trial $t$.
The total sum of squares is decomposed as:
\[
  \text{SS}_{\text{total}} = \text{SS}_{\text{method}} + \text{SS}_{\text{question}} + \text{SS}_{\text{residual}}
\]
where $\text{SS}_{\text{method}} = \sum_i n_i (\bar{y}_{i\cdot} - \bar{y})^2$
sums over the 12 methods,
$\text{SS}_{\text{question}} = \sum_j n_j (\bar{y}_{\cdot j} - \bar{y})^2$
sums over the 400 questions,
and the residual includes inter-trial variability and any method--question
interaction not captured by the additive model.

The $F$-statistic tests whether the group means differ significantly:
$F = \text{MS}_{\text{effect}} / \text{MS}_{\text{residual}}$,
where $\text{MS} = \text{SS}/\text{df}$ is the mean square.
A large $F$ (relative to the $F$-distribution with appropriate degrees
of freedom) indicates the effect is significant.

\begin{table}[h]
\centering
\small
\caption{
  \small
  Two-way ANOVA on trial-level BI scores
  (FB A$\cup$B, 12 methods $\times$ $\nqAB$ questions $\times$ 5 trials).
  A perfectly balanced design would yield $12 \times 400 \times 5 = 24{,}000$
  observations; the residual df of 23{,}442 implies $N{=}23{,}853$ actual
  trials, i.e.\ 147 (0.6\%) were dropped because the agent hit the
  per-question timeout or returned no parseable probability
  (see \cref{app:aggregation} for how trial-level dead/missing values are
  handled in the aggregation step). 
}
\label{tab:anova}
\begin{tabular}{@{}l rrrr@{}}
\toprule
Source & SS & \% total & df & $F$ \\
\midrule
Method ($M{=}12$)   & 38.1  & 2.7\%  & 11 & 164.5 \\
Question ($Q{=}400$) & 877.6 & 62.2\% & 399 & 104.3 \\
Residual           & 494.3 & 35.1\% & 23{,}442 & --- \\
\midrule
Total              & 1{,}410.1 & 100\% & 23{,}852 & \\
\bottomrule
\end{tabular}
\end{table}

\begin{figure}[h]
  \centering
  \includegraphics[width=\textwidth]{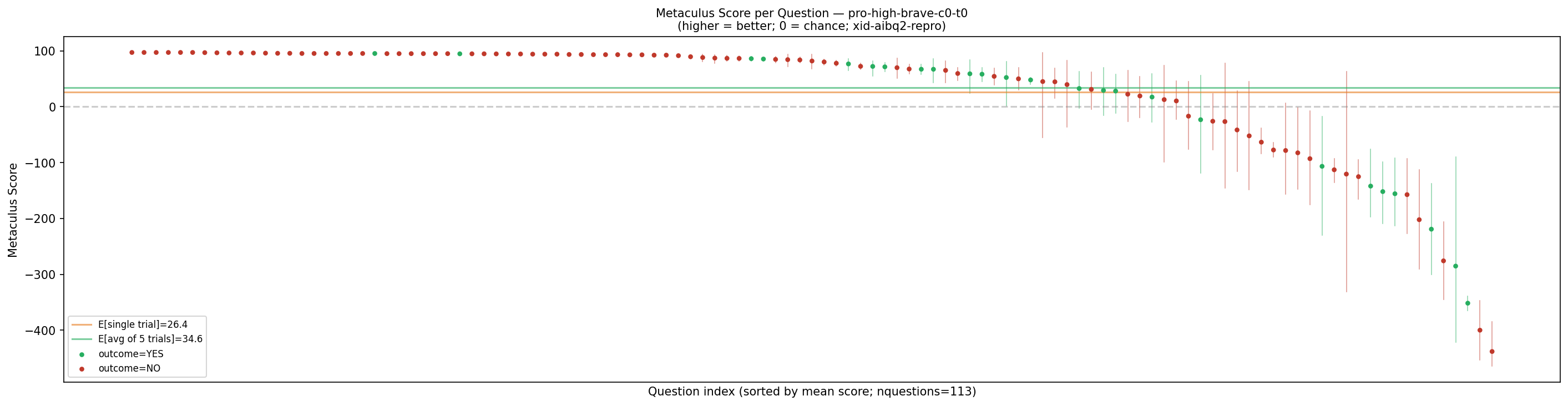}\\
  \includegraphics[width=\textwidth]{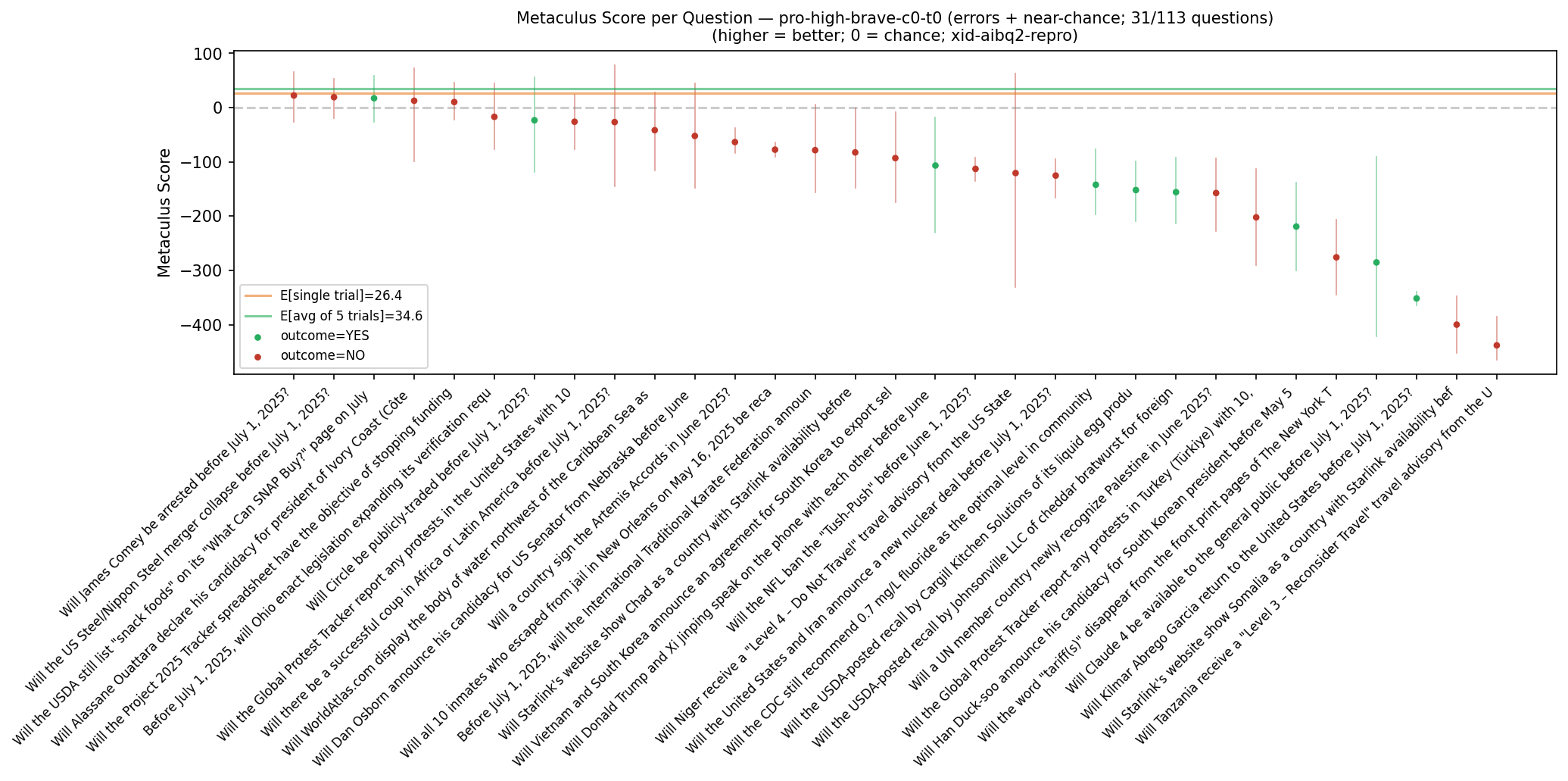}
  \caption{Top: Metaculus Baseline Score
  for all $n=113$ AIBQ2 questions (most are easy).
  Bottom: Zoom into questions at or below chance.
  Error bars: bootstrap 95\% CIs across 5 trials.
  }
  \label{fig:aibq2_ms_per_que}
\end{figure}

From \cref{tab:anova}, we see that
question difficulty dominates (62.2\% of total variance),
consistent with the observation that some questions are
inherently easy (e.g., ``Will a vaccine for Sepsis exist
by next month?'') while others are genuinely hard
(e.g., stock price movements),
cf. \citep{promptToLeaderboard}.
See also \cref{fig:aibq2_ms_per_que},
which shows that on AIBQ2, most questions are very easy,
so performance differences are dominated by a subset of hard questions.

Method effects account for only 2.7\% of total variance,
but the omnibus effect is significant ($F=164.5$, $p\ll0.001$).
This does not establish significance for every pairwise contrast.
The residual (35.1\%) includes inter-trial stochasticity and
unmodeled method--question interactions; it is not a pure estimate of
LLM sampling variability.

\subsection{Paired Analysis}
\label{app:mixed-effects}

For interpreting paired comparisons, consider an additive two-way linear model
\begin{equation}
  \BI_{ij} = \mu + \alpha_i + \gamma_j + \epsilon_{ij}
  \label{eq:mixed}
\end{equation}
where $\mu$ is the grand mean,
$\alpha_i$ is the effect of method $i$
(with $\sum_i \alpha_i = 0$),
$\gamma_j$ is the effect of question $j$
(with $\sum_j \gamma_j = 0$),
and $\epsilon_{ij}$ is the residual.

We consider both a fixed-effects  (FE) estimation method,
that computes point estimates (MLEs) of all parameters,
and a mixed-effects (ME) estimation method,
that treats the question effects as random,
$\gamma_j \sim \mathcal{N}(0,\sigma^2_q)$.
The ME method is statistically preferable,
since it accounts for uncertainty
  in $\gamma_j$ and enables inference on new questions,
  but requires iterative algorithms such as REML.
Under exact balance the fixed-effects and mixed-effects
  estimates for $\alpha_i$ coincide (a standard result on
  orthogonal designs); under approximate balance they differ
  by an amount on the order of the imbalance fraction,
  which in our case is very small, as we 
  showed in \cref{tab:anova}.

\paragraph{FE estimation via alternating projections.}
We estimate $(\mu, \{\alpha_i\}, \{\gamma_j\})$
using an alternating least-squares (ALS) algorithm,
which iterates:
\begin{align}
  \gamma_j &\gets \frac{1}{|\mathcal{I}_j|}\sum_{i \in \mathcal{I}_j}
    (\BI_{ij} - \mu - \alpha_i) \label{eq:als-q} \\
  \alpha_i &\gets \frac{1}{|\mathcal{J}_i|}\sum_{j \in \mathcal{J}_i}
    (\BI_{ij} - \mu - \gamma_j) \label{eq:als-m}
\end{align}
where $\mathcal{I}_j$ is the set of methods that answered
question $j$, and $\mathcal{J}_i$ is the set of questions
answered by method $i$.
This is equivalent to coordinate descent on the
least-squares objective
$\sum_{ij} (\BI_{ij} - \mu - \alpha_i - \gamma_j)^2$,
which is convex and separable in the two sets of parameters,
so iteration converges to a global minimiser of the fit.

The minimiser is not unique in $(\alpha, \gamma)$ alone --- adding any
constant $c$ to every $\alpha_i$ and subtracting $c$ from every
$\gamma_j$ leaves the fit unchanged --- so we impose the
identifying sum-to-zero constraints
$\sum_i \alpha_i = 0$ and $\sum_j \gamma_j = 0$
already stated in \cref{eq:mixed} by recentering after each
sweep ($\alpha_i \!\gets\! \alpha_i - \bar\alpha$,
$\gamma_j \!\gets\! \gamma_j - \bar\gamma$,
$\mu \!\gets\! \mu + \bar\alpha + \bar\gamma$); this
fixes a unique parameter configuration.

In our setting, every method attempts every question and the
0.6\% dropped trials  are spread thinly across
methods, so the design is balanced for practical purposes.
Under exact balance, the effects reduce to centered marginal means.
With missing trials, we use the iterative fit rather than assuming that
one sweep yields the exact solution.

\paragraph{Bootstrap analysis of pairwise differences.}
For the reported BI contrasts we pair methods on common resolution events,
recompute each method's BI from its mean squared error, and take the
difference. On ForecastBench, Overall BI is the equal-weight mean of
Market and Data BI. This nonlinear statistic is not the mean of per-event
BI differences. The resampling unit is stated with each analysis:
the external comparisons use paired events, while belief, memory, and
trial-count analyses resample questions and retain their resolution horizons.
Pairing controls differences in the scored sample, but does not by itself
remove confounding from model, tools, or compute budgets. Event resampling
does not capture within-question dependence across horizons; resampling
question--date instances also leaves cross-date dependence unmodeled.

Percentile intervals are obtained from the paired bootstrap distribution.
Where stars are reported, the two-sided bootstrap tail estimate is twice
the smaller fraction of differences at or below zero and at or above zero,
capped at one. Tests are exploratory and not adjusted for multiple comparisons.
Intervals for fitted post-processing are conditional on the fits unless
otherwise stated; they do not include a full nested refitting analysis.

\clearpage
\section{Datasets and their analysis}
\label{app:data-viz}
\label{app:datasets}

\subsection{AIBQ2}
\label{app:aibq2-data}
\label{app:AIBQ2}

\begin{figure}[h]
  \centering
  \includegraphics[width=0.8\textwidth]{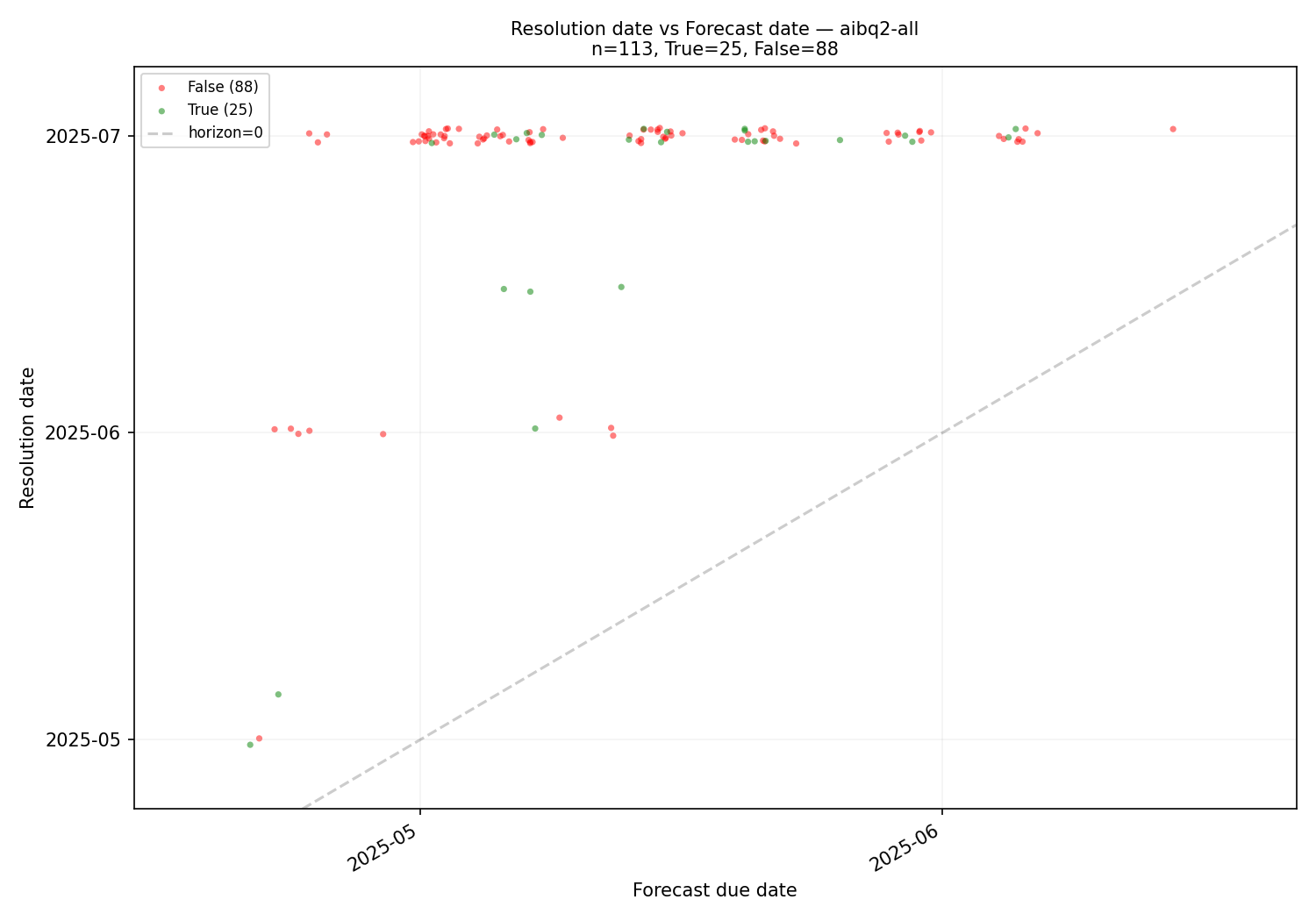}
  \caption{AIBQ2 date distribution of questions and answers.
    Green = resolved True (25), Red = resolved False (88).}
  \label{fig:aibq2-scatter}
\end{figure}

\begin{figure}[h]
  \centering
  \includegraphics[width=\textwidth]{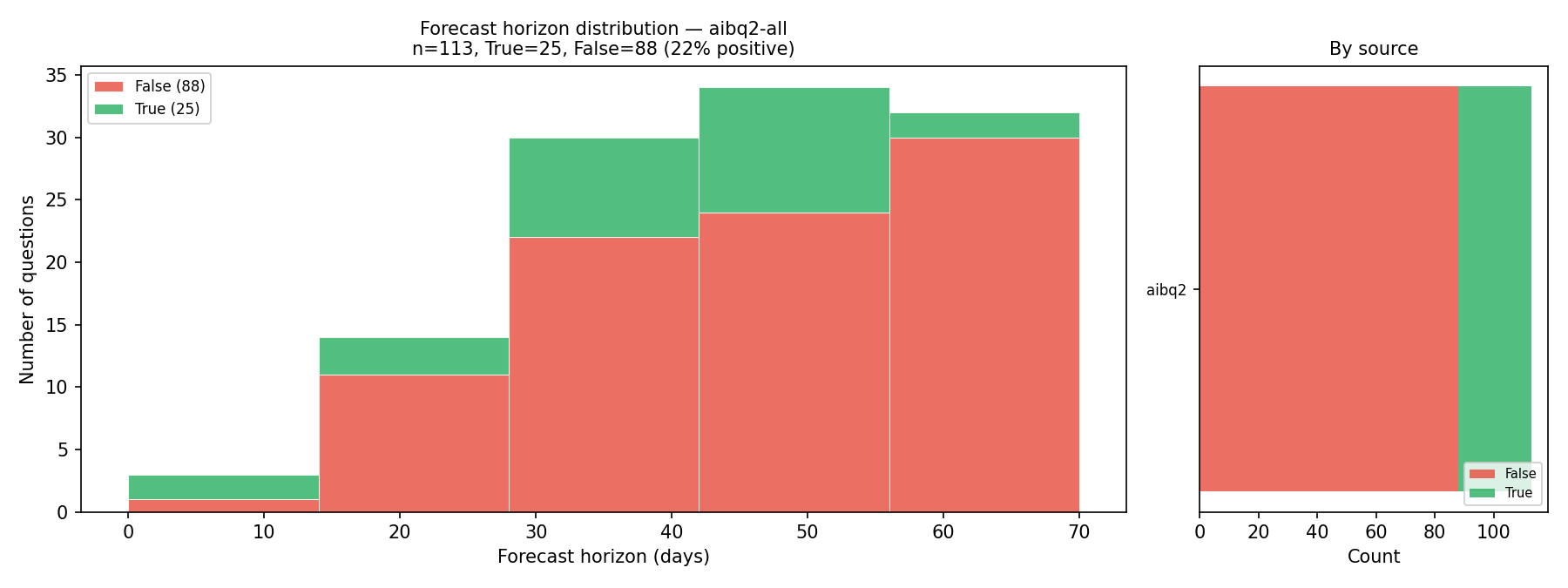}
  \caption{AIBQ2 forecast horizon distribution ($n=113$).
     Right panel shows
    outcome balance.
  }
  \label{fig:aibq2-horizon}
\end{figure}

The AIBQ2 benchmark consists of 113 binary questions from
the Metaculus AI Benchmark Tournament (Q2 2025).
\footnote{
  See \url{https://www.metaculus.com/tournament/aibq2/}.
The full list of questions and outcomes is available at
\url{https://gist.github.com/enjeeneer/86e24a52e6041a3d78e333bcab16984d},
taken from \citep{scott2026forecasting}.
}
Questions were asked between 2025-04-22 and 2025-06-15,
with resolution dates up to 2025-07-01.
The base rate is 22\% (25 True, 88 False).
Questions do not have a market or crowd base rate.
Figure~\ref{fig:aibq2-scatter} shows the forecast horizon
distribution: most questions have short horizons (1--8 weeks),
with many resolving on 2025-07-01 (the tournament deadline).
See also \cref{fig:aibq2-horizon} for another visualization of this data.
Table~\ref{tab:aibq2-tag-examples} shows some example questions,
which we have clustered into 10 manually chosen topics.
(The clustering is done  by applying an LLM classifier to each question,
and then grouping questions with the same class label or tag into
the same cluster).
We see that the most common topics are
Domestic Politics (35/113) and Geopolitics \& Conflict (29/113).
By contrast, the FB topic distribution is quite different
(see \cref{fig:fb-categories}).

\begin{table}[h]
  \centering
  \small
  \caption{AIBQ2 topic distribution with example questions
    across 10 categories ($n=113$).
    Domestic Politics dominates (35 questions).}
  \label{tab:aibq2-tag-examples}
\begin{tabular}{@{}lrl@{}}
\toprule
Category & $n$ & Example question \\
\midrule
Domestic Politics & 35 & Will a state of emergency be in effect in Samoa on April 30, 2025? \\
Geopolitics \& Conflict & 29 & Will there be a successful coup in Africa or Latin America before Jul\dots \\
Business \& Industry & 16 & Will Uber be available in the Turks and Caicos Islands on July 1, 2025? \\
Society \& Law & 9 & Will Harvard University lose its tax exempt status before July 1, 2025? \\
Macroeconomics & 7 & Will South Sudan ratify AfCFTA (the pan-African free trade agreement)\dots \\
Science \& Technology & 6 & Will Starlink's website show Lesotho as a country with Starlink avail\dots \\
Health \& Biology & 5 & Will the CDC raise the Travel Health Notice status of Colombia to abo\dots \\
Financial Markets & 2 & Will any of these companies get dropped from the Dow Jones Industrial\dots \\
Other & 2 & Will WorldAtlas.com display the body of water northwest of the Caribb\dots \\
Sports \& Entertainment & 2 & Before July 1, 2025, will the International Traditional Karate Federa\dots \\
\midrule
Total & 113 & \\
\bottomrule
\end{tabular}
\end{table}

\subsection{ForecastBench}
\label{app:fb-data}
\label{app:FB}
\label{app:datasets-construction}

In this section, we discuss the 
ForecastBench dataset from \citep{forecastbench}.
This consists of
binary prediction questions from market
sources (Polymarket, Manifold, Metaculus, Rand Forecasting Initiative (RFI)),
which  assess the ability to do ``judgemental forecasting''
\citep{lawrence2006} on various topics for various forecast horizons.
We can formalize this as follows.
Let $Y(t)$ denote the random variable of interest (e.g.,
``Will A be president of country B at time $t$?'').
Let $f$ be the forecast date and $r$ the resolution date.
Our task is to estimate
$P(Y(r)=1 \mid \text{data}(\leq f))$.

FB also contains
 questions derived from dataset sources
(yfinance, FRED, DBnomics, Wikipedia, ACLED), which
 assess the ability to do (univariate) time series
forecasting.
These are converted to binary prediction problems by asking the forecaster
what the probability will be
that the value of interest
at multiple future resolution dates, out to a horizon of
$h \in \{7, 30, 90, 180, 365, 1095, 1825, 3650\}$ days,
will be larger or smaller
(possibly by some factor)
compared to  its value at the forecast date.
More formally we are required to estimate
$P(Y(r_i) > v \mid \text{data}(\leq f))$,
where  $r_i = f + h_i$ is the resolution date for the $i$'th forecast,
for $i=1:8$,
$Y(r_i)$ is the unknown future value,
and $v = y(f)$ is the known threshold or reference value.
For example, $Y(r_i)>v$ might represent the event
``the temperature at location X on date $r_i$ will
be higher than its current value $v$''.
(Note that the value $v$ is not always given in the question,
a subtlety we discuss in \cref{app:tools}.)

500 new FB questions are released every two weeks (250 market and 250 dataset),
and these resolve over time at different rates.

The combined tranches contain 400 question-date instances but 355 distinct
underlying (source, question) IDs. Date-specific instances can therefore be
dependent even when treated as separate questions in an analysis.

\subsubsection{Tranches}
\label{sec:tranches}

\begin{figure}[h]
  \centering
  \includegraphics[width=\textwidth]{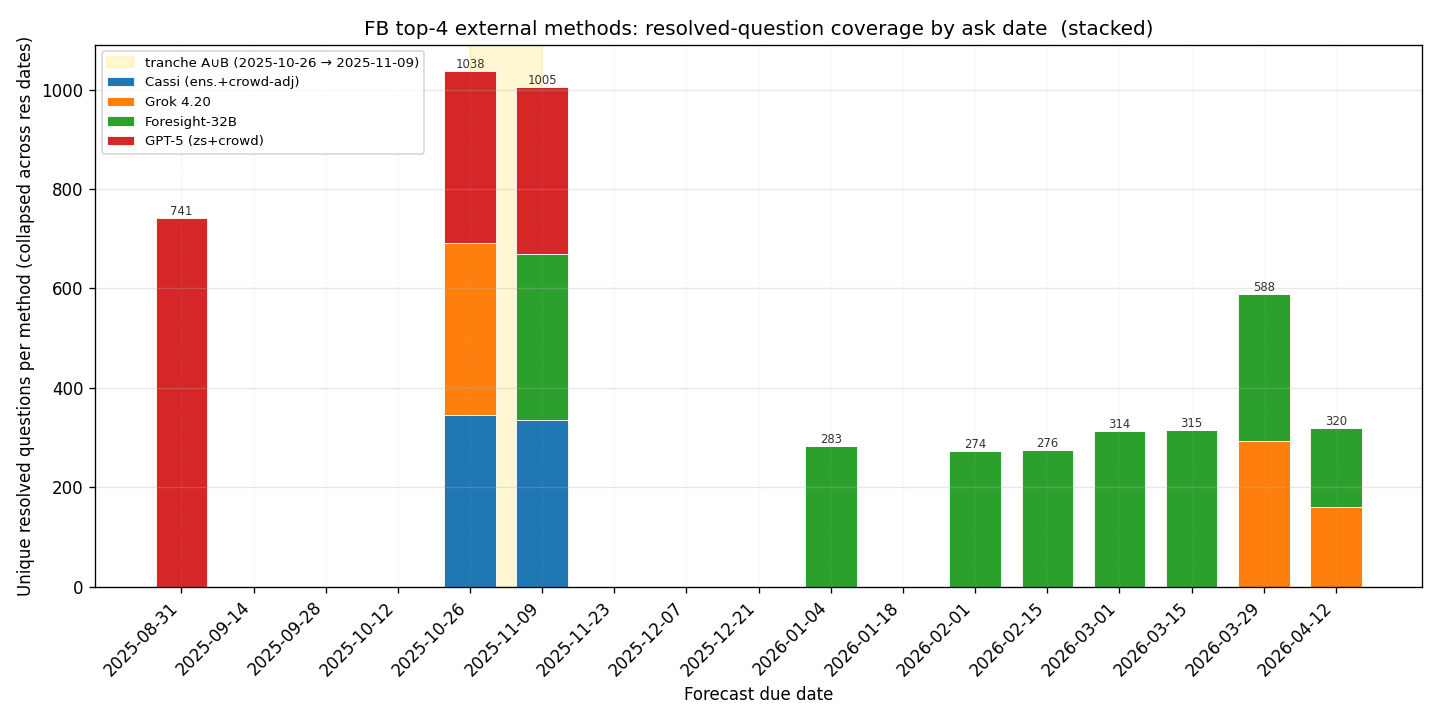}
  \caption{
    Number of resolved questions by the top external methods across time.
    The golden region corresponds to tranches A and B that we use.
  }
  \label{fig:method-coverage}
\end{figure}

We construct two evaluation datasets (which we call ``tranches'')
from ForecastBench,
selecting dates that satisfied several criteria:
after the knowledge cutoff of all models of interest (2025-08-31),
has many resolved values (especially important for market questions, which resolve less often),
and maximizes overlap with the submissions from existing leading methods
(see \cref{fig:method-coverage}).
(See \cref{app:alternatives}  for details on these existing methods which we compare to.)
Each tranche uses questions from a single \texttt{forecast\_due\_date}
and selects those which are resolved by 2026-04-10 (ensuring ground
truth is available).
 Of the $\sim$500
questions per forecast date, approximately 100 market and 200
dataset questions had resolved by 2026-04-10; we select all 100
market questions and 100 dataset questions (20 per source, via
stratified sampling with a fixed seed) for each tranche,
to ensure neither dominates the performance metrics.
(Note that each dataset question has multiple resolution dates,
so the total number of binary questions with answers is $n{=}\nrdAB$.)
More precisely, we use these tranches:

\begin{itemize}
  \item \textbf{Tranche~A} (forecast\_due\_date = 2025-10-26):
    100 market + 100 dataset questions (200 total).
    Comparisons available: Cassi, xAI Grok~4.20, GPT-5.
  \item \textbf{Tranche~B} (forecast\_due\_date = 2025-11-09):
    100 market + 100 dataset questions (200 total).
    Comparisons available: Cassi, Foresight-32B, GPT-5.
\end{itemize}

\noindent The combined dataset (\textbf{Tranche~A$\cup$B}, $n=400$)
is used for the main comparison. Paired statistical tests are
conducted within each tranche (where all methods forecast the same
questions).
Note that for market questions (but not dataset/timeseries questions),
ForecastBench includes a crowd estimate,
based on the market price for Polymarket, Manifold, and RFI,
and the Community Prediction score for Metaculus.
This is a very strong baseline, and most methods on the leaderboard rely
on it quite heavily.

\eat{
\begin{figure}[h]
  \centering
  \includegraphics[width=\textwidth]{figs/per_date_resolved.png}
  \caption{
    Number of questions per source with resolutions, as of 2026-03-31.
  }
  \label{fig:per-date-resolved}
\end{figure}
}

\subsubsection{Statistical properties of the tranches}

\begin{table}[h]
\centering
\small
\caption{
  \small
  Question composition by source for FB Tranches~A and~B.
  ``BR'' = base rate (fraction resolving True).
  Market sources provide judgemental forecasting questions;
  dataset sources provide univariate time-series questions
  with multiple resolution dates per question.
  ACLED = Armed Conflict Location and Event Data.
 FRED = Federal Reserve Economic Data.
 DBnomics = database of international economic statistics.
}
\label{tab:source-composition}
\setlength{\tabcolsep}{4pt}
\begin{tabular}{@{}llrrrrr@{}}
\toprule
Source & URL & $n_A$ & $n_B$ & $n$ & BR(A) & BR(B) \\
\midrule
RFI & randforecastinginitiative.org & 5 & 5 & 10 & 0.00 & 0.00 \\
Manifold & manifold.markets & 18 & 26 & 44 & 0.22 & 0.15 \\
Metaculus & metaculus.com & 10 & 10 & 20 & 0.10 & 0.20 \\
Polymarket & polymarket.com & 67 & 59 & 126 & 0.13 & 0.08 \\
\cmidrule{1-7}
\emph{Market total} & & 100 & 100 & 200 & 0.14 & 0.11 \\
\midrule
ACLED & acleddata.com & 20 & 20 & 40 & 0.25 & 0.00 \\
DBnomics & db.nomics.world & 20 & 20 & 40 & 0.95 & 0.80 \\
FRED & fred.stlouisfed.org & 20 & 20 & 40 & 0.60 & 0.55 \\
Wikipedia & wikipedia.org & 20 & 20 & 40 & 0.20 & 0.25 \\
Yahoo Finance & finance.yahoo.com & 20 & 20 & 40 & 0.20 & 0.45 \\
\cmidrule{1-7}
\emph{Dataset total} & & 100 & 100 & 200 & 0.44 & 0.41 \\
\midrule
\textbf{Total} & & 200 & 200 & 400 & 0.29 & 0.26 \\
\bottomrule
\end{tabular}
\end{table}

\begin{figure}[h]
  \centering
  \includegraphics[width=1.0\textwidth]{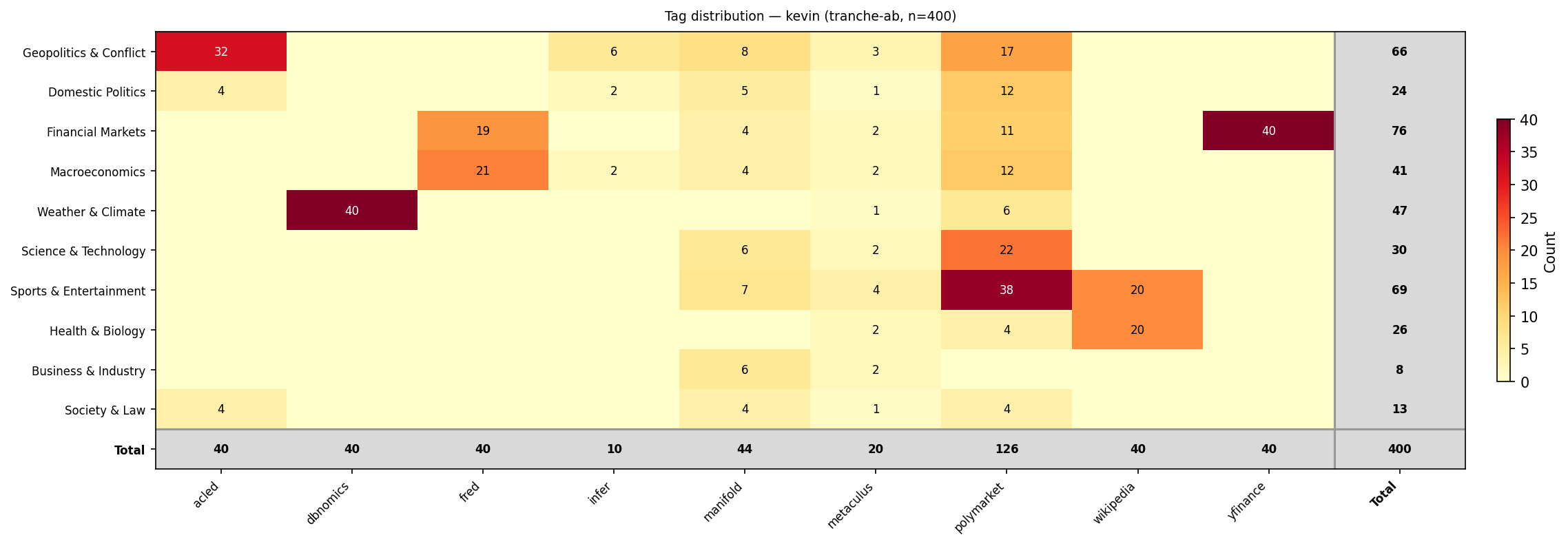}
  \caption{FB A$\cup$B topic distribution across 9 sources.}
  \label{fig:fb-categories}
\end{figure}

In this section, we discuss some properties of the FB dataset.
\Cref{tab:source-composition} shows some key statistics.
\Cref{fig:fb-categories} shows the topic distribution:
Financial Markets (76) leads, followed by
Sports \& Entertainment (69) and Geopolitics \& Conflict (66),
with the remaining 7 categories more evenly distributed.

Figure~\ref{fig:bi-by-source} shows
per-source BI for four methods.
Performance of our best system (\system{}+crowd+emp+cal)
varies dramatically across sources:
Wikipedia (\bi{83.0}) and Manifold (\bi{90.7}) are near-perfect,
while yfinance (\bi{49.6}) is essentially chance.
FRED (\bi{53.2}) is the second-weakest source.

\label{app:fb-market}

\begin{figure}[h]
  \centering
  \includegraphics[width=\textwidth]{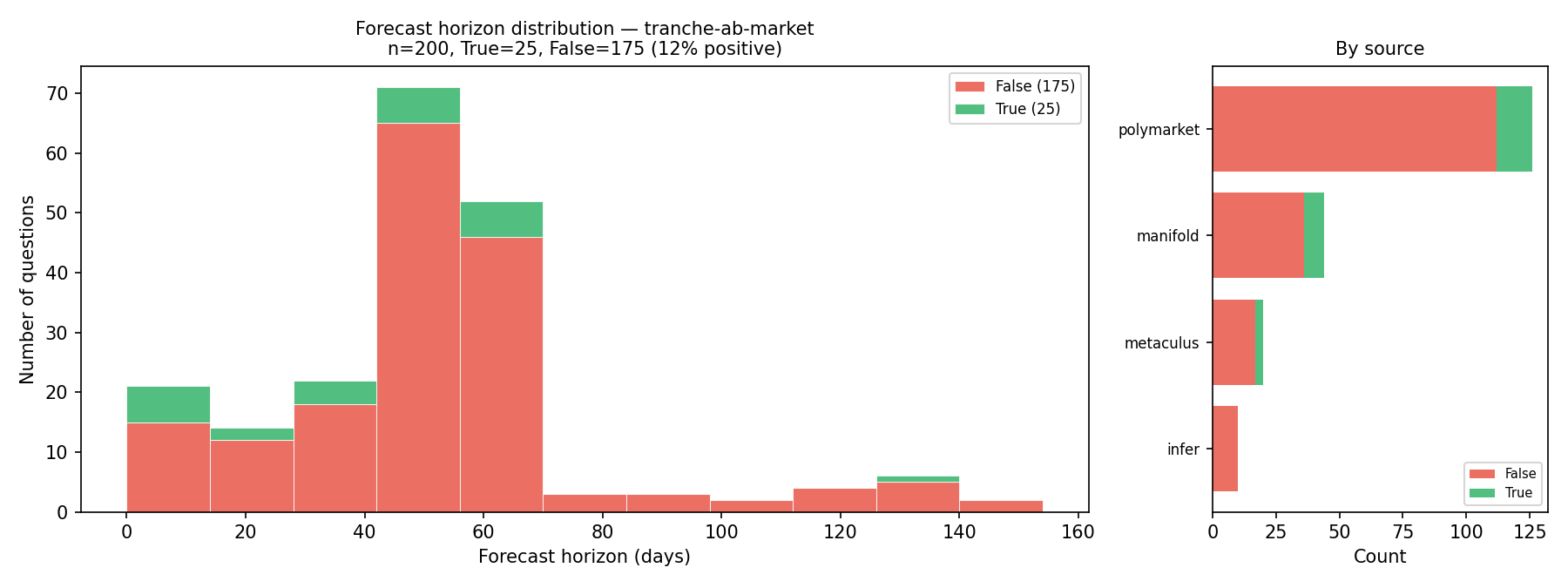}
  \caption{FB market questions: forecast horizon distribution
    ($n=200$). Most horizons are 1--5 months.
    Base rate is 12\% (heavily skewed toward False).}
  \label{fig:fb-horizon-market}
\end{figure}

Market questions (Polymarket, Manifold, Metaculus, RFI) are
judgemental forecasting questions similar in nature to AIBQ2
(see Table~\ref{tab:aibq2-tag-examples} for example topics),
but spanning a wider range of sources and with a different
topic distribution (Figure~\ref{fig:fb-categories}).
Forecast horizons range from 1 week to 5 months
(Figure~\ref{fig:fb-horizon-market}).

The base rates are very low (Table~\ref{tab:source-composition}):
Polymarket 11\%, Manifold 18\%, Metaculus 15\%, RFI 0\%.
This means most questions resolve ``No,''.
Furthermore, these questions come with a market estimate;
using this as the predicted probability achieves high market BI
--- a strong baseline that is very hard to beat.
The belief-only comparison in \cref{tab:component-effects} isolates
market gains for Pro and Flash without calibration.

\label{app:fb-dataset}

\begin{figure}[h]
  \centering
  \includegraphics[width=\textwidth]{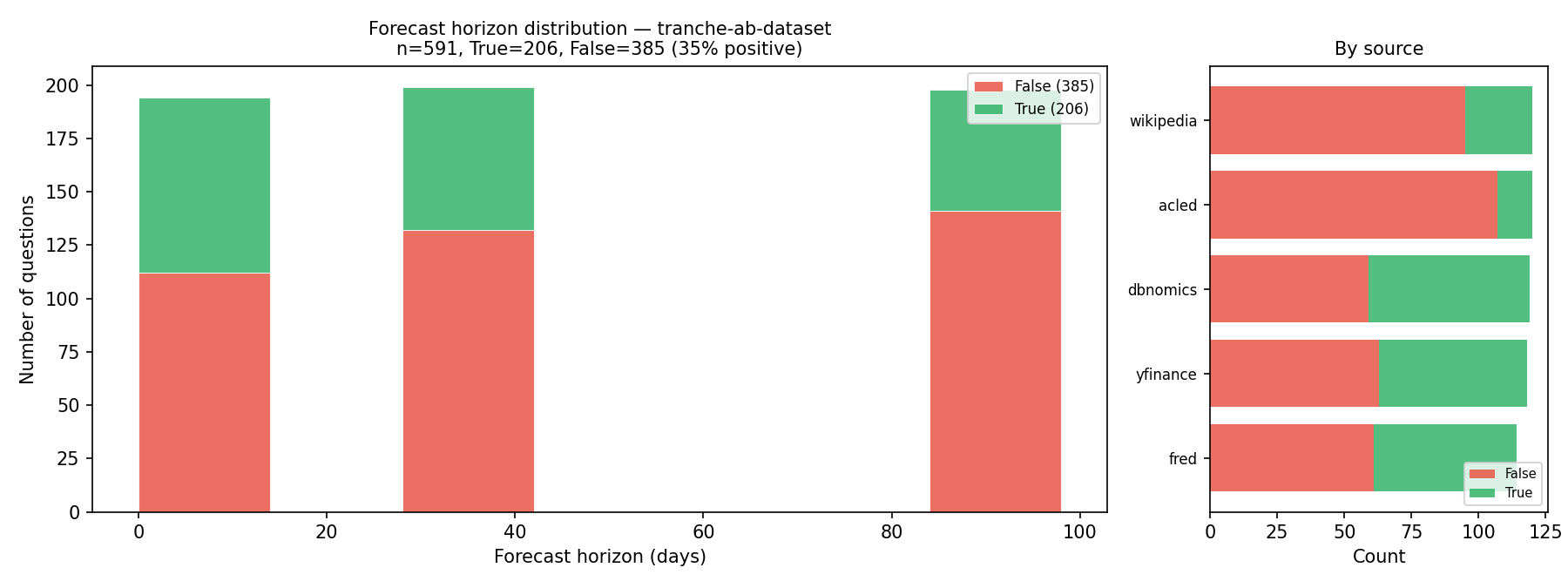}
  \caption{FB dataset questions: forecast horizon distribution
    ($n=200$ questions, 591 resolution dates).
    Horizons are fixed at 7, 30, and 90 days
    (only the first 3 of 8 standard resolution dates have
    resolved by our evaluation cutoff).
    Base rate is 35\%.}
  \label{fig:fb-horizon-dataset}
\end{figure}

Dataset questions are univariate time series forecasting problems,
which are converted to binary form by asking whether  (some function of)
the value of a quantity of interest at a future resolution
date will be higher or lower than the value at the forecast date.
Topics are very narrow, and depend on the source, as shown in 
Figure~\ref{fig:fb-categories}.
Forecast horizons are fixed at 7, 30, and 90 days,
as shown in 
Figure~\ref{fig:fb-horizon-dataset}.
We give more details on individual dataset sources below.

\subsubsection{Representativeness of the tranches}
\label{app:representativeness}

After completing the main experiments, we extended our evaluation
by running our SOTA model
 on every FB question that the agent could ever have
forecasted since after Gemini knowledge cutoff
($n{=}23{,}150$ resolved (source, question, resolution-date)
entries spanning 14 ask dates from 2025-03 through 2026-04).
This allowed us to estimate how representative the A,B tranches are.

\begin{figure}[h]
  \centering
  \includegraphics[width=\textwidth]{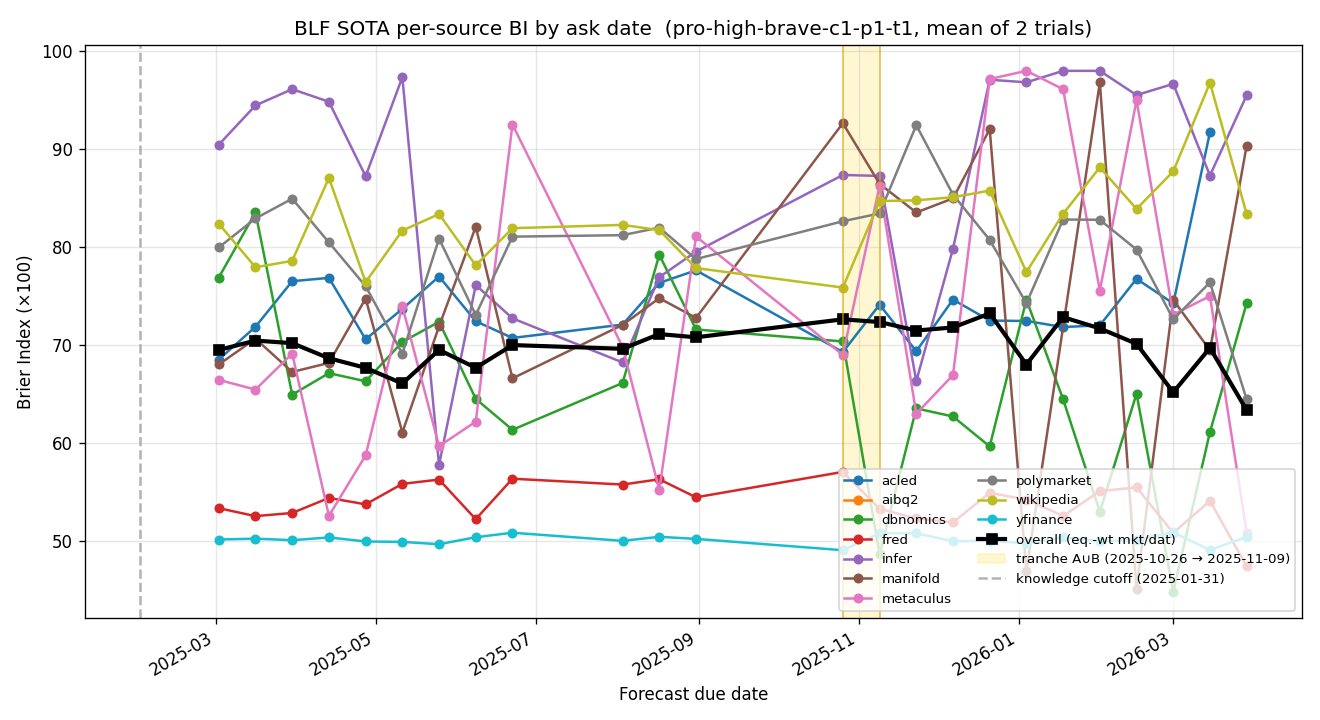}
  \caption{
    Brier Index of our method across all resolved questions since after the Gemini
    knowledge cutoff.
    Gold band represents tranches A and B.
  }
  \label{fig:per-date-bi}
\end{figure}

As we see from \cref{fig:per-date-bi},
the tranche window sits in the middle of a fairly stable region of the BI timeline,
although the overall BI level ($\sim$72) is slightly higher than the wider corpus mean ($\sim$69).
The difference is mostly 
due to the market questions in the tranches being easier (+6.5 BI),
driven mostly by differences in the manifold/infer/metaculus  sources.
However, the dataset questions in the tranches were slightly harder (-1.2 BI),
and the net effect (after averaging over all sources) is that the BI only differs by about 3. 
This suggests our tranche BI numbers may be higher than what one would expect
on the overall FB leaderboard.
However, we expect this bias to affect other methods in the same way.

\subsection{FB: Yahoo Finance (stock prices)}
\label{app:fb-yfinance}

\begin{figure}[h]
  \centering
  \includegraphics[width=\textwidth]{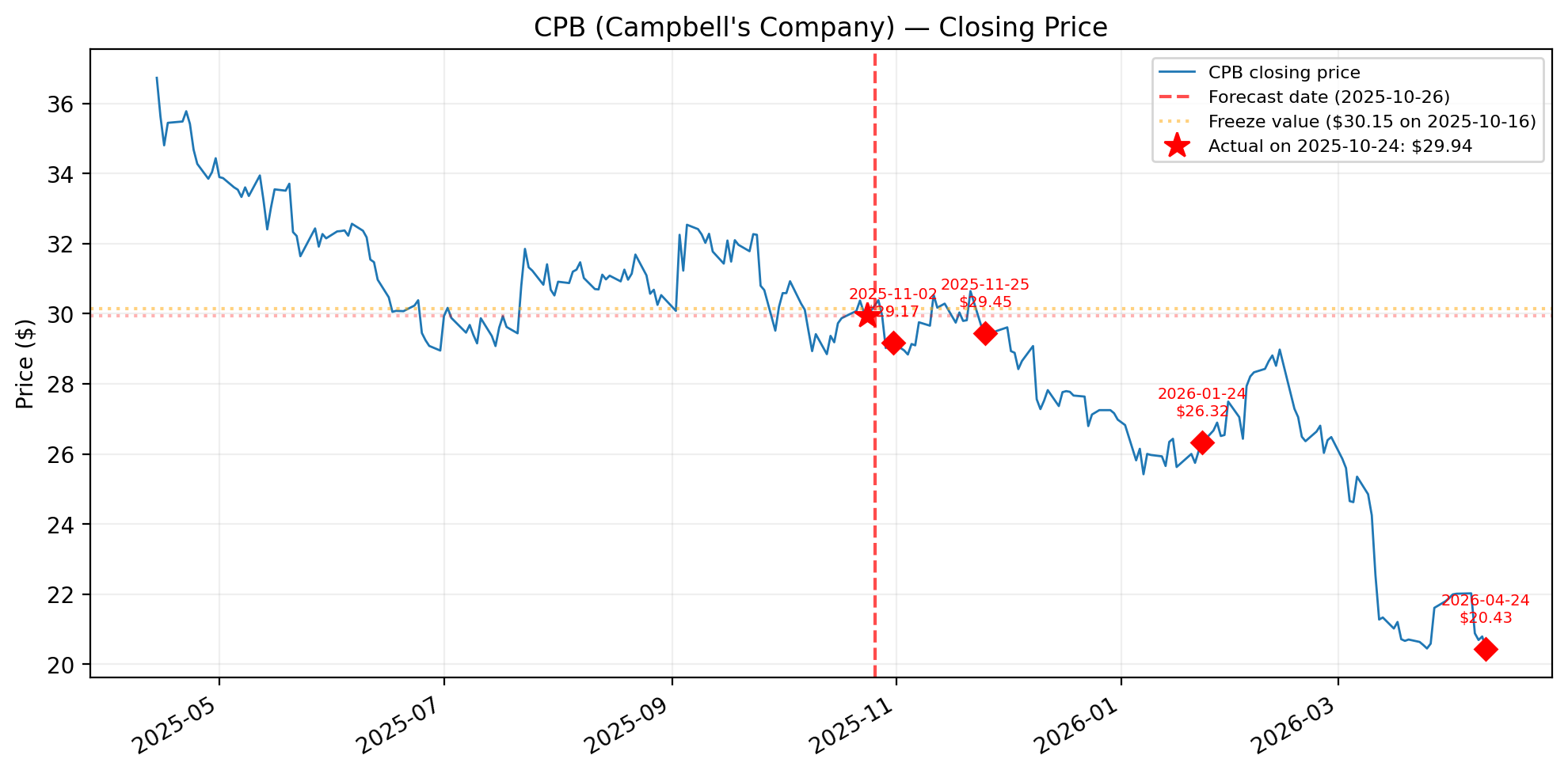}
  \caption{CPB (Campbell's Company) closing price.
    Red dashed line: forecast date (2025-10-26).
    Star: actual price on 2025-10-24 (\$29.94, since 10-26 was Sunday).
    Orange: freeze value (\$30.15 from 10-16).
    Red diamonds: resolution date prices (all below reference,
    all resolved False).
    The decline to $\sim$\$20 was unpredictable at forecast time.}
  \label{fig:yfinance-cpb}
\end{figure}

Yahoo Finance questions ask whether a stock's closing price on
a future resolution date will be higher than its closing price
on the forecast due date.
Each question has 8 resolution dates ranging from 1 week to
10 years out. Figure~\ref{fig:yfinance-cpb} shows an example.

The question text uses placeholders \verb|{resolution_date}|
and \verb|{forecast_due_date}| that are substituted at
preprocessing time. Crucially, the question does \emph{not}
specify the reference price on the forecast due date ---
it only provides a \texttt{freeze\_datetime\_value} from
an earlier date (when the question was created, typically
10 days before the forecast due date).
For example, the CPB question below has
\texttt{freeze\_datetime\_value} = 30.15 from 2025-10-16,
but the actual closing price on the forecast due date
(2025-10-24, since 10-26 is a Sunday) was 30.79.
Our system retrieves the correct reference value via the
\texttt{fetch\_ts\_yfinance} tool, which downloads the
price history and returns the closing price on or before
the forecast due date.

Figure~\ref{fig:yfinance-cpb} illustrates why yfinance
questions are inherently difficult: stock prices follow
an approximate random walk, making it nearly impossible to
predict whether the price will be higher or lower than the
reference value on any given future date.
The optimal strategy is to predict $p \approx 0.5$, which
our system correctly learns (yfinance BI $\approx 0.5$ across
both tranches; see Table~\ref{tab:source-composition}).

\paragraph{Example question (verbatim, abbreviated).}
\begin{small}
\begin{verbatim}
{
  "forecast_due_date": "2025-10-26",  // global
  "id": "CPB",
  "source": "yfinance",
  "question": "Will CPB's market close price on
    {resolution_date} be higher than its market
    close price on {forecast_due_date}?",
  "resolution_criteria": "Resolves to the market
    close price at https://finance.yahoo.com/quote/CPB.
    If the resolution date coincides with a day the
    market is closed, the previous close is used.",
  "freeze_datetime": "2025-10-16T00:00:00+00:00",
  "freeze_datetime_value": "30.15",
  "resolution_dates": ["2025-11-02", "2025-11-25",
    "2026-01-24", ...],
  "resolved_to": [0, 0, 0, ...]  // all False
}
\end{verbatim}
\end{small}

\subsection{FB: FRED (economic indicators)}
\label{app:fb-fred}

FRED (Federal Reserve Economic Data) questions ask whether an
economic indicator will have increased by each resolution date
compared to the forecast due date.
An example question is shown below:

\begin{quote}
\small
\textbf{Question (DTB6):} ``Will the Federal Reserve's 6-month
secondary market treasury bill rate have increased by
\texttt{\{resolution\_date\}} compared to its value on
\texttt{\{forecast\_due\_date\}}?''\\[2pt]
\textbf{Freeze value:} 3.69\% (as of 2025-10-16)\\
\textbf{Resolution dates:} 2025-11-02, 2025-11-25, 2026-01-24,
  2026-04-24, 2026-10-26, 2028-10-25, 2030-10-25, 2035-10-24
\end{quote}

\noindent
The 40 FRED questions in our tranches (\cref{tab:source-composition}) ask about
36 distinct economic series; the series span interest rates (DTB6, DGS30),
corporate bond spreads (BAMLC0A0CM), exchange rates (DEXUSUK),
stock indices (DJIA), and banking data (DPSACBW027SBOG).

\begin{figure}[h]
  \centering
  \includegraphics[width=\textwidth]{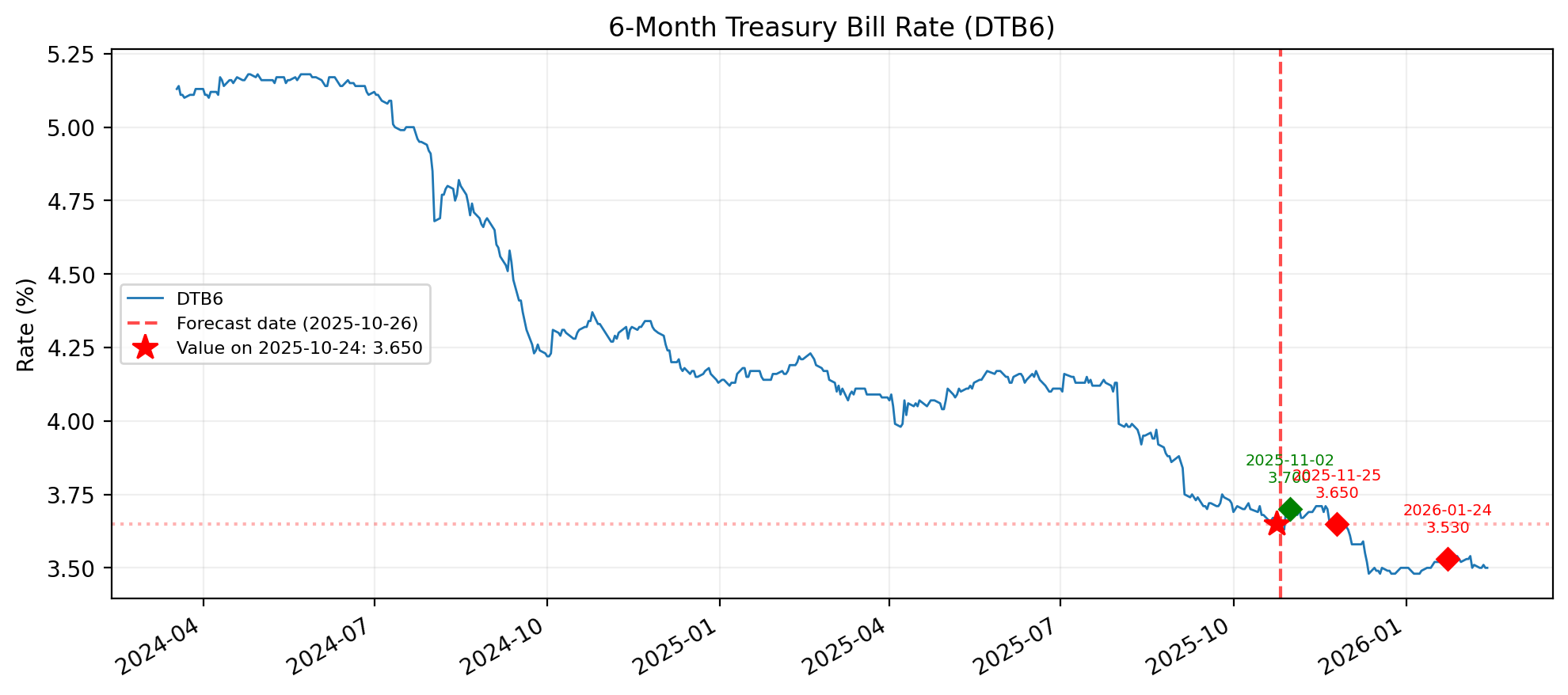}
  \caption{6-Month Treasury Bill Rate (DTB6).
    Forecast date is 2025-10-26 (Sunday), so the reference
    value is from 2025-10-24 (red star).
    Clear downward trend from Fed rate cuts.
    Green diamond = resolved True (rate increased);
    red = resolved False.}
  \label{fig:fred-dtb6}
\end{figure}

Unlike yfinance (approximate random walk), many FRED series
exhibit persistent trends. For example,
Figure~\ref{fig:fred-dtb6} shows the 6-month T-bill rate
declining steadily from $\sim$5.25\% to $\sim$3.5\% over
2024--2026, reflecting Fed rate cuts. The first resolution
date (2025-11-02) caught a brief uptick (resolved True),
but subsequent dates resolved False as the rate continued falling.
This trend is partially predictable from recent history,
giving our \texttt{fetch\_ts\_fred} tool useful context.

\begin{figure}[h]
  \centering
  \includegraphics[width=\textwidth]{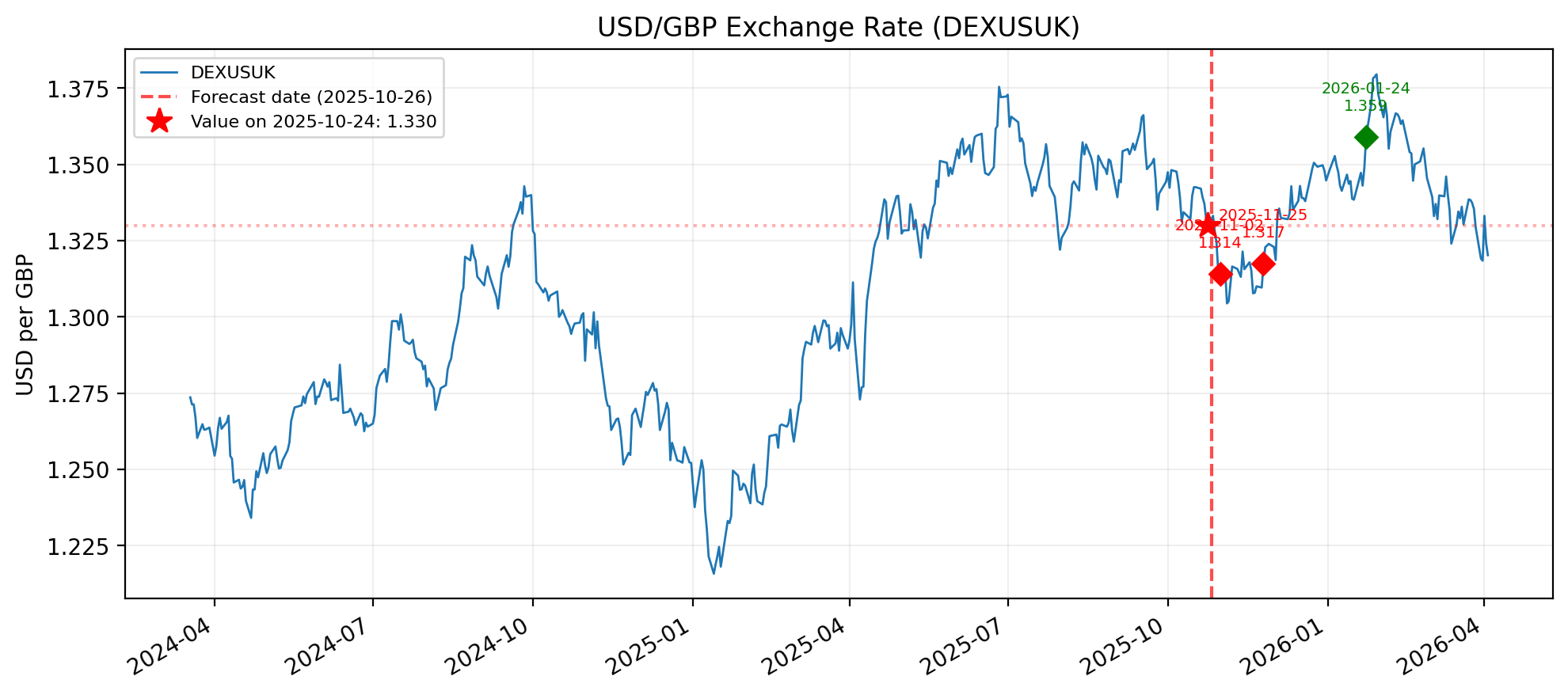}
  \caption{USD/GBP exchange rate (DEXUSUK).
    Reference value from 2025-10-24 (red star, since
    forecast date 2025-10-26 is a Sunday).
    More erratic than interest rates --- harder to predict.}
  \label{fig:fred-dexusuk}
\end{figure}

Figure~\ref{fig:fred-dexusuk} shows the USD/GBP exchange rate,
which is harder to predict --- more similar to a random walk,
with a mix of True and False resolutions.

Overall, FRED has mixed predictability: of our 36 distinct series,
roughly 8 exhibit strong autocorrelation (trending),
18 are effectively random walks, and the rest are intermediate.
Trend-following series (interest rates, banking aggregates)
reward tools that detect recent direction, while noisy series
(exchange rates, bond yields) are closer to chance.
The combined base rate for FRED is 57\%
(Table~\ref{tab:source-composition}), between DBnomics (88\%)
and yfinance (33\%).

A natural improvement would be to classify each FRED series
into a small number of patterns (random walk, trending,
mean-reverting) and apply the appropriate statistical model
--- e.g., $p = 0.5$ for random walks, a trend-extrapolation
model for persistent series.
ForecastBench's full question bank contains 165 unique FRED
series, so this per-series approach would require a lightweight
classifier, which we leave to future work.

\subsection{FB: DBnomics (temperature)}
\label{app:fb-dbnomics}

\begin{figure}[h]
  \centering
  \includegraphics[width=0.6\textwidth]{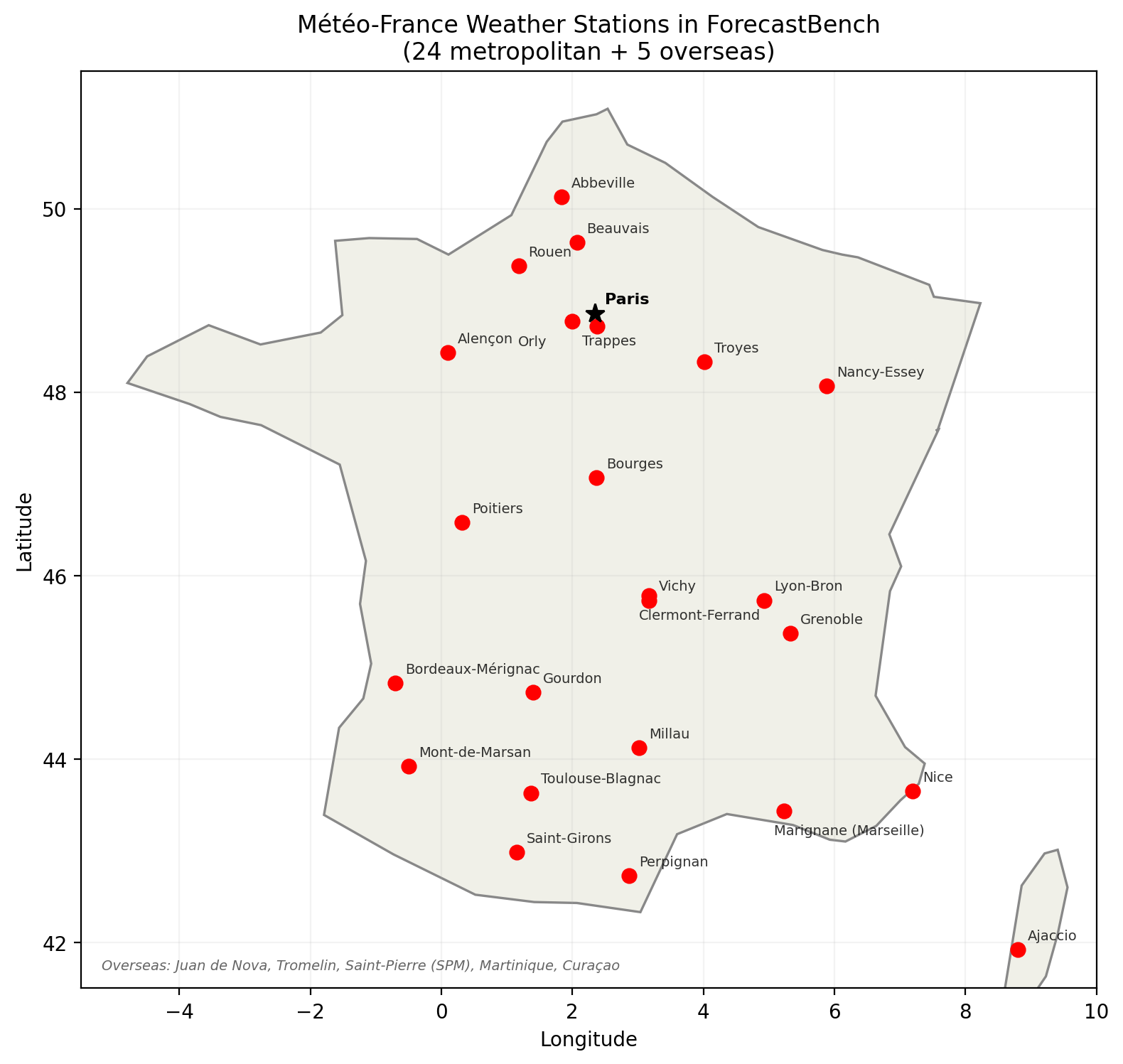}
  \caption{Météo-France weather stations used in ForecastBench.
    Stations are highly correlated (mean pairwise $r=0.93$).}
  \label{fig:dbnomics-map}
\end{figure}

\begin{figure}[h]
  \centering
  \includegraphics[width=\textwidth]{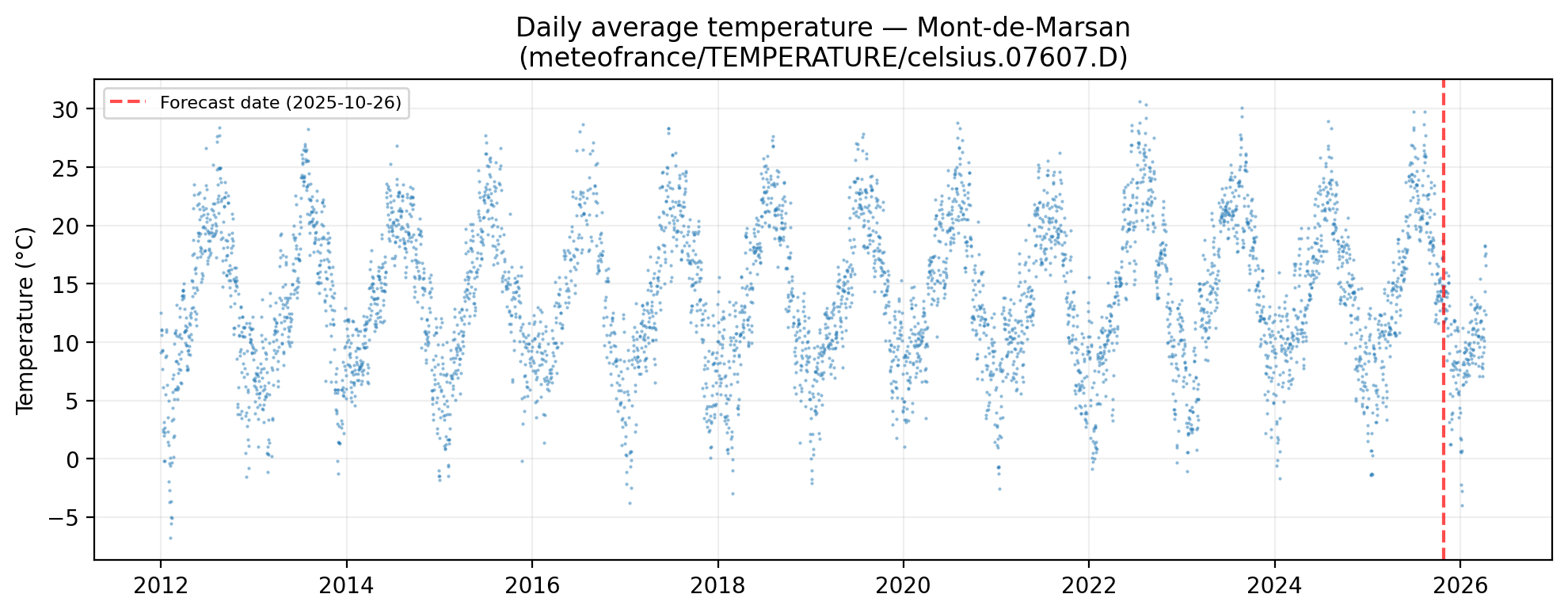}
  \caption{Daily average temperature at Mont-de-Marsan,
    2012--2025. Strong annual seasonality makes statistical
    forecasting effective.}
  \label{fig:dbnomics-full}
\end{figure}

All DBnomics questions in ForecastBench ask whether the daily
average temperature at a French weather station will be higher
on a future resolution date than on the forecast due date.
Our tranches contain 29 stations (24 metropolitan France,
5 overseas), as shown in Figure~\ref{fig:dbnomics-map};
each generates one question per forecast date with 8
resolution dates.
Figure~\ref{fig:dbnomics-full} shows $\sim$14 years of daily
temperature data for one station (Mont-de-Marsan), exhibiting
strong annual seasonality.

Although the stations are spatially distributed, we found
that pooling bias-corrected observations from nearby stations
does not improve forecasts (negligible BI gain).
This is because each station already has $\sim$200 historical
observations per resolution date, so the per-station
empirical exceedance estimate is already well-determined.
The main bottleneck for dbnomics is near-horizon weather
forecast data (e.g., 7-day forecasts), not statistical
estimation from historical data.

\begin{figure}[h]
  \centering
  \includegraphics[width=\textwidth]{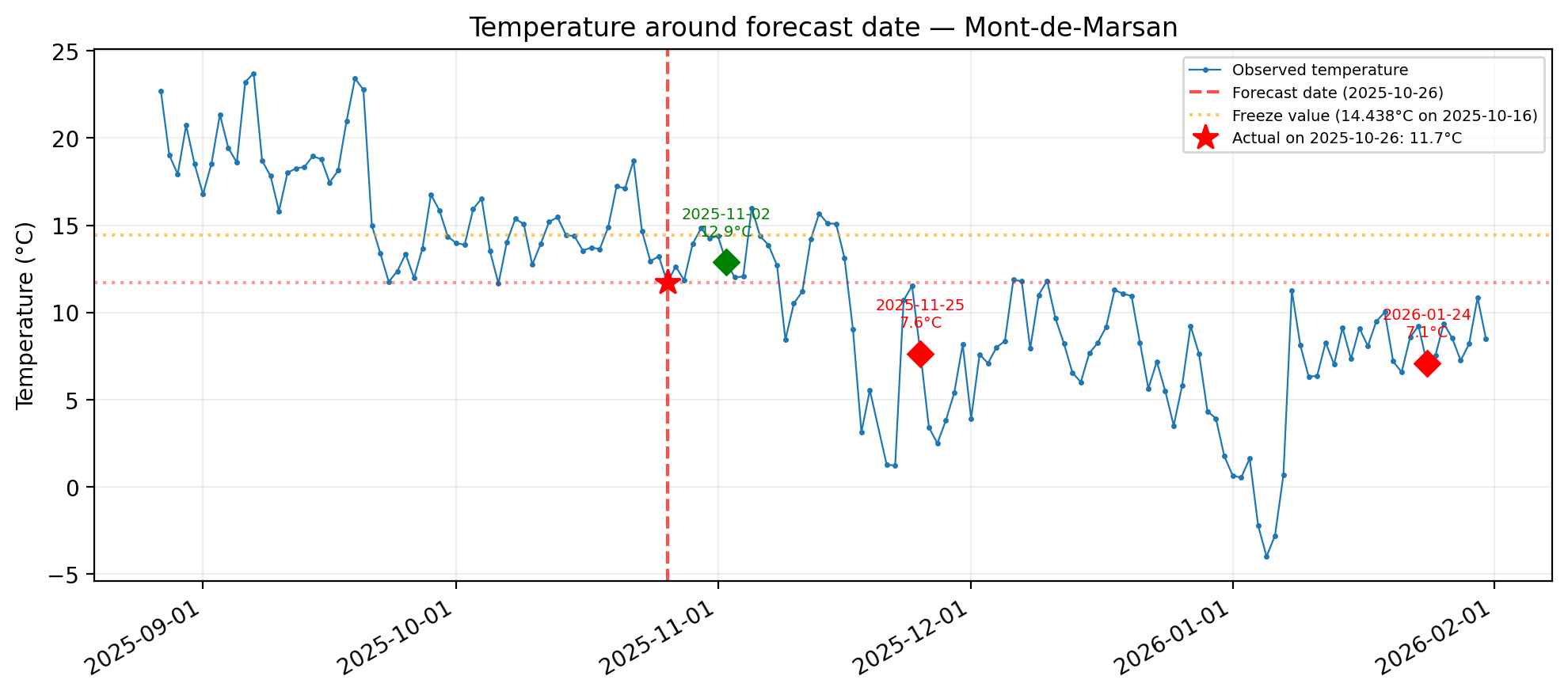}
  \caption{Temperature around the forecast date for
    Mont-de-Marsan. Star: actual value on forecast date
    (11.7$^\circ$C). Orange: stale freeze value (14.4$^\circ$C).
    Green diamond: resolution date warmer than reference;
    red: colder. The seasonal decline from October to January
    is predictable from historical data.}
  \label{fig:dbnomics-zoom}
\end{figure}

Figure~\ref{fig:dbnomics-zoom} zooms into the period around
the forecast date. The key challenge is the same as yfinance:
the question provides a \texttt{freeze\_datetime\_value} of
14.4$^\circ$C from 2025-10-16, but the actual value on the
forecast date (2025-10-26) was 11.7$^\circ$C --- the
threshold the agent must predict against.
Our \texttt{fetch\_ts\_dbnomics} tool retrieves the correct
value and computes an empirical exceedance probability
(see Appendix~\ref{app:ts-models}).

Unlike yfinance (random walk), temperature has strong seasonal
structure: the October-to-January decline is predictable from
$\sim$14 years of historical data.
Our empirical exceedance model achieves good BI by computing
$P(\text{temp} > \text{threshold})$ from same-calendar-date
observations in prior years
(see \cref{app:ts-models} for model comparison).
We bypass the LLM entirely for DBnomics,
since LLM reasoning does not improve on the statistical model.
Our system outperforms Cassi on this source
(BI \bi{60.8} vs Cassi \bi{51.0}).
Further improvement likely requires
short-term weather forecast data for near-horizon
resolution dates.

\paragraph{Example question (verbatim, abbreviated).}
\begin{small}
\begin{verbatim}
{
  "forecast_due_date": "2025-10-26",  // global
  "id": "meteofrance_TEMPERATURE_celsius.07607.D",
  "source": "dbnomics",
  "question": "What is the probability that the
    daily average temperature at Mont-de-Marsan
    will be higher on {resolution_date} than on
    {forecast_due_date}?",
  "url": "https://db.nomics.world/meteofrance/
    TEMPERATURE/celsius.07607.D",
  "freeze_datetime": "2025-10-16T00:00:00+00:00",
  "freeze_datetime_value": "14.438",
  "resolution_dates": ["2025-11-02", "2025-11-25",
    "2026-01-24", ...],
  "resolved_to": [1, 0, 0, ...]  // True, False, False
}
\end{verbatim}
\end{small}

\subsection{FB: ACLED (armed conflict)}
\label{app:fb-acled}

\begin{figure}[h]
  \centering
  \includegraphics[width=\textwidth]{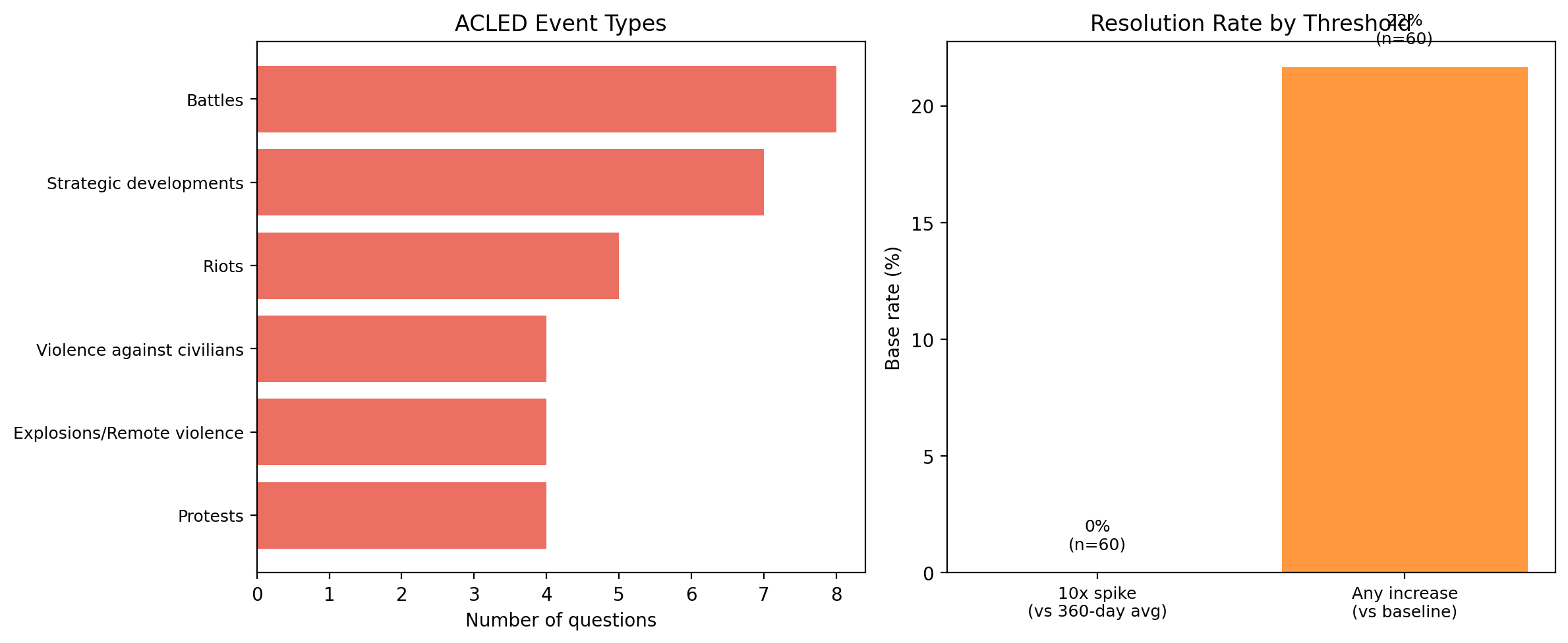}
  \caption{ACLED questions in A$\cup$B.
    Left: event type distribution.
    Right: base rate by threshold type.
    The ``10x spike'' questions never resolved True.}
  \label{fig:acled-summary}
\end{figure}

ACLED (Armed Conflict Location \& Event Data) questions ask
whether the number of conflict events of a given type in a
given country will exceed a threshold over a future 30-day
window. There are two question templates
(Figure~\ref{fig:acled-summary}):

\begin{itemize}
  \item \textbf{Any increase} (20 questions): will events exceed
  the baseline level? These have a 25\% base rate (5/20).
  Example:
  \begin{quote}
\small
\textbf{Any increase:}
``Will there be more `Protests' in Sri Lanka for the 30 days
before \texttt{\{resolution\_date\}} compared to the 30-day
average of `Protests' over the 360 days preceding 2025-11-09?''\\
Resolution dates: 2025-11-16, 2025-12-09, 2026-02-07.\\
Resolved: False, False, True (protest activity surged later).
\end{quote}

\item \textbf{10x spike} (20 questions): will events in the
  next 30 days exceed \emph{ten times} the 360-day rolling
  average? These have a 0\% base rate --- no country in our
  tranches experienced such an extreme spike. Example:
  \begin{quote}
\small
\textbf{10x spike:}
``Will there be more than ten times as many fatalities in
Finland for the 30 days before \texttt{\{resolution\_date\}}
compared to \emph{one plus} the 30-day average of fatalities
over the 360 days preceding 2025-11-09?''\\
The ``one plus'' ensures the threshold is non-zero even when
the baseline average is~0.
Resolved False on all 3 dates.
\end{quote}
\end{itemize}

\noindent
Questions span 30 countries (from Antarctica to Venezuela)
and 6 event types (Battles, Strategic developments, Riots,
Violence against civilians, Explosions/Remote violence,
Protests).
No source-specific data tool is available for ACLED (the API
requires special access), so the agent relies entirely on
web search.

The very low overall base rate (12.5\%, 5 True out of 40) means that
predicting ``No'' for all questions is a strong baseline.
The 10x-spike questions are particularly easy: since none
resolved True, any well-calibrated system should predict
near 0 for these. The agent's main challenge is the
``any increase'' questions, which require understanding
recent conflict dynamics via web search.

\subsection{FB: Wikipedia}
\label{app:fb-wikipedia}

\begin{figure}[h]
  \centering
  \includegraphics[width=\textwidth]{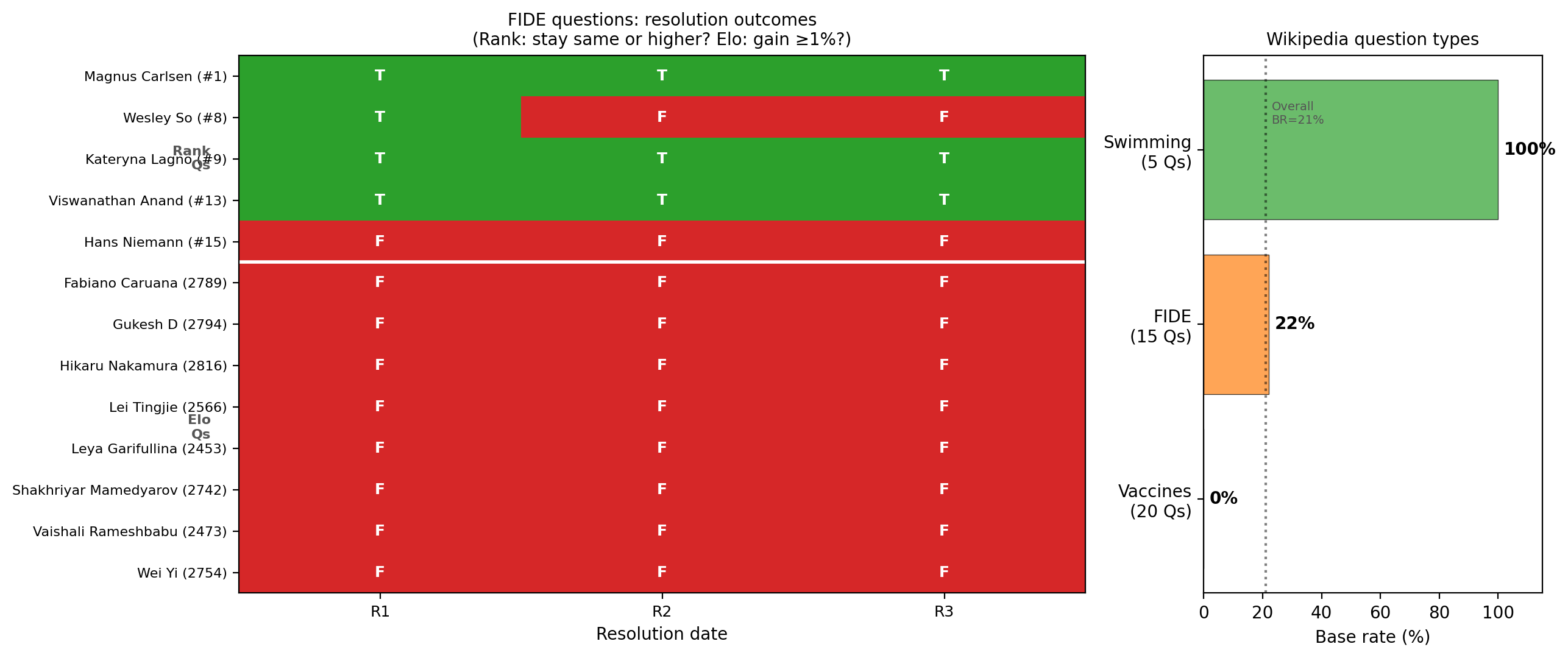}
  \caption{Left: FIDE question outcomes. Rank questions (top 5)
    mostly resolve True; Elo $\geq$1\% questions (bottom 10)
    all resolve False. Wesley So (\#8) is the only mixed case
    (True, False, False --- his ranking briefly improved then dropped).
    Right: base rates for the three Wikipedia question types.}
  \label{fig:wiki-fide}
\end{figure}

Wikipedia questions  ask whether
specific facts on a Wikipedia page will change by the
resolution date. Our 40 questions come from three articles:

\begin{itemize}
\item \textbf{List of infectious diseases} (20 questions,
  BR=0\%): ``Will a vaccine have been developed for
  [disease] by [resolution date]?'' Since vaccine development
  takes years, none of these resolved True over our
  1--3 month horizons.
\item \textbf{FIDE rankings} (15 questions, BR=27\%, 4/15):
  either ``Will [player]'s FIDE ranking be as high or higher?''
  (5 rank questions) or ``Will [player]'s Elo rating be at least
  1\% higher?'' (10 Elo questions).
  The rank questions are easier (4/5 always True),
  while gaining 1\% Elo in 1--3 months is very rare (0/10 True).
  See Figure~\ref{fig:wiki-fide}.
\item \textbf{List of world records in swimming}
  (5 questions, BR=100\%): ``Will [swimmer] still hold the
  world record for [event]?'' All resolved True (world
  records are rarely broken in a few months).
\end{itemize}

The overall base rate is 22.5\% (9 True out of 40), but it varies
dramatically by article (0\%, 27\%, 100\%).
Our system achieves very high BI on Wikipedia
(\bi{83.0}; see Figure~\ref{fig:bi-by-source})
because the questions are often answerable
from the structure of the question itself:
vaccine questions are almost certainly No,
world record questions are almost certainly Yes.
The \texttt{fetch\_wikipedia\_toc/section} tools
allow the agent to verify these priors by checking
the actual Wikipedia page content.

\clearpage
\section{Time-Series Forecasting Models}
\label{app:ts-models}

In  this section we discuss specialized statistical modeling tools which we 
optionally make available to the agent for certain data sources.
These all return an estimate of
$\hat{p}_q^r$ defined in \cref{eqn:classifier}.
These are all simple heuristic estimators that we developed to see if they
helped the LLM with its task.
Clearly many more sophisticated methods could be used,
but that is not the focus of this paper.

\subsection{KNN model for DBnomics}
\label{app:dbnomicsModel}

\begin{table}[h]
\centering
\small
\caption{Time-series model comparison on DBnomics questions
  (A$\cup$B, $n{=}40$ questions, 119 resolution dates).
  All models above the line are evaluated standalone (no LLM).
  KNN is our default for DBnomics and outperforms Cassi.}
\label{tab:ts-models-dbnomics}
\begin{tabular}{lr}
\toprule
Model & BI \\
\midrule
KNN ($\pm 10$ days, 10+ years) & \Bbi{59.1} \\
Always 0.5 (baseline) & \bi{50.0} \\
\midrule
Cassi (reference) & \bi{51.0} \\
\bottomrule
\end{tabular}
\end{table}

We tried several simple models for estimating
$\hat{p}_q^r$ for DBnomics data.
The best performing method, which we call KNN,
is  a non-parametric estimator which is analogous
to a binary $k$-nearest-neighbor classifier, where
``neighbors'' are selected by day-of-year proximity.
More precisely, define the neighbor set as follows:
\begin{equation}
  \mathcal{N}_{a,w}(q,r) =
  \{y_q(t_i) : d_{\text{cyc}}(\text{doy}(t_i), \text{doy}(r)) \leq w,
  \; f_q - a < t_i < f_q \},
\end{equation}
where $\text{doy}(\cdot) \in \{1, \ldots, 365\}$ extracts the day-of-year and
\begin{equation}
  d_{\text{cyc}}(a, b) = \min(|a - b|,\, 365 - |a - b|)
  \label{eqn:doy-cyclic}
\end{equation}
is the cyclic (year-wrap-aware) distance --- so e.g.\ Dec~31 and Jan~2 are at
distance~2, not~363. $a$ is the maximum age (to encourage focusing on data
close to $q$'s forecast date), $w$ is the window size around $\text{doy}(r)$,
and $i$ indexes across all available data. (Leap days are folded onto
day~365 by Pandas' \texttt{dayofyear}; the resulting $\pm 1$ slack is
negligible compared to our $w{=}10$ window.)
The forecast is the Laplace-smoothed empirical exceedance
frequency:
\begin{equation}
  \hat{p}_q^{\text{knn}}(a,w) = 
  \frac{|\{y_q(t_i) \in \mathcal{N}_{a,w}(q,r) : y_q(t_i) > v_q\}| + 1}
                     {|\mathcal{N}_{a,w}(q,r)| + 2}.
  \label{eqn:knn}
\end{equation}
This directly estimates
\cref{eqn:classifier} without
distributional assumptions.
For DBnomics, we use a window of $w=10$ days, and set $a=\infty$ 
to remove age restrictions,
reflecting an assumption the periodic distribution is stationary.
With $\sim$14 years of daily data, each resolution date
typically has $|\mathcal{N}_{a,w}(q,r)| \approx 270$ neighbors.\footnote{
   $2w+1=21$ days/yr times 14 years $\approx 294$.
   But some locations drop about $5\%$ of their data, 
    bringing the typical realised count down to about 270.
}

Table~\ref{tab:ts-models-dbnomics} shows results on A$\cup$B.
Our KNN method achieves BI \bi{59.1},
outperforming Cassi (\bi{51.0}) on this source.
We also tried feeding the time series directly to the LLM,
as well as linear and harmonic models,
but KNN was the best standalone approach.
Further improvement might come from short-term weather
forecast data for near-horizon resolution dates.

\subsection{Linear trend models for yfinance and FRED}
\label{app:yfinanceModel}

For non-seasonal time series, such as yfinance and FRED,
we use a parametric approach, which combines
a local linear model with the baseline estimate of 0.5:
\begin{equation}
  p_q^{\text{safe-linear}}(\alpha) = \alpha \cdot p_q^{\text{linear}} + (1 - \alpha) \cdot 0.5
  \label{eqn:safe-linear}
\end{equation}
where $\alpha$ is a source-specific shrinkage factor
controlling how much to trust the linear extrapolation.
We set $\alpha = 0.1$ for yfinance,
which reflects that stock prices follow
an approximate random walk, so the linear extrapolation
is unreliable.
We set $\alpha=0.5$ for FRED.
(All $\alpha$  values were hand-tuned based on
informal experimentation; systematic cross-validation
is left to future work.)

The linear probability estimate is given by
\begin{equation}
  p_q^{\text{linear}} = 
    \Phi^c\!\left(\frac{v_q - \hat{y}(r)}{\hat\sigma}\right)
    \label{eqn:linear}
\end{equation}
where $\Phi^c$ is the standard normal survival function,
and  $\hat{y}(t) = at + b$ is the linear model
fit by  OLS on 
the most recent $W$ observations,
$\mathcal{H}(s_q,f_q;W)$.
(We use $W=30$ for FRED and $W=60$ for yfinance.)
$\hat{\sigma}$ is the standard deviation of the residuals
for this data source $s_q$.
(We could make this a function of forecast horizon, $r-f_q$, as well as source type,
$s_q$, but we leave that to future work.)

\subsection{Hybrid KNN and linear trend model for yfinance and FRED}

For FRED and yfinance,
we also consider a hybrid parametric and non-parametric
estimate of the form 
\begin{eqnarray} 
\hat{p}_q^{\text{hybrid}}(\alpha, \beta) &=
 \beta \cdot p_q^{\safeLinear}(\alpha)
 + (1-\beta) \cdot p_q^{\text{prior-year}} 
 \label{eqn:hybrid}
 \end{eqnarray}
 where  $p_q^{\text{prior-year}}=p_q^{\text{knn}}(a,w)$
 uses the same nonparametric
exceedance estimate as \cref{eqn:knn}, but with a
narrower window ($w=7$ days) and limited to the 5 most recent years,
by setting $a=5 \times 365$ days.
This captures annual periodicity: for a given
resolution date, we check whether the value exceeded the
threshold at the same time of year in prior years.
Expanding out, this estimate is a convex combination of three terms,
the linear forecast, the prior-year (KNN) forecast, and the
uniform $0.5$ fallback:
\begin{eqnarray}
\hat{p}_q^{\text{hybrid}}(\alpha,\beta) &=
\alpha \beta \, p_q^{\text{linear}}
 + (1-\beta) \, p_q^{\text{prior-year}}
 + (1-\alpha) \beta \cdot 0.5,
 \label{eqn:phybrid}
\end{eqnarray}
where the three weights sum to one (since
$\alpha\beta + (1{-}\beta) + (1{-}\alpha)\beta = \beta + (1{-}\beta) = 1$).

We set $\beta=0.5$ for both FRED and yfinance when prior-year
data is available ($\beta=1$ otherwise, falling back to
$p^{\safeLinear}$ alone).
For DBnomics, the hybrid model is not used --- we apply the
KNN model (\cref{eqn:knn}) directly.

\clearpage
\section{Extended Related Work}
\label{app:related}

We organize the growing literature on LLM-based forecasting
into eight themes.

\paragraph{Benchmarks.}
\citet{Zou2022} introduced Autocast, the first large-scale
dataset for neural forecasting of real-world events,
and found that LM performance was far below human experts
but improved with model scale and retrieval.
\citet{halawi2024approaching} scaled this up with
5{,}000+ binary questions from five platforms,
showing that retrieval-augmented GPT-4 approaches the
human crowd's Brier score.
ForecastBench~\citep{forecastbench}
provides a rolling benchmark with market and dataset questions,
using difficulty-adjusted Brier scores
\citep{forecastbenchUpdate}
to compare methods on a common leaderboard.
TFRBench~\citep{TFRBench} evaluates
forecasting \emph{reasoning} (not just accuracy).
FutureX~\citep{FutureX} 
and FutureX-Pro~\citep{FutureXPro} introduces a live benchmark
with daily-updated questions and an automated pipeline
to eliminate data contamination,
but focuses on 0-1 accuracy rather than probabilistic forecasting.
Metaculus FutureEval~\citep{FutureEval} provides a
continuously updated live benchmark that resists
contamination (since answers are not yet known),
with \$175K in annual prizes for bot tournaments.
OpenEP~\citep{OpenEP} moves beyond binary questions
to open-ended outcome prediction.
\citep{Liptay2026} present ``Bench to the Future 2'',
a benchmark of 1417 binary judgemental forecasting questions,
asked between 2025-10-18 and 2025-10-28,
with a resolution window in 2025-10 to 2025-12;
BTF2 also ships with a frozen
pre-retrieved corpus of documents.
\eat{
\footnote{
The document corpus is  a set of "research summaries"
(median length 12k chars/ 3k tokens), one per question,
not the raw text,  due to copyright reasons.
However, the benchmark also contains a list of scraped pages:
Each question has an average of
10,100 web pages scraped and stored offline (range: 6,000-19,000),
for a total of 16.2 million documents, of which 8.7 million pages are unique.
(For each question, the pipeline
gathers pages in a rolling window that starts 1 week before the question's
own present date and ends approximately on that date, to avoid outcome from search leakage.)
Note that BTF2  is a bigger and more recent version of
``Bench to the Future'' from \citep{FutureSearch2025}.
}
}
\citep{Goel2026} present FutureSim, which is a 
benchmark for evaluating online forecasting (continual learning)
using a fixed text corpus
(to avoid leakage). 
ProphetArena~\citep{ProphetArena} is a benchmark that bundles multiple
related binary forecasting questions into a "market" (e.g., "will A,B or C win US election"
becomes 3 binary questions), but at the time of writing, they do not release historical data,
precluding backtesting.

\paragraph{Evaluation.}
\citet{Paleka2025} provide a critical analysis of
offline evaluation pitfalls, including temporal leakage
and retrieval leakage.
\citet{SimulatedIgnorance} show that LLMs fail to
``simulate ignorance'' of events before their knowledge
cutoff, further complicating backtesting validity.
Our four-layer leakage defense (\cref{app:leakage})
addresses several of these concerns.
Prophet Arena~\citep{ProphetArena} argues that
economic utility (profit from bets) may be more relevant
than Brier scores;
they find that LLMs show impressive calibration
but suffer from inaccurate event recall.

\paragraph{Financial forecasting and trading.}
Several benchmarks evaluate LLMs as financial predictors
and traders.
Prediction Arena~\citep{PredictionArena} deploys
six frontier models with real capital
on Kalshi and Polymarket, finding that most models
lose money ($-16\%$ to $-31\%$ on Kalshi).
PolyBench~\citep{PolyBench} evaluates seven LLMs
on $\sim$38K Polymarket questions with order-book data,
finding that only two models achieve positive returns.
FinTradeBench~\citep{FinTradeBench} evaluates financial
reasoning combining fundamentals and trading signals,
finding that retrieval helps fundamental analysis
but not time-series reasoning ---
consistent with our finding that LLMs struggle with
raw numerical data (\cref{app:tools}).
\citep{Wang2026trade} proposes a new trading benchmark.

\paragraph{Zero-shot and prompt-based forecasting.}
\citet{Karkar2025} show that LLM forecasting ability
is ``unevenly distributed'' across domains and question types
(c.f.,  \cref{fig:aibq2_ms_per_que}).
\citet{Pratt2024} found that superforecasting prompting
strategies (decomposition, base rates, retrieval) failed to
improve PaLM~2 over a basic prompt, attributing the model's
apparent accuracy to a negativity bias that happens to
align with low base rates.
\citet{Schoenegger2025prompts} tested 38 prompt variants
across GPT-4o, Claude~3.5, and Llama~3.1, finding that
most modifications yield negligible gains and some
(e.g., explicit Bayesian reasoning prompts) actively hurt.
These negative results motivate our agentic approach:
rather than engineering better prompts, we provide tools
(search, data access) and structured belief tracking.

\paragraph{Agentic and tool-augmented systems.}
\citet{Hsieh2024} proposed RTF (Reasoning and Tools
for Forecasting), a zero-shot framework using hierarchical
agents with Python REPL and Google Search,
achieving Brier scores competitive with human crowds
on Manifold Markets without any weight updates.
The AIA Forecaster~\citep{AIA} employs adaptive
iterative search
and statistical calibration (Platt scaling),
similar to us, but uses simple text aggregation rather than a belief state.
Also, they use an LLM-based calibration system,
which we found to be inferior to simple mean aggregation.
CogForecast~\citep{CogForecast} uses multi-agent debate
with diverse cognitive profiles to mitigate inherited biases.

\paragraph{Bayesian agents.}
BayesAgent~\citep{BayesAgent} also targets Bayesian reasoning under
uncertainty in LLM agents, but via \emph{verbalized probabilistic
graphical models} (vPGM): the LLM is prompted to identify
latent variables and dependencies, then they use standard Bayesian inference
algorithms applied to this model.
This is complementary to our approach: BayesAgent constructs
an explicit PGM per query, while \system{} maintains a
non-Markovian natural-language belief state updated
sequentially from tool-call evidence, avoiding the
need for an explicit graph structure.
In the position paper, \citep{Papamarkou2026,Amin2026} argue that LLM agents
should model uncertainty using a more classical sequential Bayesian paradigm,
in which the likelihood of observed text data $o_t$ (e.g. the results of
a search tool call $a_t$)
is modeled using $p(o_t|s, a_t)$, where $s$ is the unknown outcome
we are trying to forecast.
In \cref{sec:explicit-bayes} we present some preliminary results evaluating
this kind of method, and find lower accuracy than linguistic BLF for the tested likelihood
elicitation schemes on fixed replayed evidence.

\eat{
The BALAR system of \citep{Echarghaoui2026} implements
a variant of this  idea by prompting an LLM to generate likelihood functions
of the form $p(f(x_t)|h)$ for each possible hidden value $h$,
where $f(x)$ converts free-form text into a finite number of
string-valued responses;
they apply this to a sequential medical information gathering problem.
(\citep{Amin2026} adopts a similar approach for
a simulated job candidate screening agent.)
In \cref{sec:explicit-bayes}, we tested this idea (but setting
$f$ to be the identity function), but it did much worse than BLF.
Finally, \citep{Dai2026} recently proposed
an extension of BLF
in which the point estimate $\hat{p}$ is replaced by a Beta distribution,
to better capture uncertainty and calibration.
}

\paragraph{Reinforcement learning for forecasting.}
Several recent works apply RL to improve forecasting.
Time-R1~\citep{TimeR1} uses a two-stage curriculum
(temporal comprehension then prediction) to build
``temporal logic'' into the model's representations.
\citet{Turtel2025} apply RLVR (RL with Verifiable Rewards)
to train a 14B model on historical Polymarket questions,
achieving frontier-level Brier scores (0.190) and
demonstrating economic utility (10\%+ ROI in simulated trading).
OpenForecaster~\citep{Chandak2026} synthesizes 50K+
training questions from historical news and uses GRPO
with a composite accuracy+Brier reward to combat
hedging bias, showing that specialized 8B models can
match 120B+ generalist models.
\citet{scott2026forecasting,Jeen2026} use RLFT on open source models
for the Metaculus AI Benchmark Tournament,
achieving the current AIBQ2 SOTA (MS=45.8).
 \citep{Damani2026} use RLFT to improve calibration.
\citet{Turtel2026} introduce ``Foresight Learning'' (the
training method behind the \emph{Foresight-32B} model from
Lightning Rod Labs that we compare against in
\cref{tab:mega}), using resolved outcomes as free supervision
via proper scoring rule rewards, showing that a trained
Qwen3-32B outperforms the untrained Qwen3-235B
(27\% better Brier score).
\citep{Auzina2026} uses RL where the reward function is related
to the change in the belief state, rather than just the final outcome.
These RL approaches are complementary to our work:
they improve the \emph{base model}, while we improve the
\emph{agent harness} (tools, belief tracking, calibration).

\paragraph{Ensemble methods.}
\citet{Schoenegger2024crowd} showed that aggregating
12 diverse LLMs (a ``silicon crowd'') matches
human crowd accuracy, and that frontier models can
update beliefs when shown human median forecasts.
Ensemble gains depend on diversity among the component forecasts
(see also \citealp{Aitchison2026ensemble}).

\clearpage
\section{Prompts}
\label{app:prompts}

This appendix lists the prompts used in our system
and the zero-shot baseline.

\subsection{System Prompt}
\label{app:system-prompt}

The following is the system prompt used for atyes
ll
\system{} configurations (abbreviated for space;
source-specific tool descriptions omitted).

\begin{quote}
\small\ttfamily
You are an expert superforecaster. Your task is to predict the
probability that a binary question will resolve to YES given
information up to a certain date.

You work in a tool-use loop:\\
1. Read the question, its resolution criteria, and background.\\
2. Form a base rate estimate (outside view / reference class reasoning).\\
3. Perform a loop:\\
\quad 3a. Choose ONE tool to call.\\
\quad 3b. After each tool call, your belief state is updated.\\
4. When you have gathered enough evidence, call submit(probability, reasoning).

Belief state rules:\\
- Evidence lists should ACCUMULATE across steps.\\
- Each evidence item MUST cite its source (search\_X\_result\_Y).\\
- Include update\_reasoning explaining WHY evidence changed your probability.\\
- Consider RECENCY and AUTHORITATIVENESS of sources.

Rules:\\
- You MUST call submit before step \{max\_steps\}.\\
- Call submit once probability has stabilized.\\
- Probabilities must be between 0.05 and 0.95.
\end{quote}

\subsection{NoBel System Prompt}
\label{app:nobel-prompt}

For the NoBel ablation we use a separate, simplified system prompt
that does \emph{not} mention belief state, evidence-accumulation
rules, or an \texttt{updated\_belief} field --- the tool schemas
also have the \texttt{updated\_belief} parameter stripped (see
\cref{app:tools}). The prompt is reproduced below (abbreviated
the same way as \cref{app:system-prompt}). Crucially, neither the
prompt nor the tool schema asks the agent to maintain a structured
belief state, so the agent operates by accumulating raw search
results in its message history.

\begin{quote}
\small\ttfamily
You are an expert superforecaster. Your task is to predict the
probability that a binary question will resolve to YES given
information up to a certain date.

You work in a tool-use loop:\\
1. Read the question, its resolution criteria, and background.\\
2. Perform a loop:\\
\quad 2a. Choose ONE tool to call.\\
\quad 2b. The tool result will be added to the conversation.\\
3. When you have gathered enough evidence, call submit(probability, reasoning).

Suggested strategy:\\
- Use web\_search to find specific evidence.\\
- Use summarize\_results to read the full content of promising results.\\
- Consider all the evidence you have gathered when making your final
probability estimate.

Rules:\\
- You MUST call submit before step \{max\_steps\}. If you reach the last
step without submitting, a default probability of 0.5 will be used.\\
- Call submit once your sense of the probability has stabilized.\\
- Probabilities must be between 0.05 and 0.95.
\end{quote}

\noindent
The same template-substitution machinery that fills
\texttt{\{max\_steps\}}, the source-specific tool list, and the
backtest-vs-live mode paragraph in the BLF prompt also runs on this
template; the only difference is the loop-and-rules text shown above.

\subsection{Question Prompt}
\label{app:question-prompt}

Each question is formatted as follows (with crowd/prior
sections included only when \texttt{crowd=1} or \texttt{emp=1}):

\begin{quote}
\small\ttfamily
\# Question\\
\{question text\}

\#\# Background and resolution criteria\\
\{background\}\\
\{resolution\_criteria\}

\#\# Resolution dates\\
\{dates\}\\
You must submit \{n\} probabilities (one per resolution date).
Your uncertainty should INCREASE with forecast horizon.

\#\# Knowledge cutoff\\
\{cutoff\_date\}\\
You must not use any information from after this date.

\#\# Market estimate \textrm{(if crowd=1, market questions only)}\\
The market estimate on \{date\} was \{value\}.

\#\# Prior estimate \textrm{(if emp=1, dataset questions only)}\\
\{explanation\}: \{prior\}.\\
Use this as your starting point, but adjust based on
question-specific evidence from search and tools.
\end{quote}

\subsection{Zero-Shot Prompt}
\label{app:halawi-prompt}

For the zero-shot baseline (\texttt{search=none}, \texttt{tools=0})
we use the same standard system prompt as \system{}
(\cref{app:system-prompt}); the only difference at runtime is that
the tool schema contains only \texttt{submit}, so the agent reads
the question and immediately submits a probability without
performing search or any other tool call. The loop-rules and
belief-state portions of the standard prompt are effectively
dead text in this mode.
\eat{
--- empirically the agent terminates on
its very first step in 1995/2000 ZS trials (median $n_{\text{steps}}=1$,
mean $1.00$), so what matters is the question framing rather than
the loop instructions.
 }
 Interestingly, we found that switching the entire prompt
to the simplified zero-shot template from
\citet{halawi2024approaching} (also used by ForecastBench for its
zero-shot leaderboard entries) gave worse performance ($65.3$~BI
on FB A$\cup$B for ZS+crowd+emp, vs $70.6$~BI using our standard
prompt). The Halawi template is shown below for reference; the
experiments reported in this paper use our standard prompt
throughout.

\begin{quote}
\small\ttfamily
You are an expert superforecaster, familiar with the work of
Tetlock and others. Make a prediction of the probability that
the question will be resolved as true. You MUST give a probability
estimate between 0 and 1 UNDER ALL CIRCUMSTANCES. If for some
reason you can't answer, pick the base rate, but return a number
between 0 and 1.

Question: \{question\}\\
Question Background: \{background\}\\
Resolution Criteria: \{resolution\_criteria\}\\
Question close date: \{close\_date\}\\
\textrm{[If freeze value available:]}
The freeze value is \{value\}. \{explanation\}

Output your answer (a number between 0 and 1) with an asterisk
at the beginning and end of the decimal. Do not output anything else.\\
Answer:\\
\{\{ Insert answer here \}\}
\end{quote}

\subsection{Search-Result Date-Filter Prompt}
\label{app:filter-prompt}

Web search results are often metadata-stale: a page's
\texttt{page\_age} reflects its original publish date, but the page
body may have been updated with post-cutoff content (liveblogs,
``last updated'' patches, automatically-appended related links, etc.).
Our second pass strips the metadata age flags and asks an LLM to
re-judge each result purely on its \emph{textual content}, dropping
anything that describes events or outcomes after the knowledge
cutoff. This is the layer audited in \cref{app:leakage} (the leak
detective). The exact prompt (taken from
\texttt{src/search/search\_lib.py:filter\_results}; the LLM is
Gemini-3-Flash by default):

\begin{quote}
\small\ttfamily
You are a strict information-cutoff filter. The knowledge cutoff
date is \{cutoff\_date\}.\\
For each numbered search result, decide KEEP or DROP.\\

Focus on the CONTENT of each result --- the highlights, extra
snippets, and any dates mentioned in the text body. Do NOT trust
the `Published on' date alone; pages are often updated after their
original publish date.\\

Rules:
\begin{itemize}\small\ttfamily
\item If ANY part of the content describes events, outcomes, or
updates from AFTER \{cutoff\_date\} → DROP the entire result
\item Look for dates in the text body (e.g.\ `January 25, 2026',
`February 6, 2026'). If any mentioned date is after
\{cutoff\_date\} and the text describes what happened on that
date → DROP
\item Past tense describing events scheduled after cutoff
(e.g.\ `the climb was postponed', `he completed the ascent') →
DROP
\item Liveblog/timeline pages with entries after cutoff → DROP
\item Content only about events before or on the cutoff date →
KEEP
\end{itemize}

Reply with one line per result: the number, a colon, KEEP or DROP,
then a brief reason.\\
Example:\\
1: KEEP - article from Nov 2025 about the announcement\\
2: DROP - contains Jan 25 update describing postponement\\

Results:\\
\{numbered\_results\_with\_age\_flags\_stripped\}
\end{quote}

\noindent The \texttt{KEEP/DROP} decisions are parsed by regex; any
result the LLM does not explicitly mark as \texttt{DROP} is
retained (conservative AND with the client-side \texttt{page\_age}
filter that already runs inside the search-engine wrapper).

\subsection{Search-Result Summarization Prompt}
\label{app:summarize-prompt}

After the agent calls \texttt{web\_search} and selects a subset of the
returned snippets via \texttt{summarize\_results}, the selected
results are concatenated and passed to a separate
``summarizer'' LLM (Gemini-3-Flash by default; see
\cref{app:llms}) for compression. We use a question-aware prompt
that extracts only facts relevant to the resolution criteria,
rather than producing a generic summary --- this both shortens the
text the agent must reason over and reduces the risk of injecting
irrelevant evidence into the belief state.
The exact system prompt (taken from
\texttt{src/search/search\_lib.py:summarize\_results}):

\begin{quote}
\small\ttfamily
You are an assistant to a superforecaster. Extract the facts and
data from the search results below that are most relevant to
predicting the outcome of this question.\\

Question: \{question\}\\

Resolution criteria: \{resolution\_criteria\}\\

Instructions:
\begin{itemize}\small\ttfamily
\item Extract concrete facts, statistics, dates, and named-source
expert opinions.
\item Note any quantitative data (prices, percentages, counts,
trends).
\item Distinguish hard facts from speculation or editorial opinion.
\item Omit information that is not relevant to the resolution
criteria.
\item Do NOT add your own analysis or forecast --- only extract
what the sources say.
\end{itemize}

Search results:\\
\{filtered\_search\_results\}
\end{quote}

\noindent When \texttt{resolution\_criteria} is unavailable
(short-form summarization fallback), the prompt collapses to
``Summarize the following search results clearly and concisely,
highlighting the most relevant facts for answering the question:
\{question\}'', followed by the same \texttt{\{filtered\_search\_results\}}
block.

\subsection{Likelihood estimation prompts}
\label{app:likelihoodPrompts}

\paragraph{The likelihood ratio prompt.}
Each weight uses a fixed system prompt and a user prompt instantiated once per
state $s\in\{\textsc{yes},\textsc{no}\}$; the action, prior-evidence, and summary
lines appear only for the conditioning variants that use them.
\begin{quote}
\small\ttfamily
[SYSTEM] You estimate how typical an observation is under a hypothesized outcome
of a binary forecasting question. Use forward simulation: assume the stated
outcome is the truth, then judge how typical/representative the evidence would be
of a world in which that outcome holds. Output a single integer 0-10: 10 = highly
typical, 5 = neutral, 0 = completely atypical. Output ONLY the integer.\\[4pt]
[USER] Binary question: \{question\}\\
The agent issued this search query: \{a\_t\}.\\
Evidence already gathered before this search: \{b\_\{t-1\}.h\}\\
NEW evidence retrieved by this search: \{o\_t\}\\
Assume the TRUE outcome is: \{state\}. How typical/representative is the evidence
above of a world in which the answer is \{state\}? Judge only what the NEW
evidence adds beyond what is already known. Answer with a single integer 0-10.
Output only the integer.
\end{quote}
\noindent
The model is queried twice (\{state\}$=$\textsc{yes} and \textsc{no}), giving the
two typicality scores in \cref{eqn:per-state}.

As an example, consider the NextEra Energy (NEE) stock question.
The belief summary
$b_{t-1}.h$ (the value substituted for the prior-evidence line)
 taken mid-trajectory  is as follows:
\begin{quote}
\small\ttfamily
Evidence FOR: Stocks generally go up over time (base rate \textgreater\ 0.5);
Linear trend from Aug to Oct is positive (slope = +0.16/day)\\

Evidence AGAINST: Trend analysis gives combined estimate of 0.41; Price has been
slightly trending down/flat over the last 10 days of October\\

Open questions: Earnings dates and ex-dividend dates for NEE in Q4 2025; Any
recent major news affecting NEE stock
\end{quote}

\end{document}